\PassOptionsToPackage{unicode}{hyperref}
\PassOptionsToPackage{hyphens}{url}
\documentclass[11pt]{article}
\usepackage[margin=1in]{geometry}
\usepackage{xcolor}
\usepackage{newunicodechar}
\usepackage{amsmath,amssymb}
\usepackage{iftex}
\ifPDFTeX
  \usepackage[T1]{fontenc}
  \usepackage[utf8]{inputenc}
  \usepackage{textcomp} 
\else 
  \usepackage{unicode-math} 
  \defaultfontfeatures{Scale=MatchLowercase}
  \defaultfontfeatures[\rmfamily]{Ligatures=TeX,Scale=1}
\fi
\newunicodechar{≤}{\ensuremath{\leq}}
\newunicodechar{≈}{\ensuremath{\approx}}
\newunicodechar{↔}{\ensuremath{\leftrightarrow}}
\newunicodechar{⇒}{\ensuremath{\Rightarrow}}
\newunicodechar{−}{\ensuremath{-}}
\newunicodechar{±}{\ensuremath{\pm}}
\newunicodechar{Δ}{\ensuremath{\Delta}}
\newunicodechar{ρ}{\ensuremath{\rho}}
\newunicodechar{×}{\ensuremath{\times}}
\newunicodechar{⁻}{\textsuperscript{-}}
\newunicodechar{⁰}{\textsuperscript{0}}
\newunicodechar{¹}{\textsuperscript{1}}
\newunicodechar{²}{\textsuperscript{2}}
\newunicodechar{³}{\textsuperscript{3}}
\newunicodechar{⁴}{\textsuperscript{4}}
\usepackage{lmodern}
\ifPDFTeX\else
  \newfontfamily\khmerfont[
    Path=fonts/,
    Script=Khmer,
    UprightFont=NotoSerifKhmer-Regular.ttf,
    BoldFont=NotoSerifKhmer-Bold.ttf,
    ItalicFont=NotoSerifKhmer-Regular.ttf,
    BoldItalicFont=NotoSerifKhmer-Bold.ttf
  ]{NotoSerifKhmer-Regular.ttf}
\fi
\ifXeTeX\XeTeXgenerateactualtext=1\fi
\IfFileExists{upquote.sty}{\usepackage{upquote}}{}
\IfFileExists{microtype.sty}{
  \usepackage[]{microtype}
  \UseMicrotypeSet[protrusion]{basicmath} 
}{}
\usepackage{color}
\usepackage{fancyvrb}
\usepackage{fvextra}

\DefineVerbatimEnvironment{Highlighting}{Verbatim}{commandchars=\\\{\},fontsize=\small,breaklines,breakanywhere,breaksymbolleft={}}
\newenvironment{Shaded}{}{}

\newcommand{\NormalTok}[1]{#1}

\usepackage{longtable,booktabs,array}
\usepackage{caption}
\usepackage{calc} 
\usepackage{etoolbox}
\makeatletter
\patchcmd\longtable{\par}{\if@noskipsec\mbox{}\fi\par}{}{}
\makeatother
\newcommand{\tablecaptiongap}{%
  \global\everypar{%
    \small
    \global\everypar{\global\everypar{}\normalsize\vadjust pre{\smallskip}}%
  }}
\AfterEndEnvironment{longtable}{\tablecaptiongap}
\pretocmd{\section}{\normalsize\global\everypar{}}{}{}
\pretocmd{\subsection}{\normalsize\global\everypar{}}{}{}
\pretocmd{\subsubsection}{\normalsize\global\everypar{}}{}{}
\IfFileExists{footnotehyper.sty}{\usepackage{footnotehyper}}{\usepackage{footnote}}
\makesavenoteenv{longtable}
\usepackage{graphicx}
\usepackage{pdflscape}
\makeatletter
\newsavebox\pandoc@box
\newcommand*\pandocbounded[1]{
  \sbox\pandoc@box{#1}%
  \Gscale@div\@tempa{\textheight}{\dimexpr\ht\pandoc@box+\dp\pandoc@box\relax}%
  \Gscale@div\@tempb{\linewidth}{\wd\pandoc@box}%
  \ifdim\@tempb\p@<\@tempa\p@\let\@tempa\@tempb\fi
  \ifdim\@tempa\p@<\p@\scalebox{\@tempa}{\usebox\pandoc@box}%
  \else\usebox{\pandoc@box}%
  \fi%
}
\def\fps@figure{htbp}
\makeatother
\NewDocumentCommand\citeproctext{}{}

\makeatletter
 \let\@cite@ofmt\@firstofone
 \def\@biblabel#1{}
 \def\@cite#1#2{{#1\if@tempswa , #2\fi}}
\makeatother
\newlength{\cslhangindent}
\newlength{\csllabelwidth}
\newenvironment{CSLReferences}[2] 
 {\begin{list}{}{%
  \setlength{\itemindent}{0pt}
  \setlength{\leftmargin}{0pt}
  \setlength{\parsep}{0pt}
  \ifodd #1
   \setlength{\leftmargin}{\cslhangindent}
   \setlength{\itemindent}{-1\cslhangindent}
  \fi
  \setlength{\itemsep}{#2\baselineskip}}}
 {\end{list}}
\usepackage{calc}

\providecommand{\tightlist}{%
  \setlength{\itemsep}{0pt}\setlength{\parskip}{0pt}}
\usepackage{bookmark}
\IfFileExists{xurl.sty}{\usepackage{xurl}}{} 
\makeatletter
\@ifundefined{xmpquote}{}{}
\makeatother
\hypersetup{
  pdftitle={M-GATE: Multilingual Grammar{,} Accuracy in Translation{,} and Efficiency Benchmark for Large Language Models},
  pdfauthor={Tomáš Burkert; Angelika Peljak-Łapińska; David Zelený},
  hidelinks,
  pdfcreator={LaTeX via pandoc}}

\title{M-GATE: Multilingual Grammar, Accuracy in Translation, and Efficiency Benchmark for Large Language Models}
\author{Tomáš Burkert\\{\small RWS TrainAI}\\\texttt{\small tomas.burkert@rws.com} \and Angelika Peljak-Łapińska\\{\small RWS TrainAI} \and David Zelený\\{\small RWS TrainAI}}
\date{2026-08-04}

\begin{document}
\maketitle

\begin{abstract}

Multilingual language models are deployed across a hundred or more languages, yet most benchmarks test whether a model can perform a task \emph{in} a language rather than whether it commands the language itself, conflating fluency with proficiency. We introduce M-GATE (Multilingual Grammar, Accuracy in Translation, and Efficiency), a benchmark of linguistic proficiency spanning 30 typologically diverse languages from high- to low-resource. M-GATE comprises three tasks: grammatical error detection on linguist-crafted, adversarially selected sentences that turn on hard, language-specific phenomena; round-trip translation of shared English sources across 29 target languages, scored by a three-provider LLM judge panel validated against professional annotators; and a supplementary tokenizer-efficiency measure. We evaluate over 50 models in more than 80 configurations. Fluency and proficiency come apart sharply: models that translate competently sit near chance on the adversarial grammar items, the best reaching a Matthews correlation coefficient (MCC) of only 0.36, and their errors lean systematically toward under-flagging, accepting ungrammatical text rather than raising false alarms. Translation quality closely tracks a language's share of pretraining data (r = 0.86 against log Common Crawl share), producing a steep low-resource penalty that is nonetheless narrowing with successive model releases. Enabling reasoning reliably improves translation, while its effect on error detection is smaller and for some models negative, so the best configuration is task-dependent. To resist contamination, test items are kept private behind a continuously updated public leaderboard, with illustrative examples released (\href{https://m-gate.ai}{m-gate.ai}).

\end{abstract}

\section{1. Introduction}\label{introduction}

\section{1.1 The Linguistic Proficiency Gap}\label{the-linguistic-proficiency-gap}

A large language model (LLM) today can produce a Finnish paragraph that reads like a native speaker's, yet the same model will accept a Finnish sentence whose suffix violates vowel harmony, a mistake a child would catch. The two behaviors look contradictory only if we assume that producing fluent text and understanding a language are the same capability in large language models. They are not. Fluency is the ability to generate text that looks right. It does not guarantee two abilities that many real uses depend on: telling whether text is right, and preserving meaning when moving between languages. A model can be highly fluent while being surprisingly weak at both.

This distinction is easy to miss because most of what we ask of multilingual models, and most of how we evaluate them, rewards fluency. A model that has absorbed the statistical regularities of a language will generate plausible text and will accept as normal the errors that are common in that text, including the ones native speakers themselves make. What it will not reliably do is enforce the prescriptive standard of the language or notice a subtle, language-specific violation. That is precisely the competence that matters wherever linguistic precision is the point: automated writing assistance and grammar checking, localization quality assurance, review of legal, medical, and regulatory content, and content moderation outside English. In all of these a model is asked to judge text against a standard, and a model that is merely fluent will let errors pass with apparent confidence.

The gap is growing in visibility, because models are now deployed across a hundred or more languages while our evidence about their actual command of those languages lags far behind their claimed coverage. Existing multilingual benchmarks largely test whether a model can perform a task in a language rather than whether it commands the language itself, and their language sets tend to concentrate on higher-resource languages, folding the rest into an average or omitting them, so the cases where proficiency is weakest are the least visible. As a result, there is little basis for knowing where a model is genuinely proficient and where it is only fluent. This affects everyone who depends on that knowledge: the companies deploying these models to global audiences for content and text work, from generation to localization; the researchers comparing models; and the people relying on the systems in their own languages.

M-GATE is built to provide that basis. We call this target linguistic proficiency: whether a model reliably judges what is and is not grammatical in a language and preserves meaning when working in it. We measure it with two complementary tasks, grammatical error detection on linguist-crafted sentences that turn on phenomena absent from English, and round-trip translation scored by comparing the original English with the result, across 30 typologically diverse languages that deliberately span from the highest-resource languages to low-resource ones such as Kinyarwanda and Dzongkha. A central finding is that this gap is real and large: models that translate competently often sit near chance at detecting grammatical errors, and grammatical and translation ability correlate only moderately across models, diverging sharply for some.

\section{1.2 The Gap in Existing Benchmarks}\label{the-gap-in-existing-benchmarks}

The benchmarks that dominate multilingual evaluation do not probe this competence. The most common design translates an English knowledge or reasoning benchmark into other languages, so that a correct answer depends on factual recall rather than on any sensitivity to the target language's grammar; other suites deliberately test language-agnostic skills such as mathematics and code across languages, and parallel corpora provide translation data without prescribing how meaning preservation should be judged. None directly measures whether a model has internalized a language's grammar or can preserve meaning under translation, and most report a single aggregate that obscures low-resource performance. We survey this landscape and position our approach against it in §2.

\section{1.3 Contributions}\label{contributions}

We make the following contributions:

\begin{enumerate}
\def\labelenumi{\arabic{enumi}.}
\item
  \textbf{A benchmark for linguistic proficiency.} We introduce a multilingual benchmark that measures whether models reliably detect violations of a language's grammar and can preserve meaning under translation, rather than whether they can perform knowledge or reasoning tasks phrased in that language.
\item
  \textbf{Linguist-crafted, adversarially selected grammatical items.} For each language, professional linguists author grammatical error-detection items that turn on phenomena present in the target language but not in English, cross-validate them, and select the hardest, giving a task that isolates grammatical competence from world knowledge and surface fluency (§3.2).
\item
  \textbf{Round-trip translation with a human-validated judge panel.} We adopt round-trip translation, scored by an English-to-English comparison that needs neither reference translations nor a judge more capable than the tested models, and make it trustworthy through a hardened, category-structured source set, a three-provider judge panel with measured self-preference bias, and direct validation against professional annotators (§3.3, §4). Round-trip translation for multilingual evaluation is also proposed in concurrent work (Skorobogat et al. 2026); our contribution is its combination with the grammar task and this validation, discussed in §6.
\item
  \textbf{Empirical findings across 53 models in 82 configurations.} These include a large fluency-proficiency gap with most models near chance at grammatical error detection, a task-dependent effect of reasoning that helps translation but degrades some models' grammar judgments, a sharp split in what cost buys between the two tasks, and rapid recent progress on low-resource languages (§5).
\item
  \textbf{A continuously maintained, contamination-resistant benchmark.} Results are reported on a living public leaderboard (\href{https://m-gate.ai}{m-gate.ai}) across 30 typologically diverse languages with substantial low-resource coverage, while the test items are kept private to resist contamination, with only illustrative non-live examples released (§7).
\end{enumerate}

\begin{center}\rule{0.5\linewidth}{0.5pt}\end{center}

\section{2. Related Work}\label{related-work}

\section{2.1 Multilingual Benchmarks}\label{multilingual-benchmarks}

Most multilingual benchmarks evaluate whether a model can perform a task \emph{in} many languages, rather than whether it commands the languages themselves. The dominant pattern translates an English knowledge or reasoning benchmark into other languages: MMLU (Hendrycks et al. 2021) and its harder successor MMLU-Pro (Wang et al. 2024) are English knowledge tests, and MMLU-ProX (Xuan et al. 2025) extends this format to 29 languages. Because the underlying task is factual recall, a model can answer correctly without any sensitivity to the grammar of the target language. BenchMAX (Huang et al. 2025) broadens coverage to reasoning, code, and other capabilities across languages, but these are deliberately language-agnostic skills; MultiLoKo (Hupkes and Bogoychev 2025) tests local knowledge across 31 languages, again a knowledge rather than a proficiency probe. Belebele (Bandarkar et al. 2024) offers reading comprehension in 122 language variants but in multiple-choice form, which tests selection rather than generation. FLORES-200 (NLLB Team 2022) provides high-quality parallel data across 200 languages, but it is a corpus, not an evaluation protocol: it enables translation scoring without prescribing how meaning preservation should be judged.

The common gap is that none of these directly probe a model's grasp of language-specific grammar or its ability to preserve meaning under translation. A model can score well on translated knowledge tests while failing to detect a grammatical error that any native speaker would catch.

M-GATE is designed to fill this gap by measuring what we call linguistic proficiency: whether a model reliably detects violations of a language's grammar and can preserve meaning when working in it, rather than whether it can complete a familiar task that happens to be phrased in that language. The benchmark approaches this through two complementary tasks. In the first, the model is shown individual sentences and asked to judge whether each one is grammatically correct. These sentences are written by professional linguists and deliberately turn on grammatical phenomena that exist in the target language but not in English, so that answering correctly requires sensitivity to the language's own rules rather than world knowledge or surface fluency (§3.2). In the second, the model translates an English sentence into the target language and then back into English, and comparing the original with the result reveals where meaning has been lost or distorted along the way. Because that comparison is made between two English texts, the task needs neither human reference translations nor an evaluator more capable than the models being tested (§3.3). Finally, where the benchmarks surveyed above tend to report a single aggregate score across their languages, M-GATE reports results for each of its 30 languages separately. The set is chosen for typological diversity and deliberately includes low-resource languages such as Kinyarwanda and Dzongkha, which most existing benchmarks either fold into an average or leave out altogether.

\section{2.2 Translation Evaluation and LLM-as-Judge}\label{translation-evaluation-and-llm-as-judge}

Automatic translation metrics fall into two families, both ill-suited to our goal. Reference-overlap metrics, such as BLEU (Papineni et al. 2002) and chrF (Popović 2015), and learned metrics such as COMET (Rei et al. 2020) and BLEURT (Sellam et al. 2020) all require reference translations and correlate imperfectly with human judgment, particularly for high-quality output and for low-resource languages where references are scarce; they also penalize valid paraphrases (Freitag et al. 2020; Mathur et al. 2020). The LLM-as-judge paradigm (Zheng et al. 2023; Chiang et al. 2024; Li et al. 2023) removes the reference requirement, but judging translation quality directly requires a judge at least as competent as the tested model in the target language, which is precisely where judges are least reliable, especially for low-resource languages (Fu and Liu 2025), and where they exhibit self-preference (Panickssery et al. 2024).

Round-trip translation sidesteps both problems by reducing the judgment to an English-to-English comparison. Back-translation has a long history as an evaluation device (Brislin 1970) and in machine translation data augmentation (Sennrich et al. 2016), and its use as a quality estimator has been criticized on the grounds that a faithful round-trip does not guarantee a faithful intermediate translation (Somers 2005), and partially rehabilitated, with the quality of a round trip (translating out and back) shown to correlate with the quality of the corresponding one-way translation under human assessment (Aiken and Park 2010). Most closely related is concurrent work by Skorobogat et al. (2026), who likewise argue that translated knowledge benchmarks fail to measure multilingual proficiency and propose round-trip translation as the alternative, releasing the LiT benchmark.

The two benchmarks also differ in how round-trips are judged and validated, and in the surrounding task design. We defer a detailed comparison to the discussion (§6), after our benchmark and its human validation have been presented.

\section{2.3 Grammatical Acceptability and Error Detection}\label{grammatical-acceptability-and-error-detection}

A separate line of work evaluates grammatical knowledge through acceptability judgments and error detection. For English, CoLA (Warstadt et al. 2019) collects sentences from the linguistics literature labeled grammatical or ungrammatical, and BLiMP (Warstadt et al. 2020) uses automatically generated minimal pairs to isolate specific phenomena; together they established acceptability judgment as a meaningful probe of linguistic competence. The approach has since been extended beyond English. CLiMP (Xiang et al. 2021) ports the minimal-pairs method to Chinese, and MELA (Zhang et al. 2024), the largest multilingual acceptability benchmark to date, assembles roughly 46K judgments across ten languages from several families (Germanic, Romance, Slavic, Sino-Tibetan, Japonic, and Semitic). On the error-detection side, the MultiGED-2023 shared task (Volodina et al. 2023) covers binary grammatical error detection in five languages (Czech, English, German, Italian, and Swedish), with data drawn from second-language learner corpora.

Our grammar task adopts the same acceptability-judgment paradigm but differs along several dimensions. First, items are crafted by professional linguists to target phenomena that exist in the target language but not in English, rather than generated from templates (BLiMP, CLiMP) or drawn from existing corpora and learner data (CoLA, MELA, MultiGED) (§3.2). Second, items are adversarially selected to be difficult for current frontier models and cross-validated by linguists, rather than sampled to reflect a natural or learner-error distribution. Third, the task spans 30 typologically diverse languages with substantial low-resource representation, including Fijian, Kinyarwanda, and Dzongkha, well beyond the ten higher-resource languages of MELA or the five European languages of MultiGED. We are not aware of prior work that pairs adversarially selected, linguist-authored grammatical items with round-trip translation, or that extends grammatical acceptability and error detection to this range of low-resource languages.

\section{2.4 Data Contamination and Living Benchmarks}\label{data-contamination-and-living-benchmarks}

Static, publicly released benchmarks are increasingly vulnerable to data contamination: once test items appear on the web, they are liable to be absorbed into the pretraining corpora of later models, inflating scores without reflecting genuine capability (Sainz et al. 2023). Contamination can sometimes be detected after the fact (Oren et al. 2024), but it cannot be undone, and the risk compounds as a benchmark ages. Proposed mitigations include withholding or encrypting test data and demanding training-exclusion guarantees from providers (Jacovi et al. 2023), as well as continuously updated, human-in-the-loop leaderboards that are harder to contaminate (Chiang et al. 2024). Contamination is a live concern in our setting specifically: the concurrent LiT benchmark was constructed from new data expressly to avoid it (Skorobogat et al. 2026). M-GATE keeps its test set private behind a public leaderboard and releases only illustrative, non-live examples; we describe this design and its reproducibility trade-offs in §7.

\begin{center}\rule{0.5\linewidth}{0.5pt}\end{center}

\section{3. Benchmark Design}\label{benchmark-design}

\section{3.1 Language Selection}\label{language-selection}

M-GATE covers 30 languages (Table 1), deliberately spanning the full resource spectrum: from high-resource languages such as English, Spanish, and Chinese to low-resource languages such as Dzongkha, Fijian, and Kinyarwanda (Joshi et al. 2020). This design serves two purposes. First, it lets us test whether linguistic proficiency tracks the availability of training data (Blasi et al. 2022). Second, because the benchmark is run continuously, it lets us track whether the gap between high- and low-resource languages narrows as new models are released. This is a question that existing benchmarks, concentrated on high-resource languages, cannot answer.

A language was included only if it met the following criteria:

\begin{itemize}
\tightlist
\item
  A clearly defined standard variety that is widely accepted and not actively contested.
\item
  A prescriptive authority or institution (language academy, government body, or equivalent) that codifies grammar, spelling, and usage.
\item
  Settled orthography with no competing spelling systems or unresolved script debates.
\item
  No significant diglossia; the written standard and spoken form should not be so divergent that grammaticality judgments depend on register.
\item
  Low dialect fragmentation relative to the standard; regional variation is fine, but the standard should be broadly recognized across the speech community.
\item
  Grammaticality questions should be non-disputable; educated native speakers should broadly agree on whether a sentence is grammatical in the standard language.
\end{itemize}

These criteria reduce ambiguity in the gold standard: when a language has a single, broadly accepted written standard, grammaticality labels are less likely to be confounded by unresolved disagreement about correct usage. We do not claim that any standard is wholly uncontested. Rather, the benchmark deliberately targets standardized languages across the resource spectrum, meaning languages that have a codified written norm with an institutional or widely recognized basis, whether high- or low-resource. This is a selection choice, and we acknowledge that it introduces bias: deciding which languages have a settled enough standard is ultimately our own judgment, and in places a politically sensitive one. The criteria favor institutionally supported languages while under-representing or excluding those with contested or fragmented standards. The diglossia criterion concerns the written standard rather than spoken usage; we include Arabic, for example, because Modern Standard Arabic has codified grammaticality norms even though it diverges from spoken varieties (Ferguson 1959). Conversely, the criteria excluded languages such as Kurdish, which has two competing standards (Kurmanji and Sorani), two writing systems (Latin and Arabic), and no unified prescriptive authority. Inclusion was assessed using Ethnologue classifications (Eberhard et al. 2024), the Expanded Graded Intergenerational Disruption Scale (EGIDS; Lewis and Simons (2010)), and Digital Language Support levels (Simons et al. 2022), together with our own review.

Quechua (Cusco-Collao) and Guarani sit closest to the boundary of these criteria, with enough dialectal variation that some grammaticality judgments may be contested; we relied on the variety-specific standards and our linguists' cross-validation.

Within these constraints, we selected for typological breadth: the 30 languages span 14 language families plus one isolate (Basque), 13 writing systems, and the major morphological types (fusional, agglutinative, and isolating). They also span more than five orders of magnitude in estimated Common Crawl share (41\% for English to 0.0001\% for Fijian and Dzongkha), our proxy for representation in common pretraining corpora.

We group the 30 languages into three resource tiers of ten (Table 1), used throughout for sampling and reporting. Tiers combine three signals: Ethnologue's Digital Language Support level (Simons et al. 2022) and EGIDS (Lewis and Simons 2010) for digital and institutional support, and Common Crawl share for pretraining-data presence. Tier 1 comprises languages with top-level digital support that receive dedicated post-training and evaluation from every major frontier lab; Tier 2 comprises languages with strong institutional support and meaningful pretraining presence but without that dedicated investment; Tier 3 comprises languages whose pretraining footprint is weak across signals. No single axis determines membership, and the boundaries involve judgment: Korean and Arabic remain in Tier 1 despite lower Common Crawl share than Czech or Indonesian because of dedicated frontier-lab investment, while Tagalog sits in Tier 3 despite roughly 80 million speakers because the Philippine internet is largely English-dominant, so models likely see less Tagalog text than its speaker count suggests. The Tier 2/3 boundary (Khmer, Kyrgyz, Tagalog, within 1.5× of one another on Common Crawl share) is the most sensitive to these judgments, and reasonable alternative cuts exist. The full tiering methodology and per-language rationale are given in Appendix A.

{\setlength{\tabcolsep}{4pt}\def\LTcaptype{none} 
\begin{longtable}[]{@{}
  >{\raggedright\arraybackslash}p{(\linewidth - 16\tabcolsep) * \real{0.0500}}
  >{\raggedright\arraybackslash}p{(\linewidth - 16\tabcolsep) * \real{0.0900}}
  >{\raggedright\arraybackslash}p{(\linewidth - 16\tabcolsep) * \real{0.1550}}
  >{\raggedright\arraybackslash}p{(\linewidth - 16\tabcolsep) * \real{0.1125}}
  >{\raggedright\arraybackslash}p{(\linewidth - 16\tabcolsep) * \real{0.0850}}
  >{\raggedright\arraybackslash}p{(\linewidth - 16\tabcolsep) * \real{0.1100}}
  >{\raggedright\arraybackslash}p{(\linewidth - 16\tabcolsep) * \real{0.1125}}
  >{\raggedright\arraybackslash}p{(\linewidth - 16\tabcolsep) * \real{0.1750}}
  >{\raggedright\arraybackslash}p{(\linewidth - 16\tabcolsep) * \real{0.1100}}@{}}
\toprule\noalign{}
\begin{minipage}[b]{\linewidth}\raggedright
Tier
\end{minipage} & \begin{minipage}[b]{\linewidth}\raggedright
Code
\end{minipage} & \begin{minipage}[b]{\linewidth}\raggedright
Language
\end{minipage} & \begin{minipage}[b]{\linewidth}\raggedright
L1+L2 (M)
\end{minipage} & \begin{minipage}[b]{\linewidth}\raggedright
EGIDS
\end{minipage} & \begin{minipage}[b]{\linewidth}\raggedright
DLS
\end{minipage} & \begin{minipage}[b]{\linewidth}\raggedright
CC share (\%)
\end{minipage} & \begin{minipage}[b]{\linewidth}\raggedright
Language Family
\end{minipage} & \begin{minipage}[b]{\linewidth}\raggedright
Script
\end{minipage} \\
\midrule\noalign{}
\endhead
\bottomrule\noalign{}
\endlastfoot
1 & en-US & English (US) & 1500+ & 0 & Thriving & 41.06 & Indo-European (Germanic) & Latin \\
1 & de & German & 135 & 1 & Thriving & 5.98 & Indo-European (Germanic) & Latin \\
1 & ja & Japanese & 125 & 1 & Thriving & 5.72 & Japonic & Japanese (mixed) \\
1 & zh-CN & Simplified Chinese & 1100 & 0 & Thriving & 4.99 & Sino-Tibetan (Sinitic) & Han (Simplified) \\
1 & es & Spanish & 560 & 0 & Thriving & 4.66 & Indo-European (Romance) & Latin \\
1 & fr & French & 310 & 0 & Thriving & 4.61 & Indo-European (Romance) & Latin \\
1 & pt-BR & Portuguese (Brazil) & 215 & 1 & Thriving & 2.53 & Indo-European (Romance) & Latin \\
1 & pl & Polish & 40 & 1 & Vital & 2.04 & Indo-European (Slavic) & Latin \\
1 & ko & Korean & 80 & 1 & Thriving & 0.84 & Koreanic & Hangul \\
1 & ar & Arabic\footnote{The operational tag targets Modern Standard Arabic. Ethnologue distinguishes the Arabic macrolanguage (\texttt{ara}) from Standard Arabic (\texttt{arb}); this code distinction does not affect the Tier 1 placement.} & 370 & 0 & Thriving & 0.65 & Afro-Asiatic (Semitic) & Arabic \\
2 & cs & Czech & 10 & 1 & Vital & 1.15 & Indo-European (Slavic) & Latin \\
2 & id & Indonesian & 200 & 1 & Vital & 1.10 & Austronesian (Malayo-Polynesian) & Latin \\
2 & fi & Finnish & 5.5 & 1 & Vital & 0.37 & Uralic (Finnic) & Latin \\
2 & th & Thai & 70 & 1 & Vital & 0.37 & Kra-Dai (Tai) & Thai \\
2 & ne & Nepali & 32 & 1 & Vital & 0.053 & Indo-European (Indo-Aryan) & Devanagari \\
2 & sq & Albanian\footnote{Albanian is spoken by around 6--7 million people (Coretta et al. 2023), rather than a combined Ethnologue total. Ethnologue lists separate Gheg (\texttt{aln}) and Tosk (\texttt{als}) entries.} & 6--7 & 1 & Vital & 0.053 & Indo-European (Albanian) & Latin \\
2 & ta & Tamil\footnote{EGIDS 2 reflects the targeted variety (India, where Tamil has state-level status); Tamil is EGIDS 1 in Sri Lanka and Singapore.} & 85 & 2 & Vital & 0.049 & Dravidian & Tamil \\
2 & eu & Basque & 0.75 & 2 & Vital & 0.039 & Isolate & Latin \\
2 & is & Icelandic & 0.37 & 1 & Vital & 0.038 & Indo-European (Germanic) & Latin \\
2 & km & Khmer & 17 & 1 & Vital & 0.009 & Austroasiatic (Khmeric) & Khmer \\
3 & ky & Kyrgyz & 5 & 1 & Vital & 0.013 & Turkic (Kipchak) & Cyrillic \\
3 & tl & Tagalog\footnote{Ethnologue distinguishes Tagalog (\texttt{tgl}) from Filipino (\texttt{fil}); published DLS values differ between them. The Tier 3 placement follows Tagalog's low Common Crawl share and the English-dominant Philippine web, so an Ascending classification for \texttt{tgl} would reinforce rather than weaken it.} & 80 & 1 & Vital & 0.010 & Austronesian (Malayo-Polynesian) & Latin \\
3 & as & Assamese\footnote{The 23.6 million speaker total is the Ethnologue 29th-edition L1+L2 figure, included as a footnoted exception to the fixed 27th-edition snapshot.} & 23.6 & 2 & Vital & 0.004 & Indo-European (Indo-Aryan) & Bengali--Assamese (Eastern Nagari) \\
3 & mt & Maltese & 0.52 & 1 & Vital & 0.003 & Afro-Asiatic (Semitic) & Latin \\
3 & rw & Kinyarwanda & 13 & 1 & Vital & 0.003 & Niger-Congo (Bantu) & Latin \\
3 & ti-ET & Tigrinya\footnote{EGIDS 2 reflects the Ethiopian variety (ti-ET); Tigrinya is EGIDS 1 (National) in Eritrea.} & 10 & 2 & Vital & 0.001 & Afro-Asiatic (Semitic) & Ge'ez (Ethiopic) \\
3 & qu & Quechua (Cusco-Collao)\footnote{Ethnologue (27th ed.) reports only a population band (10K--1M) for Cusco-Collao Quechua. Peru's 2017 census counted 3.8M mother-tongue speakers of Quechua across all varieties (Instituto Nacional de Estadística e Informática (INEI) 2017); the Quechuan macrolanguage total is \textasciitilde7M. Ethnologue's variety codes, including Cusco Quechua (\texttt{quz}) and the separately scored Ayacucho Quechua (\texttt{qve}), are not interchangeable.} & \textless1 & 2 & Vital & 0.001 & Quechuan & Latin \\
3 & gn & Guarani (Paraguayan) & 6 & 1 & Vital & 0.001 & Tupian (Tupí--Guaraní) & Latin \\
3 & fj & Fijian & 0.35 & 1 & Vital & 0.0001 & Austronesian (Malayo-Polynesian) & Latin \\
3 & dz & Dzongkha & 0.17 & 1 & Vital & 0.0001 & Sino-Tibetan (Tibetic) & Tibetan \\
\end{longtable}
}

\textbf{Table 1:} \emph{The 30 benchmark languages, grouped into three resource tiers (§3.1). \emph{Code} is the benchmark's operational language tag (BCP-47; region subtags indicate the targeted variety). \emph{L1+L2}: speakers in millions, Ethnologue 27th ed., rounded (fixed at that edition for reproducibility; see Appendix A.4). Exceptions are footnoted. \emph{EGIDS}: Expanded Graded Intergenerational Disruption Scale (Lewis and Simons 2010); 0 = International, 1 = National, 2 = Provincial. \emph{DLS}: Digital Language Support level (Simons et al. 2022). \emph{CC share}: percentage of pages in Common Crawl CC-MAIN-2026-12 identified as the language by CLD2; shares for non-Latin-script languages (e.g., Khmer, Tigrinya, Assamese, Dzongkha) are likely undercounted due to language-identification bias. Within each tier, rows are ordered by CC share.}

\section{3.2 Task 1: Grammatical Error Detection}\label{task-1-grammatical-error-detection}

The first task probes a core diagnostic of linguistic competence: whether a model can detect that a sentence is ungrammatical, framed as a binary judgment (error / no error) (Warstadt et al. 2019, 2020). Unlike error correction or explanation, binary detection isolates recognition of ungrammaticality from the ability to localize or repair it (§3.2.6). For each language we curated 100 ``stumper'' sentences targeting phenomena that exist in the target language but not in English; for example, Czech verbal case government, Basque ergative--absolutive case marking with transitivity-conditioned auxiliary selection, and Finnish illative case formation. Each item set is unique to its language and was constructed and validated in-language by professional linguists (§3.2.1).

Table 2 shows illustrative items from the non-live pool (§7); none is part of the scored benchmark. In each case the error is invisible in an English gloss, which is the point: judging these sentences requires competence in the language itself.

{\def\LTcaptype{none} 
\begin{longtable}[]{@{}
  >{\raggedright\arraybackslash}p{(\linewidth - 8\tabcolsep) * \real{0.1111}}
  >{\raggedright\arraybackslash}p{(\linewidth - 8\tabcolsep) * \real{0.2222}}
  >{\raggedright\arraybackslash}p{(\linewidth - 8\tabcolsep) * \real{0.2222}}
  >{\raggedright\arraybackslash}p{(\linewidth - 8\tabcolsep) * \real{0.2963}}
  >{\raggedright\arraybackslash}p{(\linewidth - 8\tabcolsep) * \real{0.1481}}@{}}
\toprule\noalign{}
\begin{minipage}[b]{\linewidth}\raggedright
Language (Tier)
\end{minipage} & \begin{minipage}[b]{\linewidth}\raggedright
Sentence
\end{minipage} & \begin{minipage}[b]{\linewidth}\raggedright
English gloss
\end{minipage} & \begin{minipage}[b]{\linewidth}\raggedright
Error and correction
\end{minipage} & \begin{minipage}[b]{\linewidth}\raggedright
Panel models tricked
\end{minipage} \\
\midrule\noalign{}
\endhead
\bottomrule\noalign{}
\endlastfoot
Finnish (T2) & Meidän piti mennä Thaimaaseen. & We were supposed to go to Thailand. & Illative case formation: the correct form is \emph{Thaimaahan}. & 3 of 5 \\
Czech (T2) & Musíme se vyvarovat těmto zbytečným chybám v budoucnu. & We must avoid these unnecessary mistakes in the future. & Verbal case government: \emph{vyvarovat se} requires the genitive (\emph{těchto chyb}); the dative here is contaminated from the near-synonym \emph{vyhnout se}. & 3 of 5 \\
Khmer (T2) & {\khmerfont ខ្ញុំទិញខោអាវពីរក្បាល។} & I bought two sets of clothes. & Classifier selection: {\khmerfont ក្បាល} is not the correct classifier for clothing; it should be {\khmerfont ឈុត (ខ្ញុំទិញខោអាវពីរឈុត។)}. & 4 of 5 \\
Fijian (T3) & Keitou sa kana na ika ka siwa e na mataka. & We ate the fish that was caught in the morning. & Verb transitivity: the intransitive \emph{kana} (to eat) is used where the transitive \emph{kania} (to eat something) is required. & 3 of 5 \\
\end{longtable}
}

\textbf{Table 2:} \emph{Illustrative grammar items from the non-live pool, one item each from four languages, with the linguists' error explanations (abridged). Trick counts are against the five-model construction panel (§3.2.1). The pool also contains fully grammatical sentences that draw false alarms: 35\% of valid non-live error-free items were wrongly flagged by two or more panel models.}

The grammar task covers all 30 benchmark languages, including English; only the translation task excludes English, which serves there as the source language and cannot be a target (§3.3.1). The requirement that items turn on phenomena absent from English exists to ensure that judging a sentence demands competence in the target language itself rather than knowledge transferable from English; for the English item set this constraint is vacuous, and items instead target difficult phenomena of standard written American English, such as the mandative subjunctive (\emph{it is important that he go}, including the subjunctive required after \emph{lest}) and the counterfactual subjunctive (\emph{were}, not \emph{was}), alongside collective-noun and correlative-subject agreement and dangling modifiers. In every other respect the English items were constructed, validated, and selected through the same pipeline (§3.2.1--3.2.3).

\subsection{3.2.1 Stumper dataset creation}\label{stumper-dataset-creation}

The stumper dataset was created by three professional linguists per language. All were native or near-native speakers of their assigned language, with formal linguistics training and at least five years of professional experience in translation or language work (most had 10+ years). Each linguist was asked to author ``stumper'' sentences: sentences designed to trick large language models into misjudging grammaticality, either by wrongly flagging a correct sentence as containing an error or by wrongly accepting an incorrect one. Because the final dataset is balanced (§3.2), we asked for roughly equal numbers of each type, with a target of about 50 sentences per linguist (25 with errors, 25 without) and at least 150 per language in total, since we expected some to be eliminated in later steps.

While authoring, linguists worked with a web app that ran each candidate sentence they came up with through the live grammar-task pipeline and reported which models it tricked, that is, which produced the wrong binary judgment: a false positive on a grammatical sentence, or a false negative on an ungrammatical one. A sentence qualified only if it tricked at least two models from a fixed five-model panel: Gemini 3 Flash, GPT-4o, Llama 4 Maverick, Grok 4.1 Fast, and DeepSeek V3.1 (exact versions in Appendix B, Table B.2). The testing app evaluated each candidate once per model at temperature 0 (see §3.5.2 for further discussion), so the panel's judgments during collection were near-deterministic; a sentence's ``tricks \emph{N} models'' status was therefore stable rather than a single draw from a noisy distribution.

We deliberately excluded frontier/flagship models from this construction panel. Selecting items against the strongest models risks overfitting the dataset to those specific models' idiosyncratic weaknesses; because the same model families headline the evaluation leaderboard, this would inflate apparent difficulty. The five models above are capable but non-flagship, span multiple providers, and include at least one non-US lab. Requiring a sentence to trick at least two of the five rather than one selects for items whose difficulty generalizes across independent models rather than reflecting a single model's quirk. We further asked linguists to include at least five sentences that also trick Gemini 3 Flash, which our internal testing found the strongest of the five at multilingual tasks; this is single-model selection in miniature, which we accepted because the two-of-five requirement still enforces cross-model generality and because items that defeat the strongest of the five proved the most likely to defeat the others. The construction-panel models themselves appear on the leaderboard, but their grammar scores should be read with this selection bias in mind: the items were chosen expressly to defeat them, and they occupy the bottom of the grammar range accordingly (§5.2).

A leave-one-family-out analysis (Appendix D.4) quantifies this bias and its reach: re-scoring all configurations on the subset of items whose qualification never consulted a given panel model's family, the panel models' families gain relative to the rest of the field (Meta +0.11, DeepSeek +0.07, xAI +0.07 MCC), and the effect persists when the panel member itself is excluded: Grok 4.20 and 4.3 inherit the depression from Grok 4.1 Fast's panel membership despite never being consulted. By contrast, the families that headline the grammar leaderboard show no material effect (Google +0.02, OpenAI −0.01), so the strongest models' scores (high or low) are not artifacts of the panel's composition.

The instructions constrained what counts as a valid item. Each error had to be indisputable according to the language's prescriptive authorities; stylistic issues, contested usages, forms that authorities have come to accept over time, and rules stemming from very recent reforms (first half of 2025 and later) were all excluded. For sentences with errors, linguists explained the error in an English comment. To ensure variety, any specific error instance (a particular word form or construction) could appear in at most two of a linguist's sentences. We did not prescribe error categories, since these are language-dependent, and instead directed linguists toward common native-speaker and learner mistakes. To limit contamination, linguists were asked not to use sentences found online verbatim but to adapt or paraphrase them; they were permitted, though not required, to use LLMs to brainstorm candidate errors provided every item was verified against authoritative sources, and in a post-collection survey most reported not using LLM assistance. Compensation covered qualifying sentences only; we return to the incentive this creates, and the validation step that checks it, in §3.2.2. The complete instructions are reproduced in Appendix C.

\subsection{3.2.2 Stumper dataset linguist validation}\label{stumper-dataset-linguist-validation}

Our aim was a high-quality dataset that avoids disputes arising from regional variants, subjective interpretation of grammatical rules, or other factors that could make an item's status contestable. In the second step, the three linguists for each language cross-validated the candidate sentences: each item, authored by one linguist, was reviewed by the other two. Reviewers saw the sentence, its category label (error / no error), and the author's error explanation, along with the pseudonymized author identity; they did not see which models the item had tricked, so validation judged linguistic validity independently of difficulty. A reviewer could invalidate an item for any defect in the sentence, label, or explanation, including duplicates and near-duplicates. Review took place in a shared environment in which a reviewer could see the other reviewer's response: the intent was to surface disagreements and converge on a defensible label rather than to measure independent agreement. In practice the two reviewers agreed in the large majority of cases; rather than litigate invalidated items toward consensus, we discarded them and collected replacements. A single invalidation excluded an item from the candidate pool. Across all languages, approximately 83\% of candidate sentences passed validation. Because compensation was tied to a sentence qualifying against the model panel, linguists had an incentive to submit marginal items; the independent cross-validation described here is the check on that incentive, and it removed roughly one in six candidates.

Because false positives (flagging a fully grammatical sentence as erroneous) were the harder category to collect sentences for in most languages, several languages (Tamil, Korean, and Arabic) fell short of the required 50 error-free and 50 error candidates after validation. In those cases we asked all three linguists to contribute ten additional sentences in both categories and repeated the validation, so top-up items met the same bar as the original pool.

\subsection{3.2.3 Final stumper dataset selection}\label{final-stumper-dataset-selection}

After validation, each language had roughly 110--140 valid candidates. To produce a fully balanced 100-item set (50 with errors, 50 without) per language, we selected items as follows. Within each category, we first ranked candidates by the number of the five testing models they tricked (§3.2.1) and took the hardest items; we then adjusted the selection at the margin to minimize the difference in mean sentence length between the two categories, breaking remaining ties with two held-out frontier models (Gemini 3.1 Pro Preview and GPT-5.4),\footnote{These two frontier models influence only tie-breaking among the already-selected hardest items, so their effect on the final set is marginal. We note their role here when interpreting their leaderboard results.} preferring the items these models more often got wrong.

We select the hardest available items deliberately, to keep the strongest models from clustering near the top of the scale and to preserve separation among them. As a consequence, scores reflect performance on each language's most difficult items rather than a fixed difficulty bar shared across languages (see §3.2.7).

Dataset artifacts such as length can act as shortcuts for predicting labels (Gururangan et al. 2018; Jiang et al. 2022), so we balance sentence length across the two categories: pooled across all items, the median sentence length is 35 characters for items with errors and 39 for items without, a 4-character difference. Within individual languages the residual correlation between length and label is larger in a few cases, but this does not threaten the scores: models are evaluated zero-shot on independent single-sentence requests, so they cannot learn a dataset-level correlation between length and label the way a supervised model trained on biased data could. The only residual pathway is a pre-existing, dataset-independent length prior. We tested for this directly in the four languages where length is most correlated with the label (Assamese, Fijian, Quechua, and Kinyarwanda; item-level length--label agreement of 72--90\%): for each model we split items by whether a median-length heuristic agreed with the gold label and compared accuracy. Across the models we tested, the aggregate gaps ranged from −0.096 to +0.056, with the largest values negative. No model scored systematically higher on items where the length heuristic agreed with the gold label. We conclude that the length--label correlation present in some languages' items is not exploited under zero-shot evaluation. Per-language length distributions and the full per-model results are reported in Appendix D.

\subsection{3.2.4 Grammar task evaluation}\label{grammar-task-evaluation}

Models are evaluated on the grammar task on a single-sentence basis: our custom Python app iterates over individual stumper sentences and inserts them, together with the language name, into the following prompt:

\begin{Shaded}
\begin{Highlighting}[]
\NormalTok{Reply with exactly one word: Yes or No.}
\NormalTok{Are there any grammar or spelling issues with the following \{language\} sentence?}
\NormalTok{\{stumper\_sentence\}}
\end{Highlighting}
\end{Shaded}

The \texttt{\{language\}} field is filled with the full English language name, with a country or region qualifier where needed to disambiguate the targeted variety (for example, ``Tigrinya (Ethiopia)''). Each sentence is sent as an independent, zero-shot request with no system prompt or in-context examples. We evaluate each model under two temperature settings: temperature 0 (one run) and the model's default temperature (three runs) to capture run-to-run variability. We send no other generation parameters, except for reasoning-effort levels where these are part of the model identifier (for example, the -high, -adaptive, or -medium suffixes). The full temperature and reporting policy, including how the three default-temperature runs are aggregated (Avg 3-Run / Worst 3-Run), is described in §3.5.2.

We prompt for ``grammar or spelling issues'' rather than grammar alone because the two can overlap: a given error may be construed as either a typo or a grammatical mistake, and we want models to flag the error regardless of how they categorize it rather than withhold a ``Yes'' because they judge it a spelling issue. In practice this is a precaution only, since linguists were instructed to construct grammatical errors; spelling errors were not solicited (§3.2.1).

We frame the instruction in English, rather than the target language, for two reasons. First, although prompting in the target language is often considered best practice, in deployed multilingual systems the framing prompt is frequently in English while only the content is in the target language; our setup mirrors that. Second, English framing makes the model more likely to answer in English, which simplifies and disambiguates parsing. In our testing, instructing the model in the target language to produce an English answer increased parsing failures.

\subsubsection{3.2.4.1 Grammar task response parsing}\label{grammar-task-response-parsing}

When the model response is received, the parser strips leading and trailing whitespace and lowercases the text. It then checks whether the response begins with ``yes'' or ``no'', optionally followed by a space or punctuation. When this fails, we trigger a fallback: we send the response to Claude Haiku 4.5 and ask it to classify the answer using the following prompt:

\begin{Shaded}
\begin{Highlighting}[]
\NormalTok{The answer below may be in any language. Determine whether it means yes or no. Reply in English with exactly one word: Yes or No.}

\NormalTok{Answer:}
\NormalTok{"\{text\}"}
\end{Highlighting}
\end{Shaded}

The fallback model sees only the answer text; it is blind to which model produced the response and never judges grammatical correctness, it only normalizes an already-produced answer into a yes/no label. The fallback is triggered for fewer than 2\% of responses. To check its reliability, the authors manually reviewed approximately 50 fallback cases per language; the intended yes/no meaning was confirmed by a native-speaker linguist where available, or otherwise via machine translation. We found no miscategorizations in this sample. Given the low fallback rate and the absence of errors in the reviewed cases, we did not pursue further changes to the fallback.

In fewer than 0.03\% of responses, the fallback model could not classify the message at all. Of these unparseable responses, most came from small models (Gemma 4 31B, 47\%; ByteDance Seed 2.0 Lite and GLM 5.2, 16\% each) and low-resource languages (Basque, 34\%; Icelandic, Albanian, and Fijian, roughly 9\% each). In every such case the model's response was empty, leaving nothing to classify. We score these as incorrect: an empty response to an item with an error counts as a false negative, and an empty response to an item without an error counts as a false positive.

\subsection{3.2.5 Grammar score calculation}\label{grammar-score-calculation}

Each grammar check produces a binary Yes/No output, and the dataset is balanced 50/50: 50 sentences with errors (expecting ``Yes'') and 50 without (expecting ``No''). With this balance, a model that always returns the same answer, regardless of the sentence, scores exactly 50\% accuracy and an MCC of 0 (by the convention below), so neither metric rewards such a strategy.

We score the grammar task primarily with the Matthews correlation coefficient (MCC) (Matthews 1975), a chance-corrected measure computed from the four confusion-matrix counts, that is true positives (TP), true negatives (TN), false positives (FP), and false negatives (FN):

\[
\mathrm{MCC} = \frac{TP \times TN - FP \times FN}{\sqrt{(TP+FP)\,(TP+FN)\,(TN+FP)\,(TN+FN)}}
\]

MCC ranges from −1 to +1, where +1 is perfect prediction, 0 is no better than chance, and −1 means the predictions are the exact reverse of the gold labels. Unlike accuracy or F1, MCC yields a high value only when the classifier performs well across all four confusion-matrix categories, which makes it more informative and reliable for binary classification (Chicco and Jurman 2020). When a model returns the same label for every item, one or more denominator terms are zero and MCC is undefined; following common convention, we set MCC to 0 in these cases.

We compute MCC and F1 for each temperature setting separately (temperature 0, and the Avg 3-Run and Worst 3-Run aggregates at default temperature as per §3.5) and report them side by side; the public leaderboard reports both metrics.

\subsection{3.2.6 Remarks on grammar task design}\label{remarks-on-grammar-task-design}

We deliberately chose an easy-to-interpret task: a binary judgment over a balanced dataset. This design has a known limitation. The task records whether a model correctly flags a sentence as containing an error, but not why: it cannot tell whether the model detected the specific error the linguist constructed, detected a different (possibly spurious) issue, or reached the correct label by chance. Binary detection therefore measures whether a model flags ungrammaticality, not how it reaches that judgment.

Recovering that finer signal would require either human linguists to read every response (which is infeasible at the scale of 100 items across 30 languages, 80+ model configurations, and multiple runs) or a highly capable multilingual LLM judge to assess each model's stated reasoning. We avoid the latter because LLM judges are documented to be unreliable in exactly our setting: a judge's judgments become inconsistent across languages, degrading sharply for low-resource languages even in multilingual-specialized models (Fu and Liu 2025). This is a different setup from our round-trip translation task (§3.3), where the judge compares English to English, a task on which we validate judge--human agreement directly (§4), rather than reasoning about grammar in a low-resource target language.

A further limitation concerns what a prompted judgment represents. Asking a model whether a sentence is grammatical tests its ability to \emph{report} a judgment, which can diverge from what its internal representations encode: metalinguistic yes/no prompting has been shown to undersell models' linguistic knowledge, so a wrong answer is evidence about prompted behavior, not proof that the underlying knowledge is absent (Hu and Levy 2023). Our task therefore also folds in instruction-following and, because we prompt in English about a non-English sentence (§3.2.4), cross-lingual instruction-following. We treat this as a feature rather than a confound, for two reasons. First, it mirrors how multilingual systems are prompted in deployment. Second, the main alternative, direct probability measurement, carries its own multilingual confound: in a preliminary Mandarin experiment, Hu and Levy find that direct scoring underperformed metalinguistic prompts on their sentence-comparison task (accuracy roughly 0.6 against 0.8), which they attribute to English-optimized models scoring context-free Chinese sentences poorly rather than to a genuine advantage of prompting; their word-prediction results in Chinese matched the English pattern. Neither method is a clean window onto latent competence in the multilingual setting. We therefore interpret grammar results as measuring prompted, deployment-relevant error detection rather than latent linguistic competence.

\subsection{3.2.7 Cross-language comparability}\label{cross-language-comparability}

Grammar-task scores are designed to be compared \emph{within} a language, not \emph{across} languages. Each language has its own 100 items, targeting language-specific phenomena and selected to be the hardest available for that language (§3.2.3) rather than calibrated to a shared difficulty level. There is no language-independent scale of grammatical difficulty against which items could be equated, so a lower score in one language than another need not mean the language is ``harder'', only that its items are different.

This non-comparability is compounded by genuine differences in model proficiency across languages. Because English and other high-resource languages dominate pretraining corpora (Joshi et al. 2020), models exhibit systematically uneven linguistic proficiency across languages (Blasi et al. 2022). This unevenness is not static: overall scores trend upward as newer models are released, but the progression is not linear and providers improve at different rates (§5.4). The effect is most pronounced for low-resource languages, where some labs exclude a language's data entirely while others deliberately invest in it.

We therefore urge users to read grammar results within a language. Valid uses include identifying which model scores best in a given language, or tracking how a single language's scores change across model releases; for example, following Nepali performance as new models are added (Figure 5, §5.4.4). Cross-language conclusions are not supported: statements such as ``language X is twice as hard as language Y'' or ``model Z understands language X three times better than language Y'' cannot be drawn from the grammar task and could misinform deployment decisions in multilingual settings. Cross-language comparison is instead supported by the round-trip translation task, which uses the same source sentences for every language (§3.3).

\section{3.3 Task 2: Round-trip Translation}\label{task-2-round-trip-translation}

The second task in our benchmark is round-trip translation: the tested model translates a single English sentence into the target language, then, in a separate request with no context or additional cues, translates it back into English. This design sidesteps a central obstacle in evaluating LLMs for translation. LLM-as-judge is unreliable when assessing multilingual output directly: a judge's decisions are inconsistent across languages, most severely for low-resource languages (Fu and Liu 2025). The common alternatives are also unsatisfactory. Human evaluation cannot keep pace with a continuously maintained leaderboard scoring 100 items across 29 languages for every new configuration, and it depends on annotators whose availability and standards shift over time, so scores from different periods would not be comparable. Surface-overlap and edit-distance metrics correlate poorly with human judgment for high-quality output and unduly penalize valid rewrites and paraphrases (Freitag et al. 2020; Mathur et al. 2020). Round-trip translation converts the problem into an English-to-English comparison between the original sentence and its round-tripped version, a judgment that recent models make reliably (Zheng et al. 2023), in contrast to their unreliability in low-resource languages. We validate this choice by comparing our LLM judges against three human annotators (§4); we discuss the limitations of round-trip evaluation, including its conflation of forward and backward translation quality, in §3.3.6.

The same 100 English source sentences, chosen for linguistic phenomena that often do not survive translation, are used to assess all 29 target languages (English is excluded as a target). Note that these are different from the sentences used in the grammar task. Because every language is evaluated on identical source sentences, translation scores are directly comparable across languages, unlike the grammar task, which uses neither a shared nor a calibrated item set (§3.2.7).

\subsection{3.3.1 Source dataset creation}\label{source-dataset-creation}

The translation task uses a single set of 100 English source sentences, applied unchanged to all 29 languages (English is not assessed in this task) so that scores are comparable across languages (§3.3.7). Items are kept short: most (76\%) fall in the 6--20 word range (median 12 words), with a tail of longer items running up to 36 words. A minority (12 of 100) consist of two or more sentences; these short passages carry phenomena that need more than one sentence to arise, such as cross-sentence pronoun reference (§3.3.2). For brevity we refer to all source items as ``sentences'' throughout. Every item targets linguistic phenomena known to cause meaning loss, distortion, or ambiguity under translation.

The corpus was built through a human-in-the-loop process. Candidate sentences were generated with two large language models (Gemini 3 Pro and Claude Opus 4.5) prompted to target specific phenomena, then every candidate underwent expert review, revision and rewrites; the categories themselves were refined over development into the final taxonomy described in §3.3.2.\footnote{The category taxonomy was revised during development; we describe only the final scheme (§3.3.2).} We first assembled 50 sentences and subsequently expanded to 100 for a larger sample. Each item was reviewed for grammaticality, for semantic coherence (that any intended ambiguity or polysemy is naturally interpretable rather than forced), and for the accuracy of its annotation. Review materially changed the corpus: it caught and replaced sentences that were syntactically ill-formed despite surface plausibility, sentences that forced incompatible word senses together without coherence, and annotations that mischaracterized the linguistic challenge; it also flagged items whose intended reading was insufficiently cued.

For each non-control item we additionally authored a target meaning: an explicit statement of the propositional content and pragmatic features that must survive the round trip. The target meaning defines the scoring anchors used by the judges: full preservation (5), substantive but partial preservation (3), or meaning absent or reversed (1), and is paraphrase-tolerant by design. A round trip that restates the meaning in plain language (for instance, rendering an idiom as its plain-meaning equivalent, ``raining heavily'' for ``raining cats and dogs'') scores 5, because the meaning is intact even though the wording differs. For the hardest items, we also wrote sample back-translations scored 5, 3, and 1, giving judges worked examples to calibrate against (§3.3.4). All sentences use Standard American English.

\subsection{3.3.2 Error categories}\label{error-categories}

The 100 source sentences span seven categories of translation-challenging phenomena (Table 3). Category membership is editorial: each item is assigned to the principal failure mode it probes, not to a surface feature of the source text. Six categories target distinct ways meaning is lost in round-trip translation; a seventh (control) provides simple, unambiguous baseline items for floor-effect detection and pipeline calibration.

{\def\LTcaptype{none} 
\begin{longtable}[]{@{}
  >{\raggedright\arraybackslash}p{(\linewidth - 4\tabcolsep) * \real{0.2941}}
  >{\raggedright\arraybackslash}p{(\linewidth - 4\tabcolsep) * \real{0.0882}}
  >{\raggedright\arraybackslash}p{(\linewidth - 4\tabcolsep) * \real{0.6176}}@{}}
\toprule\noalign{}
\begin{minipage}[b]{\linewidth}\raggedright
Category
\end{minipage} & \begin{minipage}[b]{\linewidth}\raggedright
n
\end{minipage} & \begin{minipage}[b]{\linewidth}\raggedright
Failure mode probed
\end{minipage} \\
\midrule\noalign{}
\endhead
\bottomrule\noalign{}
\endlastfoot
Discourse pragmatics & 26 & Pragmatic flattening: indirect speech, irony, hedging, attribution, or pragmatically resolved reference collapses into a flat or misattributed reading \\
Non-compositional & 15 & Literal rendering: an idiom or fixed expression is translated word-for-word and loses its figurative meaning \\
Implicit content & 14 & Unstated meaning (allusion, cultural or institutional shorthand) is dropped or made culturally inappropriate \\
Structural complexity & 14 & Structural simplification: nested clauses flatten, scope collapses, or meaning-bearing syntax (e.g., counterfactual mood) is lost \\
Lexical disambiguation & 13 & A polysemous word is resolved to a different sense than the source intends in the presence of disambiguating context \\
Referential precision & 12 & Referential drift: a number, date, kinship side, measurement, or specific entity shifts \\
Control & 6 & None by design; baseline for floor-effect detection and pipeline diagnostics \\
\end{longtable}
}

\textbf{Table 3:} \emph{Round-trip translation source categories (100 items total). The post-hardening distribution reflects the recategorization of pragmatically driven items from structural complexity to discourse pragmatics (§3.3.5).}

Illustrative items, one per category (none is part of the live benchmark):

\begin{itemize}
\tightlist
\item
  \textbf{Discourse pragmatics:} ``I'm not saying he lied, but the figures he presented don't quite match what we found in the source documents.'' --- a deniable accusation whose pragmatic force must survive the explicit denial.
\item
  \textbf{Non-compositional:} ``It's raining cats and dogs, so the game is off.'' --- a literal rendering of the idiom does not round-trip back to ``raining heavily.''
\item
  \textbf{Implicit content:} ``He thinks he's a real Casanova, but he's more of a Walter Mitty.'' --- the meaning lives in the contrast between two allusions; if either is dropped, the characterization is lost.
\item
  \textbf{Structural complexity:} ``Had the agreement, which the negotiators had spent eighteen months drafting, been ratified before the election, the dispute that subsequently escalated would likely have been avoided.'' --- a past-counterfactual conditional carrying a nested relative clause; both the counterfactual structure and the reality it implies must survive.
\item
  \textbf{Lexical disambiguation:} ``The composer scored the film in under three weeks, despite the studio's demands for revisions.'' --- ``scored'' must resolve to the musical sense, forced by ``composer.''
\item
  \textbf{Referential precision:} ``Each of the fourteen claimants will receive a one-fortieth share of the settlement, with the remainder reverting to the estate.'' --- the round trip must preserve fourteen claimants, the one-fortieth share each, and the remainder reverting to the estate.
\item
  \textbf{Control:} ``The children played in the yard until dinner was ready.'' --- expected to round-trip with little loss in most languages; used as a baseline to confirm the pipeline is working.
\end{itemize}

Discourse pragmatics is the largest category in the post-hardening corpus. It includes classical pragmatic phenomena such as irony, hedging, and attribution, as well as sentences with two similar referents where the round trip preserves the events but loses track of who did what, which was the strongest discriminator found during hardening (§3.3.5).

We exclude several phenomena that cannot, in principle, be scored under round-trip methodology. Intentionally ambiguous items require preserving multiple simultaneous readings, but round-trip translation forces sense commitment; garden-path structures lose their ambiguity once resolved; and wordplay depending on English-internal homophony cannot survive a non-English intermediate. Each produces uniformly low scores that do not discriminate between models.

\subsection{3.3.3 Translation task evaluation}\label{translation-task-evaluation}

Models are evaluated on the translation task on a single-sentence basis: our custom Python app iterates over individual sentences and inserts them, together with the language name, into the following prompt:

\begin{Shaded}
\begin{Highlighting}[]
\NormalTok{Translate the following English sentence to \{language\_name\}.}
\NormalTok{Reply with only the translation, no explanations.}
\NormalTok{"\{sentence\}"}
\end{Highlighting}
\end{Shaded}

The \texttt{\{language\_name\}} field is filled with the full English language name, with a country or region qualifier where needed to disambiguate the targeted variety (for example, ``Tigrinya (Ethiopia)''). Each sentence is sent as an independent, zero-shot request with no system prompt or in-context examples. We evaluate the translation task at temperature 0 only; the rationale and the equivalence analysis supporting this choice are given in §3.5. We send no other generation parameters, except for reasoning-effort levels where these are part of the model identifier (for example, the \texttt{-high}, \texttt{-adaptive}, or \texttt{-medium} suffixes).

The translation produced in the target language is then translated back into English with the following prompt:

\begin{Shaded}
\begin{Highlighting}[]
\NormalTok{Translate the following \{language\_name\} sentence to English.}
\NormalTok{Reply with only the translation, no explanations.}
\NormalTok{"\{sentence\}"}
\end{Highlighting}
\end{Shaded}

The same model under evaluation performs both the forward and the back translation, and the parameters and options sent with the back-translation request are identical to the forward translation. The round-trip therefore measures one model's combined encode-and-decode behavior; we discuss the consequences of this design in §3.3.6.

\subsection{3.3.4 Judge panel}\label{judge-panel}

Each round-trip is scored independently by three LLM judges: GPT-5.4 Mini, Claude Haiku 4.5, and Gemini 3.1 Flash Lite\footnote{Exact judge model versions: \texttt{gpt-5.4-mini}, \texttt{claude-haiku-4.5}, \texttt{gemini-3.1-flash-lite-preview}. See Appendix B, Table B.3.}, run at temperature 0. We chose smaller models from three leading providers to diversify the panel and reduce cost. The only judgment required is an English-to-English comparison, on which models of this size prove reliable: on a stratified sample of 300 round-trips, the mean of the three judges agrees with the mean of three professional human annotators at Pearson r = 0.78 and Spearman \(\rho\) = 0.79, and the judges are more internally consistent than the annotators themselves (Krippendorff's \(\alpha\) = 0.82 versus 0.69). We report this validation in full, including per-tier results, in §4.

Because all three judges come from providers whose models we also evaluate, the panel is exposed to self-preference bias, in which a judge may favor outputs from its own model family (Panickssery et al. 2024). Three features of our setup limit this risk: judges compare English to English rather than assessing their own target-language output; the panel spans three providers, so averaging dilutes any single-vendor effect; and we validate the panel against humans (§4). We also measured the effect directly, comparing each judge's score to the mean of the other two on the same response. Pooled across judges, the same-maker lift is small but, given the sample size, statistically significant: +0.025 points on the 1--5 scale on the full set (95\% CI {[}0.021, 0.028{]}) and +0.034 on hard (95\% CI {[}0.028, 0.040{]}; both p ≤ 10⁻³⁰), translating to roughly +0.008--0.011 on the final three-judge mean (+0.002--0.003 normalized): well below the differences that separate models, whose hard-subset scores span 0.40 to 0.82 (§5.3). The effect is concentrated in a single judge (Google); the other two show negligible or non-significant self-preference (Appendix E). It is no larger in its strongest form, where a judge scores its own exact model (all three judges are themselves evaluated models): pooled +0.035 (95\% CI {[}0.014, 0.056{]}; Appendix E).

The judge prompt is as follows:

\begin{Shaded}
\begin{Highlighting}[]
\NormalTok{You are evaluating whether a round{-}trip translation preserved the meaning of an English sentence. The original was translated into another language and then back to English. Your job is to score how well the round{-}trip preserved meaning, on a 1{-}5 scale.}

\NormalTok{Score against the key meaning the original conveys, including any pragmatic, figurative, or implicit content (sarcasm, euphemism, indirect speech acts, idiom{-}as{-}meaning). Do not penalize differences in wording, register, or stylistic choices when the meaning is preserved. A round{-}trip that paraphrases or restates the meaning in plain language can be a 5.}

\NormalTok{Scale:}
\NormalTok{5 {-} Key meaning fully preserved. Wording may differ from the original.}
\NormalTok{4 {-} Key meaning mostly preserved, minor loss of detail or nuance.}
\NormalTok{3 {-} Core of the key meaning present, but a substantive component (a pragmatic implication, a specific quantity, a structural relationship, etc.) is lost or altered.}
\NormalTok{2 {-} Substantial component of the key meaning is missing or wrong.}
\NormalTok{1 {-} Key meaning substantially absent.}

\NormalTok{Original sentence:}
\NormalTok{"\{original\}"}
\NormalTok{\{target\_section\}}
\NormalTok{\{examples\_section\}}
\NormalTok{Round{-}trip translation:}
\NormalTok{"\{roundtrip\}"}

\NormalTok{Respond in this exact format, with no other text:}
\NormalTok{Score: \textless{}1{-}5\textgreater{}}
\NormalTok{Rationale: \textless{}one or two sentences explaining your score\textgreater{}}
\end{Highlighting}
\end{Shaded}

The \texttt{\{original\}} field holds the source sentence and \texttt{\{roundtrip\}} the round-tripped English. \texttt{\{target\_section\}} is filled with the item's explicit target meaning for non-control items, and \texttt{\{examples\_section\}} with worked scoring examples for the subset of items that have them (§3.3.1); both are omitted when not applicable. We use a 1-5 Likert scale: five levels give enough resolution to separate full preservation, minor loss, substantive loss, major loss, and absent meaning, while remaining coarse enough for reliable agreement across judges. The rationale field is requested for troubleshooting and is not used in scoring.

Judge calls inherit the model-call layer's retry logic (up to four attempts on transport or API errors); there is no additional benchmark-level retry for judges. Fewer than 0.2\% of judge calls fail to return a parseable score. The item score is the mean of the successful judge scores; if all three judges fail, the item is flagged and excluded.

\textbf{Example (illustrative item, not in the live benchmark).}

\begin{itemize}
\tightlist
\item
  \textbf{Original:} ``I'm not saying he lied, but the figures he presented don't quite match what we found in the source documents.''
\item
  \textbf{Target meaning (\texttt{\{target\_section\}}):} a conventional deniable accusation --- the speaker avoids a direct charge while implying the figures may have been misrepresented; the implicit suggestion of dishonesty, with deniability preserved, is the meaning to retain.
\item
  \textbf{Round-trip scored 5:} ``I'm not formally accusing him of lying, but the figures don't match the source documents.'' --- implicit accusation preserved.
\item
  \textbf{Round-trip scored 3:} ``I'm not saying he lied; the figures just don't match.'' --- surface preserved, but the implied accusation is no longer conveyed.
\end{itemize}

\textbf{Judge succession.} As judge models are deprecated, we replace them with successors from the same providers, validated exactly as the current panel was: a candidate judge is adopted only after its agreement with the retained judges and with human annotators is confirmed on the validation set (§4). Because we store all round-trip outputs, adopting a new panel requires only re-running the judges over the existing English pairs, not re-translating. We therefore re-score with each new panel: the full active corpus where feasible, or a calibration sample sufficient to measure and correct the offset between panels otherwise, and report under a versioned panel identifier, freezing prior-panel scores so that previously published results remain reproducible and cross-panel comparisons remain possible, mirroring the versioning of the hard subset (§3.3.5). We do not mix panels within a single reported result: every model on a given leaderboard version is scored by the same panel, so score differences reflect translation quality rather than which judge evaluated a given model.

\subsection{3.3.5 Corpus hardening and the hard subset}\label{corpus-hardening-and-the-hard-subset}

When the initial corpus was first evaluated, before hardening, scores were compressed at the top of the distribution: the four recent strong models we tracked clustered between 4.4 and 4.6 on the 1--5 scale, leaving little room to separate strong models or to register future progress.

We probed candidate sentences through the full production translation-and-judging pipeline without writing to the active dataset. Each candidate was translated into seven languages spanning all resource tiers (Czech, Polish, Finnish, Japanese, Simplified Chinese, Fijian, Kinyarwanda) by a frontier target model, back-translated to English, and scored by the three judges (§3.3.4); we retained items that scored consistently low across language pairs or failed catastrophically in low-resource pairs. In total, 133 new candidate items were probed, drawn from a range of hypothesized difficulty sources including fine-grained linguistic phenomena, documented machine-translation failure modes, paragraph-level discourse structure, cross-lingual lexical asymmetries, and pragmatic coreference (Appendix F).

Most hypothesized difficulty sources proved robust to round-trip translation under current frontier models: number and date handling, idiom preservation, quantifier scope, modal force, counterfactuals, and anaphora chains all scored near ceiling, and many failure modes documented in 2024--2025 work were largely mitigated by intervening model progress. The strongest and most consistent discrimination signal came from a single mechanism: failure to preserve which referent performed which action. The pattern arises when a source passage introduces two same-gender referents whose pronouns can only be resolved by pragmatic inference, not by grammatical gender. In the round trip, models tend to preserve the propositional content but drop the attribution of each action to a specific actor. A sentence like ``she dismissed his concerns'' comes back as ``the concerns were dismissed'': the meaning is intact, but who did what is lost. This held across intermediate languages of every resource tier. A complementary hypothesis (that round-trips split a single actor's action chain across two actors) did not replicate, locating the signal specifically in resolving ambiguous pronouns to distinct actors rather than in preserving unambiguous predicate chains.

We promoted 22 such items into the corpus, replacing the 22 lowest-discrimination items (those at or near ceiling across all four tracked models, excluding any item with substantial per-language minimum-score signal); the corpus remained at 100 items per language. The resulting category distribution is given in §3.3.2.

For headline reporting we define a fixed hard subset of the 50 source sentences with the lowest mean score across four recent strong models tested (Claude Opus 4.7-adaptive, Gemini 3 Flash Preview, Gemini 3.1 Pro Preview, GPT-5.4-medium) at a frozen snapshot. After promotion, the full corpus remained compressed for these four models, with overall scores spanning 4.33--4.51 (a 0.18 range), because it still contains many easy items and the six controls. The hard subset, by contrast, spreads the same models across 3.96--4.22 (a 0.26 range), roughly a 40\% gain in discrimination with no additional data collection (the subset is drawn from the existing 100 items, so it requires no new sentences or annotation; it in fact halves per-run evaluation cost by scoring 50 items rather than 100). All hard-subset results in this paper use this initial definition; if the subset is redefined in a future version, the leaderboard will label results accordingly.

\subsection{3.3.6 Remarks on translation task design}\label{remarks-on-translation-task-design}

The design choices that make judging reliable and results interpretable also impose limitations. First, we intentionally measure only preservation of meaning, not style or other surface markers, which vary substantially with intended audience, register, and personal preference. Translation as a craft, however, involves careful shaping of exactly these surface features, so our task cannot capture everything that contributes to a high-quality translation. Capturing those aspects would require human evaluation of translation quality itself, a different undertaking from the judge-validation study in §4; that study establishes that our judges track human \emph{meaning} judgments rather than stylistic quality.

Second, we measure the survival of meaning through a round trip, so we cannot rule out preservation by luck, or by a model making compensating errors in the forward and backward directions that happen to cancel. This is the long-standing objection to round-trip translation as an evaluation method (Somers 2005): a poor translation can round-trip back to match the source, while a good one can be garbled in return. Later work nonetheless finds round-trip quality correlates positively with one-way translation quality under human assessment (Aiken and Park 2010), and recent results report strong agreement between round-trip scores and human preference ratings for frontier LLMs (Skorobogat et al. 2026).

Third, the task cannot distinguish comprehension failures from generation failures, nor identify in which translation direction a problem arose, because the same model performs both steps (§3.3.3). We store the full forward and backward outputs and may analyze them in future work, but this is out of scope here.

Fourth, the benchmark is English-anchored: every source sentence is English, so we measure English↔target round-trips rather than the full translation matrix or asymmetries between translating into versus out of English. This is a deliberate scoping choice: it is what makes judging an English-to-English comparison (§3.3.4), but it bounds what the task can claim about general translation ability.

Measuring translation quality in an automated way remains a hard problem. While we consider our measurements indicative of models' ability to perform real-world translation, we caution against over-interpreting any single number. Automated metrics, whether reference-based scores such as BLEU (Papineni et al. 2002) and chrF (Popović 2015) or learned metrics such as COMET (Rei et al. 2020), capture only a narrow slice of translation quality and correlate imperfectly with human judgment (Freitag et al. 2020; Mathur et al. 2020), particularly for high-quality output where differences are subtle, for low-resource languages where reference data is sparse, and for literary or domain-specific translation where fluency, register, terminology, and cultural nuance matter as much as adequacy. We therefore report these figures as a useful but incomplete signal, to be read alongside human evaluation rather than as a substitute for it.

\subsection{3.3.7 Cross-language comparability}\label{cross-language-comparability-1}

Unlike the grammar task (§3.2.7), the round-trip translation task is designed from the outset to support comparison across languages, and two properties make those comparisons valid. First, every language is evaluated on the same set of English source sentences, so any difference between languages reflects the translation pipeline itself rather than differences in the content or difficulty of the items. Second, the judges always compare English to English, regardless of which language the text passed through, so the measurement instrument is equally reliable from one language to the next; the very property that direct multilingual evaluation lacks (§3.3 intro). Together, these let us compare how much meaning a model actually preserves across languages, identify which languages it handles best and worst in practice when translating into and from English, and track how those gaps narrow as newer models are released (§5.3).

One caveat shapes how these comparisons should be read. A per-language score is directly comparable as a measure of realized round-trip quality: it tells us how well meaning survives when a given model is used for that language pair. What it cannot do is separate two things that both contribute to the result: how capable the model is, and how intrinsically difficult it is to carry meaning between English and that language in the first place. When a model scores lower on English↔Japanese than on English↔Spanish, that lower score reflects genuinely greater meaning loss in deployment, and is meaningful on its own terms; but part of the gap comes from the model and part from the inherent distance and thinner data representation of the language pair, and the benchmark does not tell us in what proportion. Cross-language scores are therefore best understood as realized performance (what a user would actually experience) rather than as a ranking of a model's underlying abstract ability in each language.

\section{3.4 Task 3: Tokenizer Efficiency (Supplementary)}\label{task-3-tokenizer-efficiency-supplementary}

As a supplementary, practitioner-oriented measurement, we report how efficiently each model's tokenizer encodes each language. Tokenizer efficiency does not reflect a model's ability to understand or generate text, but it bears directly on deployment: LLM APIs bill per token and context windows are measured in tokens, so a language that tokenizes inefficiently incurs a ``multilingual tax'' in cost and effective context usage (Ahia et al. 2023; Petrov et al. 2023).

We measure efficiency as the average number of characters (Unicode code points) encoded per token. For each model and language, we concatenate the grammar-task sentences (§3.2) into a single message, send it as the request payload, and divide the total character count by the input token count reported by the API. Reusing the grammar sentences means efficiency is measured on real in-language text rather than a synthetic corpus. We concatenate rather than send sentences individually because some models add a fixed per-request token overhead (for example, when reasoning is enabled), in some cases equal to the token count of the sentence itself; concatenating 100 sentences reduces this overhead's relative contribution by roughly 100×, to a low single-digit percentage. We report each language as a per-character token-cost multiplier relative to English: the ratio of English characters-per-token to that language's characters-per-token, computed within each tokenizer. Because models from the same provider almost always share a tokenizer, results are grouped by tokenizer family. Full methodology, the per-request overhead analysis, and the complete per-family table are in Appendix G.

The tax is large, concentrated in non-Latin scripts, and varies widely across tokenizer families. Relative to English, close European languages cost only ≈1.1× per character, while languages such as Chinese, Japanese or Korean cost ≈2.5--4.7× and the lowest-resource scripts far more. How much more depends heavily on the tokenizer: Google's keeps even the most expensive languages to ≤3.4× (Tigrinya 3.4×, Khmer 2.2×), whereas the Mistral and NVIDIA tokenizers reach 10--13× on the same languages. Tokenizer choice is therefore a first-order cost lever for multilingual deployment, independent of model quality: the same Khmer workload can cost roughly six times as much under one provider's tokenizer as under another's.

One caveat bears on how these multipliers should be read. Because the grammar sentences are not parallel across languages, the multiplier measures relative token cost per character, not per unit of content. Within a single language this distinction does not matter: every tokenizer is encoding the same text, so any difference in token count reflects the tokenizers themselves, and comparisons between tokenizer families on the same language (including the Khmer comparison above) remain fully valid. These within-language comparisons are the ones we emphasize.

Across scripts, however, characters carry different amounts of information: a Han character conveys more than a Latin one, and compact scripts express the same content in fewer characters, so the content-level cost of, say, Chinese text is substantially lower than its per-character multiplier suggests. Cross-script multipliers should therefore be read as per-character cost ratios rather than as the cost of processing the same document, which would require a parallel corpus. In future versions we may switch the measurement to highly rated round-trip translations of the shared source sentences (§3.3), which would make the multipliers content-matched across languages; we have not done so here because translation quality in the lowest-resource languages is not yet sufficient for the translated text to serve as a measurement corpus (§5.4.1), whereas the grammar sentences are linguist-validated native text in every language.

\section{3.5 Models Evaluated}\label{models-evaluated}

We evaluate 53 models from 13 providers in 82 configurations (Table 4). A \emph{model} is a distinct base system (e.g., GPT-5.5); a \emph{configuration} is a model together with a reasoning setting (e.g., GPT-5.5 at high reasoning effort). Where a model exposes a reasoning or thinking mode, each setting is evaluated as a separate configuration, because reasoning changes both performance and cost (§5.4.2, §5.4.3); this is why the number of configurations exceeds the number of models. Reasoning-effort settings are carried in the model identifier through suffixes such as \texttt{-high}, \texttt{-medium}, \texttt{-minimal}, and \texttt{-adaptive}. The set spans frontier and small models, open- and closed-weight systems, and release dates from May 2024 to June 2026; evaluations were run between March and July 2026. Exact model version strings and per-configuration details are listed in Appendix B, Table B.1.

{\def\LTcaptype{none} 
\begin{longtable}[]{@{}
  >{\raggedright\arraybackslash}p{(\linewidth - 10\tabcolsep) * \real{0.1373}}
  >{\raggedleft\arraybackslash}p{(\linewidth - 10\tabcolsep) * \real{0.0980}}
  >{\raggedleft\arraybackslash}p{(\linewidth - 10\tabcolsep) * \real{0.0980}}
  >{\raggedright\arraybackslash}p{(\linewidth - 10\tabcolsep) * \real{0.3529}}
  >{\raggedright\arraybackslash}p{(\linewidth - 10\tabcolsep) * \real{0.1569}}
  >{\raggedright\arraybackslash}p{(\linewidth - 10\tabcolsep) * \real{0.1569}}@{}}
\toprule\noalign{}
\begin{minipage}[b]{\linewidth}\raggedright
Provider
\end{minipage} & \begin{minipage}[b]{\linewidth}\raggedleft
Models
\end{minipage} & \begin{minipage}[b]{\linewidth}\raggedleft
Configs
\end{minipage} & \begin{minipage}[b]{\linewidth}\raggedright
Reasoning configurations
\end{minipage} & \begin{minipage}[b]{\linewidth}\raggedright
Weights
\end{minipage} & \begin{minipage}[b]{\linewidth}\raggedright
Releases
\end{minipage} \\
\midrule\noalign{}
\endhead
\bottomrule\noalign{}
\endlastfoot
OpenAI & 13 & 23 & Effort-graded (none/medium/high) on GPT-5.2, 5.4, 5.4 Mini, 5.4 Nano, 5.5; GPT-5, 5 Mini, 5 Nano, GPT-OSS 120B on by default & Closed; GPT-OSS 120B open & 2024-05 to 2026-04 \\
Anthropic & 10 & 19 & Extended thinking and adaptive reasoning on Haiku 4.5, Opus 4.5 to 4.8, Sonnet 4.5, 4.6; Fable 5, Sonnet 5 on by default & Closed & 2025-05 to 2026-06 \\
Google & 7 & 10 & Gemini 2.5 Flash/Pro, 3 Flash, 3.1 Pro, 3.5 Flash, Gemma 4 on by default; 3 Flash also run at minimal, 3.5 Flash at minimal and high & Closed; Gemma 4 open & 2025-06 to 2026-05 \\
Mistral & 5 & 7 & Reasoning-high on Medium 3.5, Small 4 & Medium and Large closed; Small open & 2025-06 to 2026-04 \\
DeepSeek & 4 & 6 & Thinking toggle on V4 Flash, V4 Pro & Open & 2025-08 to 2026-04 \\
Alibaba & 3 & 5 & Reasoning toggle on Qwen 3.6 Max, 3.7 Max & Qwen 3-235B and 3.6 open; 3.7 closed & 2025-07 to 2026-05 \\
Meta & 2 & 2 & None & Open & 2024-12 to 2025-04 \\
xAI & 2 & 3 & Reasoning toggle & Closed & 2026-03 to 2026-04 \\
Moonshot & 2 & 2 & On by default (K2.5, K2.6) & Open & 2026-01 to 2026-04 \\
NVIDIA & 2 & 2 & On by default (Nemotron 3 Super, Ultra) & Open & 2026-03 to 2026-06 \\
ByteDance & 1 & 1 & On by default (Seed 2.0 Lite) & Closed & 2026-03 \\
MiniMax & 1 & 1 & On by default (M3) & Open & 2026-05 \\
Z.ai & 1 & 1 & On by default (GLM 5.2) & Open & 2026-06 \\
\end{longtable}
}

\textbf{Table 4:} \emph{The 53 evaluated models and 82 configurations, grouped by provider. ``Reasoning configurations'' names the models evaluated in more than one reasoning setting; provider terminology varies (effort levels, extended thinking, adaptive, thinking toggle), and we group these as reasoning-enabled configurations throughout. ``On by default'' marks single-configuration models that emit reasoning by default without a toggle (identified by output-token profile; Appendix B.1) and are run only in that mode. ``None'' marks non-reasoning models. Configuration counts include the default configuration of every model. Full per-configuration details, including exact version strings, prices, and hosting endpoints, are in Appendix B, Table B.1.}\par\smallskip

We additionally group models into three capability classes: Small (under 50B parameters), Medium (50 to 500B), and Frontier (over 500B). Parameter counts are undisclosed for most closed models, and for mixture-of-experts models the total and active counts diverge too widely for either to map cleanly onto capability; in both cases we follow the provider's positioning of the model within its own lineup. Nemotron 3 Ultra (550B total, 55B active) and Llama 4 Maverick (\textasciitilde400B total, 17B active) are classed Medium on this basis. Classes are used for grouping and presentation rather than as an analytical variable, and are distinct from the language resource tiers of §3.1.

{\def\LTcaptype{none} 
\begin{longtable}[]{@{}
  >{\raggedright\arraybackslash}p{(\linewidth - 4\tabcolsep) * \real{0.3000}}
  >{\raggedright\arraybackslash}p{(\linewidth - 4\tabcolsep) * \real{0.3000}}
  >{\raggedleft\arraybackslash}p{(\linewidth - 4\tabcolsep) * \real{0.4000}}@{}}
\toprule\noalign{}
\begin{minipage}[b]{\linewidth}\raggedright
Capability class
\end{minipage} & \begin{minipage}[b]{\linewidth}\raggedright
Definition
\end{minipage} & \begin{minipage}[b]{\linewidth}\raggedleft
Configurations
\end{minipage} \\
\midrule\noalign{}
\endhead
\bottomrule\noalign{}
\endlastfoot
Frontier & over 500B dense-equivalent parameters, or flagship positioning & 41 \\
Medium & 50 to 500B dense-equivalent, or mid-line positioning & 30 \\
Small & under 50B dense-equivalent, or small-line positioning & 11 \\
\textbf{Total} & & \textbf{82} \\
\end{longtable}
}

\textbf{Table 5:} \emph{Evaluated configurations by capability class. Parameter thresholds apply to disclosed dense models; for mixture-of-experts models (where total and active counts diverge) and undisclosed models, provider positioning overrides the count (§3.5). On this basis Nemotron 3 Ultra (550B total, 55B active) and Llama 4 Maverick (\textasciitilde400B total, 17B active) are classed Medium. Per-configuration assignments are listed in Appendix B, Table B.1.}

Models are accessed through their providers' APIs or, where more practical, through hosting services (Azure, Amazon Bedrock, OpenRouter); the appendix lists the endpoint used for each configuration. Where a model is available through multiple hosts, we evaluate one endpoint and exclude duplicates.

For every evaluated configuration, we track input and output token prices (per 1M tokens) and compute the cost of each benchmark run, broken down by task and language. Reasoning tokens are counted as output tokens, because some providers do not report them separately. Prices are as recorded in the reporting snapshot (July 2026), so all configurations carry the snapshot's prices regardless of when they were tested: models released in 2025 or earlier are costed at 2026 prices rather than their launch prices, and the public leaderboard carries current figures. To keep the cost metric comparable across models, we do not account for volume discounts, batch-execution rates, long-context surcharges, or other provider-specific pricing schemes. Configurations of private models without public pricing are excluded from cost analyses. A small number of requests fail after all retries; affected items are scored as incorrect, never exceeding 0.5\% of items for any configuration.

\subsection{3.5.1 Model addition and retirement}\label{model-addition-and-retirement}

The benchmark is maintained as a living leaderboard, and the evaluated set grows on a rolling basis. A model is added if it is a general-purpose text model (rather than a coding-, agentic-, or otherwise domain-specialized one) and is accessible to us through a public API or a private API provided by its developer; providers may have unreleased models evaluated ahead of public availability through direct collaboration (§7). New models are evaluated under the current judge-panel and subset versions, and results for a given leaderboard version are always produced by a single panel, so scores remain comparable within a version (§3.3.4, §3.3.5). Evaluated models are never removed: their scores remain on the leaderboard permanently, frozen under the versions in effect when they were scored. By default the leaderboard displays models released within the last 12 months, with older models available through a filter; where a model's API has since been deprecated, its scores persist under their original panel version and are labeled accordingly.

\subsection{3.5.2 Temperature policy and equivalence}\label{temperature-policy-and-equivalence}

The grammar task is run once at temperature 0 and three times at each model's default temperature, capturing both deterministic behavior and run-to-run variability (reported as Avg 3-Run and Worst 3-Run). The translation task is run only at temperature 0. Translation is our most expensive task, since each item requires a forward translation, a back-translation, and three judge evaluations, so running it three times at default temperature would roughly quadruple its cost. Temperature 0 is also the appropriate setting for translation, where faithful, deterministic output is typically preferred to sampling diversity.

Some reasoning models enforce a default temperature when reasoning is enabled and do not accept temperature 0. For these models the provider request omits the temperature parameter, while we preserve the temperature-0 bucket in reporting; figures labeled \texttt{temp=0.00*} include such models, and the asterisk denotes this convention.

Temperature 0 is not, strictly speaking, deterministic. Setting the temperature to 0 tells the model to always pick its highest-scoring next word rather than sampling among the plausible ones, which removes the deliberate randomness but not quite all of it. The scores themselves are computed on GPUs, in an order that shifts with how many requests are being processed together and on what hardware, and the resulting rounding differences are occasionally large enough to reverse which of two closely matched words comes out on top. One such reversal can then send the rest of the response down a different path (He and Thinking Machines Lab 2025; Yuan et al. 2025). Repeated runs at nominally fixed settings have been shown to move task accuracy by a few points, most of all for reasoning models, where an early divergence has a long chain of thought to compound in (Atil et al. 2024; Yuan et al. 2025; Messina and Scotta 2026). We therefore make no claim that our runs are exactly reproducible, particularly as we query hosted APIs whose batching and hardware are outside our control. The benefit of temperature 0 is not absolute determinism but a large reduction in variance: the variation that remains is limited to the few items where the model is nearly torn between two options, whereas sampling can produce a different answer anywhere the model is less than certain (Song et al. 2024). That remaining variation is smaller than the differences we treat as meaningful, since per-model confidence intervals on grammar MCC span roughly ±0.03 to ±0.04 and we advise against reading gaps below 0.05 as rankings (§5.1). For reporting purposes, then, temperature 0 can be treated as deterministic, and we write ``near-deterministic'' where the distinction matters.

To check that reporting at temperature 0 is representative of each model's performance, we ran a paired item-level equivalence test (two one-sided tests, \(\alpha\) = 0.05) comparing temperature 0 against default temperature on a common subset for both tasks: six languages spanning all resource tiers (German, Japanese, Czech, Thai, Tagalog, Assamese) and five models. Temperature 0 is a single run; default temperature is summarized as the mean of three runs. To make the two tasks comparable, both metrics were normalized to {[}0,1{]} --- grammar accuracy directly, translation via (score − 1)/4 --- and we used a single equivalence margin of ±0.05 (5\% of scale).

{\def\LTcaptype{none} 
\begin{longtable}[]{@{}
  >{\raggedright\arraybackslash}p{(\linewidth - 10\tabcolsep) * \real{0.1500}}
  >{\raggedright\arraybackslash}p{(\linewidth - 10\tabcolsep) * \real{0.1833}}
  >{\raggedleft\arraybackslash}p{(\linewidth - 10\tabcolsep) * \real{0.0500}}
  >{\raggedleft\arraybackslash}p{(\linewidth - 10\tabcolsep) * \real{0.1833}}
  >{\raggedright\arraybackslash}p{(\linewidth - 10\tabcolsep) * \real{0.2333}}
  >{\centering\arraybackslash}p{(\linewidth - 10\tabcolsep) * \real{0.2000}}@{}}
\toprule\noalign{}
\begin{minipage}[b]{\linewidth}\raggedright
Task
\end{minipage} & \begin{minipage}[b]{\linewidth}\raggedright
Model
\end{minipage} & \begin{minipage}[b]{\linewidth}\raggedleft
n
\end{minipage} & \begin{minipage}[b]{\linewidth}\raggedleft
Mean Δ (norm.)
\end{minipage} & \begin{minipage}[b]{\linewidth}\raggedright
90\% CI
\end{minipage} & \begin{minipage}[b]{\linewidth}\centering
Equivalent
\end{minipage} \\
\midrule\noalign{}
\endhead
\bottomrule\noalign{}
\endlastfoot
Grammar & deepseek-v4-flash & 600 & −0.034 & {[}−0.065, −0.002{]} & No \\
Grammar & gemini-3-flash-preview & 600 & +0.017 & {[}−0.001, +0.034{]} & Yes \\
Grammar & gpt-5.4 & 600 & −0.001 & {[}−0.023, +0.021{]} & Yes \\
Grammar & grok-4.3 & 600 & −0.001 & {[}−0.034, +0.033{]} & Yes \\
Grammar & qwen-3.6-max-preview & 600 & +0.019 & {[}−0.005, +0.044{]} & Yes \\
Translation & deepseek-v4-flash & 600 & −0.003 & {[}−0.013, +0.007{]} & Yes \\
Translation & gemini-3-flash-preview & 598 & +0.005 & {[}−0.003, +0.012{]} & Yes \\
Translation & gpt-5.4 & 600 & +0.003 & {[}−0.007, +0.012{]} & Yes \\
Translation & grok-4.3 & 600 & −0.033 & {[}−0.046, −0.020{]} & Yes \\
Translation & qwen-3.6-max-preview & 600 & −0.040 & {[}−0.052, −0.029{]} & No \\
\end{longtable}
}

\textbf{Table 6:} \emph{Temperature-0 vs default-temperature equivalence, pooled by model. Δ is the default-minus-zero mean difference on the {[}0,1{]}-normalized metric (grammar accuracy; translation (score − 1)/4). Equivalence is declared when the 90\% confidence interval (CI) falls within the ±0.05 margin. Gemini's translation n is 598 owing to two failed generation calls excluded by pairwise-complete filtering.}

At the model level (pooling items across languages), the two settings are equivalent for four of five models on each task, with small mean differences throughout (Table 6; all \textbar Δ\textbar{} ≤ 0.04 on the normalized scale). The single exception per task is a model whose default-temperature score is modestly lower: DeepSeek on grammar (−0.034 in accuracy) and Qwen on translation (−0.16 points on the 1--5 scale). A complementary test on MCC, the headline grammar metric, confirms equivalence at the pooled level (mean difference +0.009, 90\% CI within ±0.10). Because temperature 0 is applied uniformly and the accuracy-based discrepancies run toward lower default scores, temperature 0 does not understate performance on these; for the two affected models, default-temperature output may fall slightly below the reported figures. The one exception to that direction is on the MCC metric, where Qwen's grammar MCC is modestly higher at default temperature (+0.068, CI exceeding the ±0.10 margin; Appendix H.3), so for that model and metric the temperature-0 figure is if anything conservative. Per-language slices remain in Appendix H, where we caution that slice-level equivalence tests (n ≈ 100) are underpowered at this margin and report observed differences rather than equivalence flags.

\begin{center}\rule{0.5\linewidth}{0.5pt}\end{center}

\section{4. Human Validation Study}\label{human-validation-study}

Translation scores depend on the LLM judge panel (§3.3.4). To establish that the panel tracks human judgment, three professional linguists scored a stratified sample of round-trip pairs under the same rubric, and we compared human and judge ratings.

\section{4.1 Study design}\label{study-design}

\textbf{Sample.} We drew 300 round-trip evaluations, balanced across resource tiers and score levels: 100 per tier (T1, T2, T3), and within each tier 25 items in each of four judge-score bands ({[}1,2), {[}2,3), {[}3,4), {[}4,5{]}). Most round-trips in normal use score near the top of the scale. By drawing an equal number of items from each score band, we test agreement across the full range of quality rather than mostly on near-perfect round-trips. The sample spans 12 languages, four per tier chosen for script and family diversity (T1: German, Japanese, Arabic, Simplified Chinese; T2: Czech, Thai, Tamil, Khmer; T3: Tagalog, Assamese, Tigrinya, Dzongkha). We also limited how many items could come from any one model or language, so the sample is not dominated by a few of them.

\textbf{Annotators.} Three professional linguists, all native English speakers with more than ten years of professional experience in translation and linguistics, scored the sample independently and blind to both the judge scores and one another's ratings. None had worked on the creation of the translation source sentences.

\textbf{Task parity.} Annotators saw exactly what the judges saw: the original English sentence, the round-trip English, the item's target meaning, and the worked scoring examples for items that have them (§3.3.4). They did not see the intermediate-language translation, and they used the same 1--5 rubric.

\section{4.2 Results}\label{results}

We report agreement within two cohorts (among the LLM judges, among the human annotators) using ordinal Krippendorff's \(\alpha\), Pearson r (Fisher-z averaged over rater pairs), and exact and within-one agreement (Table 7). Because exact agreement on a five-point scale is inherently low, we treat \(\alpha\) and within-one agreement as the primary pairwise measures. We additionally report the correlation between the three-judge mean and the three-human mean per item (Table 8), the most direct test of whether the panel reproduces human consensus.

{\def\LTcaptype{none} 
\begin{longtable}[]{@{}
  >{\raggedright\arraybackslash}p{(\linewidth - 12\tabcolsep) * \real{0.2436}}
  >{\raggedleft\arraybackslash}p{(\linewidth - 12\tabcolsep) * \real{0.2051}}
  >{\raggedleft\arraybackslash}p{(\linewidth - 12\tabcolsep) * \real{0.1410}}
  >{\raggedleft\arraybackslash}p{(\linewidth - 12\tabcolsep) * \real{0.0897}}
  >{\raggedleft\arraybackslash}p{(\linewidth - 12\tabcolsep) * \real{0.1538}}
  >{\raggedleft\arraybackslash}p{(\linewidth - 12\tabcolsep) * \real{0.0641}}
  >{\raggedleft\arraybackslash}p{(\linewidth - 12\tabcolsep) * \real{0.1026}}@{}}
\toprule\noalign{}
\begin{minipage}[b]{\linewidth}\raggedright
Cohort
\end{minipage} & \begin{minipage}[b]{\linewidth}\raggedleft
Krippendorff \(\alpha\)
\end{minipage} & \begin{minipage}[b]{\linewidth}\raggedleft
Pearson r
\end{minipage} & \begin{minipage}[b]{\linewidth}\raggedleft
Exact
\end{minipage} & \begin{minipage}[b]{\linewidth}\raggedleft
Within-one
\end{minipage} & \begin{minipage}[b]{\linewidth}\raggedleft
n
\end{minipage} & \begin{minipage}[b]{\linewidth}\raggedleft
Raters
\end{minipage} \\
\midrule\noalign{}
\endhead
\bottomrule\noalign{}
\endlastfoot
LLM judges & 0.82 & 0.83 & 59\% & 92\% & 300 & 3 \\
Human annotators & 0.69 & 0.71 & 51\% & 79\% & 300 & 3 \\
\end{longtable}
}

\textbf{Table 7:} \emph{Inter-rater agreement within each cohort. \(\alpha\) and within-one agreement are the primary measures; exact agreement on a five-point scale is inherently low.}

The judge panel is highly consistent (\(\alpha\) = 0.82, within-one = 92\%), and notably more consistent than the three human annotators are with one another (\(\alpha\) = 0.69, within-one = 79\%); this holds within every resource tier (Appendix I, Table I.2). Individual judge--human rater pairs agree somewhat less strongly (mean Pearson r ≈ 0.65) than judges agree among themselves (≈ 0.83), as is common when comparing across rater types (full per-pair breakdown in Appendix I, Table I.1). At the level the benchmark actually uses, however, agreement is strong: the mean of the three judges correlates with the mean of the three humans at r = 0.78 (95\% CI {[}0.72, 0.82{]}) and Spearman \(\rho\) = 0.79, which is at least as high as agreement among the human annotators themselves (pairwise r ≈ 0.71). Averaging within each panel cancels individual-rater noise, so the panel score is a reliable stand-in for human consensus.

{\def\LTcaptype{none} 
\begin{longtable}[]{@{}
  >{\raggedright\arraybackslash}p{(\linewidth - 10\tabcolsep) * \real{0.2329}}
  >{\raggedright\arraybackslash}p{(\linewidth - 10\tabcolsep) * \real{0.2329}}
  >{\raggedright\arraybackslash}p{(\linewidth - 10\tabcolsep) * \real{0.2466}}
  >{\raggedleft\arraybackslash}p{(\linewidth - 10\tabcolsep) * \real{0.1781}}
  >{\raggedleft\arraybackslash}p{(\linewidth - 10\tabcolsep) * \real{0.0548}}
  >{\raggedleft\arraybackslash}p{(\linewidth - 10\tabcolsep) * \real{0.0548}}@{}}
\toprule\noalign{}
\begin{minipage}[b]{\linewidth}\raggedright
Scope
\end{minipage} & \begin{minipage}[b]{\linewidth}\raggedright
Pearson r (95\% CI)
\end{minipage} & \begin{minipage}[b]{\linewidth}\raggedright
Spearman \(\rho\) (95\% CI)
\end{minipage} & \begin{minipage}[b]{\linewidth}\raggedleft
Judge − human
\end{minipage} & \begin{minipage}[b]{\linewidth}\raggedleft
MAE
\end{minipage} & \begin{minipage}[b]{\linewidth}\raggedleft
n
\end{minipage} \\
\midrule\noalign{}
\endhead
\bottomrule\noalign{}
\endlastfoot
Overall & 0.78 {[}0.72, 0.82{]} & 0.79 {[}0.73, 0.83{]} & −0.21 & 0.69 & 300 \\
T1 (high-resource) & 0.80 {[}0.70, 0.87{]} & 0.80 {[}0.68, 0.87{]} & −0.29 & 0.65 & 100 \\
T2 (mid-resource) & 0.77 {[}0.68, 0.83{]} & 0.78 {[}0.69, 0.85{]} & −0.35 & 0.76 & 100 \\
T3 (low-resource) & 0.78 {[}0.67, 0.86{]} & 0.80 {[}0.69, 0.88{]} & +0.00 & 0.64 & 100 \\
\end{longtable}
}

\textbf{Table 8:} \emph{Agreement between the three-judge mean and the three-human mean, per item, overall and by resource tier. ``Judge − human'' is the mean signed difference (negative means judges score lower). Bootstrap CIs from 5,000 resamples.}

\textbf{Robustness across tiers.} Panel-level agreement is essentially constant across resource tiers (Table 7): T1 r = 0.80, T2 r = 0.77, T3 r = 0.78 (\(\rho\) = 0.80). Agreement on low-resource languages is therefore no weaker than on high-resource languages, directly addressing the concern that LLM judgments become unreliable for low-resource languages (Fu and Liu 2025). This holds because the judges operate only on the English round-trip, independent of the intermediate language (§3.3).

\textbf{Systematic bias.} Judges score 0.21 points lower than humans on average (mean absolute error 0.69); the panel is mildly conservative and does not inflate model quality. Because the same panel scores every model, this constant offset does not affect model rankings. The bias is near zero for low-resource languages (T3: +0.00) and larger for higher-resource tiers (T1: −0.29, T2: −0.35). Per-score-band breakdowns are in Appendix I (Table I.3), where we note that within-band range restriction mechanically depresses correlation even under close agreement.

The validation sample is balanced across score bands rather than following the natural, ceiling-heavy score distribution, so these figures reflect a more stringent test of agreement than in-deployment scoring would face.

\begin{center}\rule{0.5\linewidth}{0.5pt}\end{center}

\section{5. Results}\label{results-1}

\section{5.1 Overview}\label{overview}

We evaluate 53 models in 82 configurations across both tasks. The full model-by-language result tables are given in Appendix J, and the public leaderboard renders them as heatmaps. The two tasks are scored on different scales: translation as a normalized meaning-preservation score on {[}0, 1{]}, and grammar as MCC on {[}−1, +1{]} with 0 at chance, so their numbers should not be read as a head-to-head comparison. Judged each against its own scale, the picture differs sharply. On translation, the strongest configurations preserve meaning well even on the hard subset, with the best (GPT-5.5 with high reasoning) reaching a normalized score of 0.82. On grammar, no model approaches reliable detection: the best reaches an MCC of only 0.36 (Gemini 3.1 Pro), and most models sit near chance (§5.4.5).

The two tasks measure related but distinct abilities. Across models, grammar and translation performance are only moderately correlated (Pearson 0.56, Spearman 0.63; 95\% CIs {[}0.44, 0.71{]} and {[}0.49, 0.77{]} from a bootstrap that resamples base models rather than configurations, so that multiple configurations of the same model are not treated as independent), and several models sit far off the diagonal: the GPT-5 family translates among the best while scoring weakly on grammar, while the Mistral models invert this ordering, ranking among the weakest translators but sitting in the upper half on grammar, a task on which scores are low across the board. A single ``multilingual'' number would average these apart, which is one reason the benchmark reports the two tasks separately rather than as a composite.

Per-model confidence intervals are reported on the leaderboard; as calibration, a 95\% bootstrap interval on a per-model grammar MCC over the full item set spans roughly ±0.03 to ±0.04, so per-model differences smaller than about 0.05 should not be read as rankings.

\section{5.2 Grammar Task Results}\label{grammar-task-results}

Per-model mean MCC, averaged across languages, runs from 0.36 at the top (Gemini 3.1 Pro) down to −0.34 (Qwen 3-235B), and the two leading models each answer about 68\% of items correctly overall; most models fall at or near chance, a pattern we take up in detail in §5.4.5. Scores well below chance are not noise or a scoring artifact (we verified the floor model's responses are parsed faithfully) but a signature of the item selection: sentences were chosen to trick a fixed construction panel (§3.2.1), so panel members and models that share their judgments face items anti-correlated with those judgments by construction. Because qualification required tricking at least two independent panel models (§3.2.1), the selected items concentrate on field-wide blind spots rather than any single model's quirk, so a model that shares those blind spots scores below chance even when it is neither on the panel nor from a panel provider --- as with the floor model, Qwen 3-235B. The bottom of the range accordingly clusters around the construction panel: two panel members (Llama 4 Maverick, GPT-4o) also sit among the lowest scores.

Because the grammar items differ by language, we do not compare languages on a difficulty scale (§3.2.7); we can, however, note where models struggle in agreement. Essentially all models perform at or below chance on the Khmer, Japanese, Polish, Tagalog, and Fijian stumper sets, while the Nepali, Albanian, Indonesian, Brazilian Portuguese, and Kinyarwanda sets are cleared by most. This reflects the difficulty of those particular item sets and the models' shared strengths and gaps, not an intrinsic ranking of the languages.

Grammar performance is also unstable across runs at default temperature. Within a language, a model's best and worst of the three default-temperature runs differ by a median of 0.22 in accuracy --- roughly double the ≈0.10 range expected from sampling noise alone on 100 balanced items, so this reflects genuine run-to-run instability rather than measurement error --- and the Worst-3-Run aggregate, which takes each language's worst of the three runs, falls about 0.11 below the three-run average; the most variable models include DeepSeek V3.2, Grok 4.20, and Mistral Small 4. This instability is why we report the near-deterministic temperature-0 score as the headline and track Worst-3-Run separately as a reliability measure (§3.5.2).

No stumper item is beyond every model: across all models and runs, each item is answered correctly by at least one, as expected for a balanced binary task with many attempts. A small set is nonetheless solved by almost no one, with 25 items answered correctly in fewer than 5\% of model-runs. The stumpers are hard but not impossible.

\section{5.3 Translation Task Results}\label{translation-task-results}

Translation scores span 0.40 (Nemotron 3 Super) to 0.82 (GPT-5.5 high) on the hard subset. Performance drops sharply with language resource level, from around 0.85 for high-resource European languages to below 0.35 for the lowest-resource languages; we quantify this relationship in §5.4.1.

Difficulty is concentrated in a specific kind of item. Averaged over models, the implicit content (0.62) and discourse pragmatics (0.63) categories are the hardest, while control items sit near the top (0.79), and the individual items with the lowest scores are almost all discourse pragmatics and implicit content. Meaning is therefore lost pragmatically rather than lexically or syntactically, consistent with the mechanism identified during corpus hardening (§3.3.5).

The three judges agree closely. On the hard subset, averaged across languages, inter-judge agreement is high (Krippendorff's \(\alpha\) = 0.77, mean pairwise Pearson r = 0.77, pairwise within-one agreement 93\%); on the full item set it is essentially the same (\(\alpha\) = 0.77, r = 0.80, within-one = 95\%). Per-language hard-subset values are in Table J.7.

No source item defeats every model either. On the hardest items, concentrated in implicit content and discourse pragmatics, the strongest model still preserves meaning well (normalized score at or above 0.83) even where the average model fails (item mean below 0.5). The hard subset thus separates strong models from weak ones rather than consisting of items that are impossible in principle, which is the behavior it was designed for (§3.3.5).

\section{5.4 Key Findings}\label{key-findings}

Beyond the per-model and per-language tables, several patterns emerge that we believe are of broader interest to model developers and to teams deploying LLMs in multilingual settings. We present the five most robust below. Each is supported by one or more figures, and for each we note the main threat to interpretation so that the result is not read more strongly than the data allows. All results use the hard subset for the translation task and temperature 0 unless noted.

\subsection{5.4.1 Translation quality tracks pretraining representation}\label{translation-quality-tracks-pretraining-representation}

The clearest signal in the translation results is that meaning preservation depends heavily on how well a language is represented in pretraining data. Using each language's share of Common Crawl as a proxy for that representation, we find a strong correlation between the log of the Common Crawl share and the average normalized translation score across the 29 target languages (Pearson r = 0.86, 95\% CI {[}0.73, 0.94{]}; Spearman \(\rho\) = 0.85, 95\% CI {[}0.67, 0.94{]}; Figure 1).\footnote{The 29 languages are not statistically independent observations, since related languages share history and, often, resource levels; ten of the targets are Indo-European. Three checks show the correlation is not an artifact of family structure. A cluster bootstrap that resamples the 15 family-level groups (the 14 language families of §3.1 plus the Basque isolate) rather than individual languages gives a 95\% CI of {[}0.77, 0.92{]}; collapsing each group to its mean leaves r = 0.89 (n = 15); and the relationship holds both within the Indo-European family alone (r = 0.86, n = 10, spanning five orders of magnitude of Common Crawl share from German to Assamese) and with Indo-European excluded entirely (r = 0.84, n = 19).} The two ends of the spectrum are far apart: high-resource European languages such as German (0.88), Portuguese (0.89) and French (0.87) sit near the top, while the lowest-resource languages collapse to a fraction of that, with Fijian at 0.29, Quechua at 0.30 and Guarani at 0.31.

\begin{figure}
\centering
\pandocbounded{\includegraphics[keepaspectratio,alt={Average normalized round-trip translation score per target language (mean across all evaluated models) plotted against the language's share of Common Crawl on a log scale, a proxy for pretraining representation. Each point is one of the 29 target languages, colored and shaped by resource tier (§3.1); the line is an ordinary least squares fit with a 95\% confidence band. Translation quality is strongly associated with pretraining representation (Pearson r = 0.86, Spearman \textbackslash rho = 0.85, n = 29). Common Crawl share is a coarse proxy and likely understates non-Latin-script languages, so the figure should be read as a strong association rather than an exact predictive relationship.}]{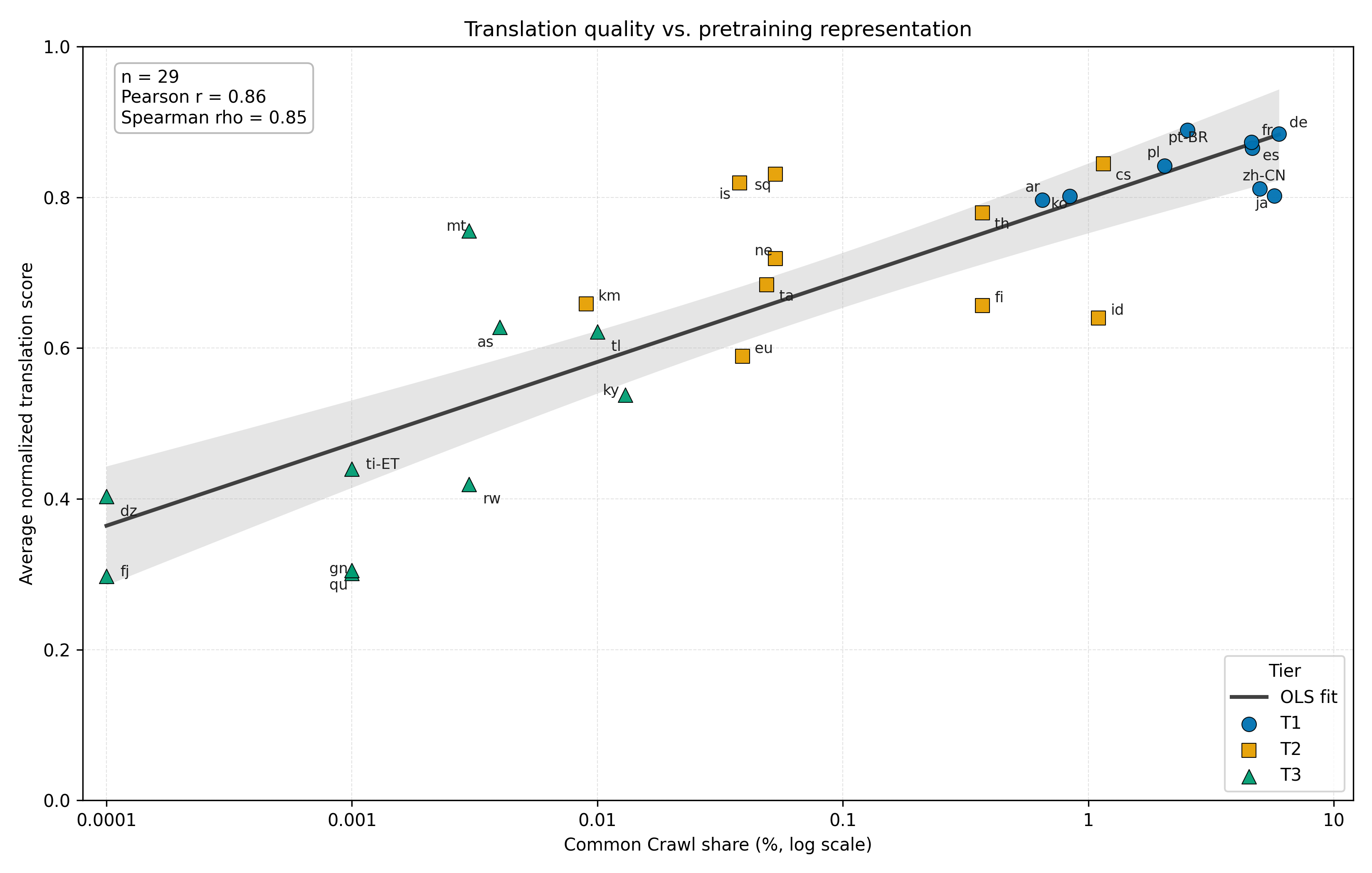}}
\caption{Average normalized round-trip translation score per target language (mean across all evaluated models) plotted against the language's share of Common Crawl on a log scale, a proxy for pretraining representation. Each point is one of the 29 target languages, colored and shaped by resource tier (§3.1); the line is an ordinary least squares fit with a 95\% confidence band. Translation quality is strongly associated with pretraining representation (Pearson r = 0.86, Spearman \(\rho\) = 0.85, n = 29). Common Crawl share is a coarse proxy and likely understates non-Latin-script languages, so the figure should be read as a strong association rather than an exact predictive relationship.}
\end{figure}

Because the translation task uses the same source sentences for every language (§3.3.7), these scores are directly comparable, and the gap reflects how well models handle each language rather than differences in test content. This result gives empirical weight to our decision to span the full resource spectrum (§3.1), and it is consistent with prior work on performance inequality across languages (Blasi et al. 2022; Joshi et al. 2020) and with the low-resource collapse reported concurrently for round-trip translation by Skorobogat et al. (2026).

A few languages sit well above the trend line, most notably Maltese, which scores around 0.74 despite having one of the lowest Common Crawl shares in our set. We read these cases as a reminder that raw web volume is only one component of a language's effective representation, since curated parallel data and sustained institutional investment, as in the case of an official EU language like Maltese, can lift translation quality above what web presence alone would suggest.

\subsection{5.4.2 Reasoning helps most tasks, but degrades some models' grammar judgments}\label{reasoning-helps-most-tasks-but-degrades-some-models-grammar-judgments}

Models can be run with or without an explicit reasoning or thinking mode, and several also expose a reasoning-effort setting. Across our set, enabling reasoning improves performance on both tasks for the large majority of models. On translation it helps universally: all 27 of the reasoning-adding comparisons show a gain (mean change +0.05 normalized), and on grammar it helps most models as well (Figure 2). Where a model exposes graded reasoning effort, more effort tends to help monotonically on the grammar task: GPT-5.4 rises from an MCC of -0.16 without reasoning to +0.03 at medium and +0.07 at high, and GPT-5.5 rises from -0.07 to +0.16 to +0.18. The minimal-effort variants confirm the direction, since reducing Gemini 3.5 Flash to minimal reasoning lowers its grammar score from 0.30 to 0.19.

\begin{figure}
\centering
\pandocbounded{\includegraphics[keepaspectratio,alt={Each of the 27 reasoning-adding configurations (n = 27) compared against its non-reasoning base of the same model, with axes showing the change in grammar MCC and hard-subset translation score (reasoning minus base). The two reasoning-reducing minimal-effort variants (Gemini 3 Flash, Gemini 3.5 Flash) are omitted, not sign-flipped, and are discussed separately in the text. All 27 configurations improve translation (all points above y = 0), while a small subset degrades grammar (points left of x = 0), and none hurt both tasks. The dissociation cases with grammar losses align with worse binary judgments rather than parsing failures.}]{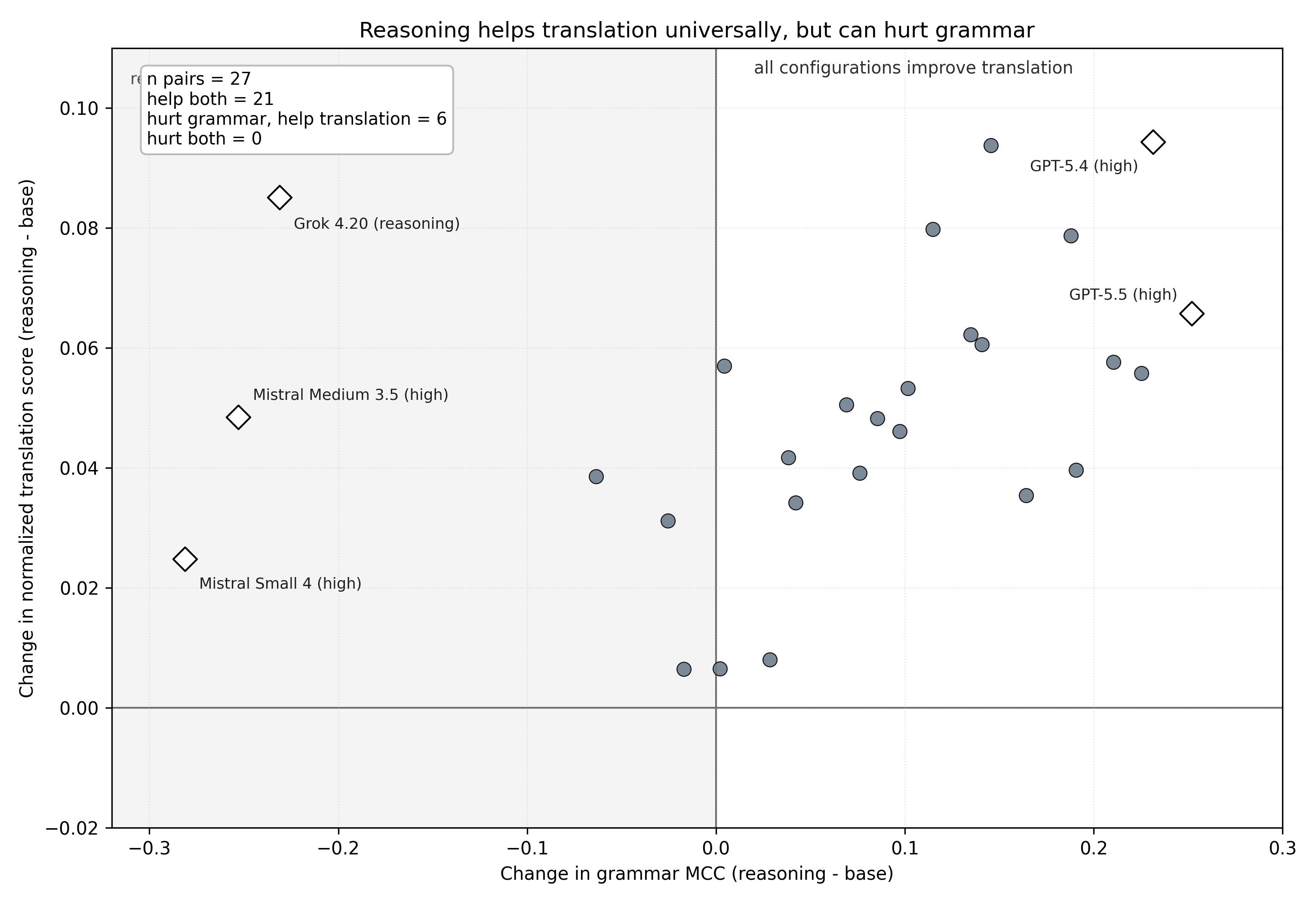}}
\caption{Each of the 27 reasoning-adding configurations (n = 27) compared against its non-reasoning base of the same model, with axes showing the change in grammar MCC and hard-subset translation score (reasoning minus base). The two reasoning-reducing minimal-effort variants (Gemini 3 Flash, Gemini 3.5 Flash) are omitted, not sign-flipped, and are discussed separately in the text. All 27 configurations improve translation (all points above y = 0), while a small subset degrades grammar (points left of x = 0), and none hurt both tasks. The dissociation cases with grammar losses align with worse binary judgments rather than parsing failures.}
\end{figure}

Against this backdrop, a small number of models lose sharply on the grammar task when reasoning is enabled: the MCC falls by 0.28 for Mistral Small 4, by 0.25 for Mistral Medium 3.5, and by 0.23 for Grok 4.20. This loss is a genuine degradation of the binary judgment rather than a scoring or parsing artifact. For Grok 4.20, no responses are unparseable, and although reasoning makes 35.5\% of its answers fail to lead with a clean yes or no (against zero percent for its non-reasoning variant), our fallback classifier (§3.2.4.1) recovers the intended answer in every such case, so the answers are scored and are simply wrong more often: accuracy falls from 0.60 to 0.49 and the MCC from 0.20 to near zero. The Mistral models keep low instruction-miss rates, so their loss is likewise a matter of worse judgments. The decisive observation is that these same reasoning variants improve on translation, where Grok 4.20 gains 0.085 and the two Mistral models gain 0.025 and 0.048. Reasoning therefore did not damage these models' multilingual ability; it specifically degraded their performance on the constrained binary error-detection judgment, while helping the generative translation task.

The practical lesson is that the reasoning setting should be chosen per task. Reasoning is broadly beneficial for multilingual meaning preservation, and for capable models it also helps grammatical error detection, but for some models it degrades the yes-or-no acceptability judgment even as it improves their translation. A model's reasoning benefit on one task therefore does not transfer to another. Our result echoes a finding reported concurrently by Skorobogat et al. (2026), who observe that reasoning variants do not reliably improve, and sometimes worsen, performance on real multilingual tasks even as they score higher on conventional reasoning and knowledge benchmarks.

\subsection{5.4.3 Cost buys translation quality, but not grammar performance}\label{cost-buys-translation-quality-but-not-grammar-performance}

The price of running a model varies enormously in our set, by more than four orders of magnitude across configurations, from about 0.02 to 240 US dollars for a full grammar run. How much that spending buys depends sharply on the task. For translation, score rises clearly with cost: the rank correlation between a configuration's normalized translation score and the total cost of its run is 0.71, and the relationship is essentially the same when cost is measured as output price per million tokens (0.70). For grammar the relationship is far weaker (Spearman 0.44, 95\% CI {[}0.21, 0.62{]}, against 0.71 {[}0.56, 0.82{]} for translation; both CIs resample base models rather than configurations\footnote{The two correlations are computed over the same 82 configurations; a paired bootstrap that resamples base models rather than configurations (5,000 resamples), so that multiple configurations of one model are not treated as independent, gives a difference of Δ\(\rho\) = 0.28 (95\% CI {[}0.13, 0.44{]}), excluding zero.}; 0.49 when each model is taken at its best configuration), and inexpensive models sit high on the frontier: Gemini 3.5 Flash reaches an MCC of 0.30 for roughly 21 US dollars, ahead of GPT-5.5 with high reasoning effort at 0.18 for close to 39 US dollars. The extremes make the split vivid. On grammar, the most expensive configuration, Claude Opus 4.6 with adaptive reasoning at about 240 US dollars, reaches an MCC of 0.17, no better than Gemini 3.5 Flash run at minimal reasoning (0.19 for around 23 cents); on translation, the most expensive configuration, Mistral Medium 3.5 with high reasoning at about 141 US dollars, sits mid-pack at 0.56 (Figure 3).

\begin{figure}
\centering
\pandocbounded{\includegraphics[keepaspectratio,alt={Performance versus total run cost for grammar (left, full item set at temperature 0, MCC) and translation (right, hard subset, normalized score), on log cost axes. Each point is one evaluated configuration (n = 82 per task). Filled circles are base configurations; open triangles are reasoning or adaptive configurations. Lines are ordinary least squares (OLS) fits with 95\% confidence bands. Total run cost is not comparable across the two panels, because the tasks differ in token volume and item count; compare the strength of each relationship, not absolute cost.}]{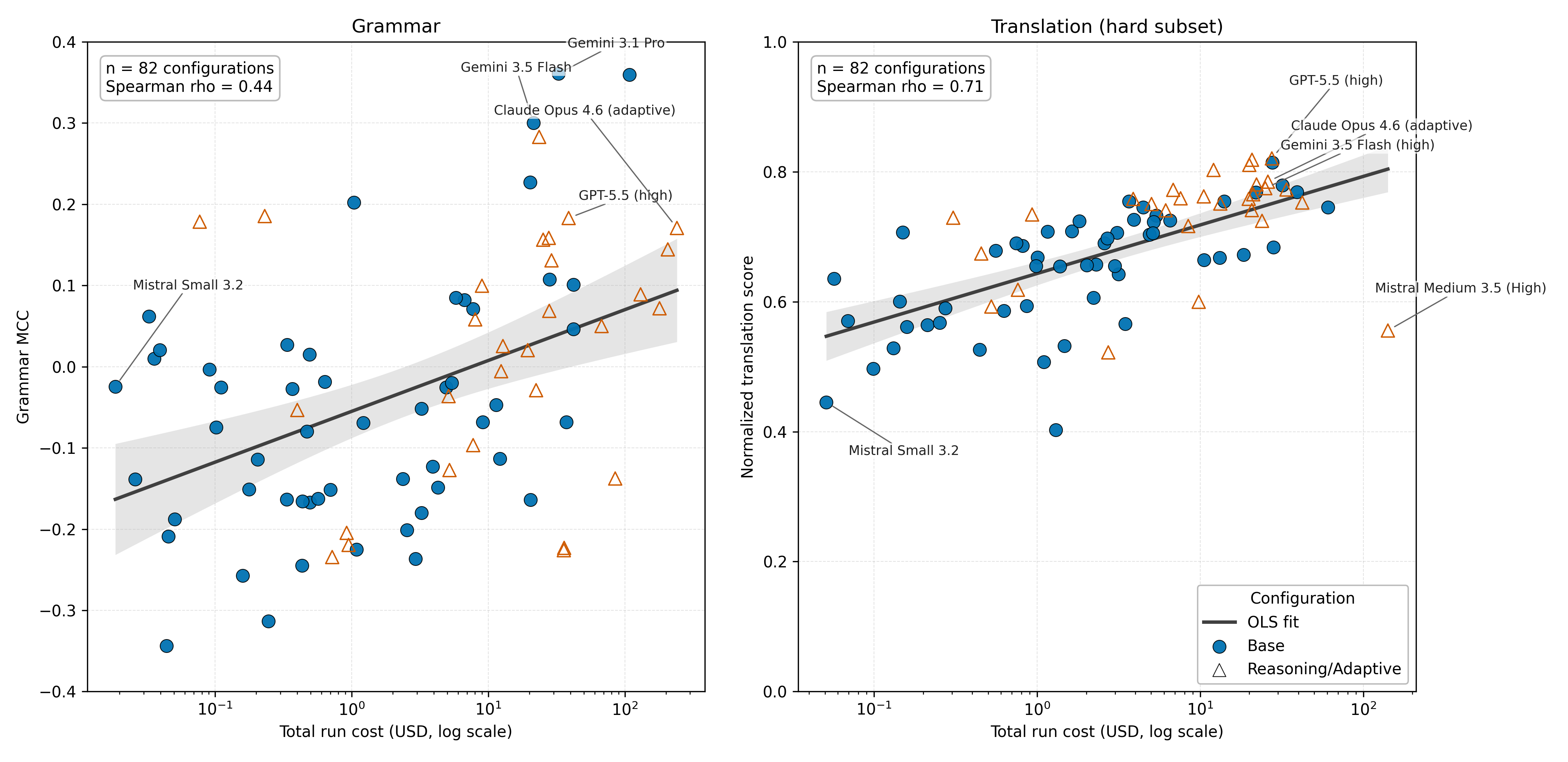}}
\caption{Performance versus total run cost for grammar (left, full item set at temperature 0, MCC) and translation (right, hard subset, normalized score), on log cost axes. Each point is one evaluated configuration (n = 82 per task). Filled circles are base configurations; open triangles are reasoning or adaptive configurations. Lines are ordinary least squares (OLS) fits with 95\% confidence bands. Total run cost is not comparable across the two panels, because the tasks differ in token volume and item count; compare the strength of each relationship, not absolute cost.}
\end{figure}

The practical implication is that the most expensive frontier configurations are justified for translation work, where they lead the field, but are difficult to justify for grammatical error detection, where a cheap model is often as good or better. This gap is widened by reasoning, which on the grammar task inflates cost through extra output tokens without a matching gain for many models: on identical items, Gemini 3.5 Flash's default configuration emits roughly 790 times the output tokens of its minimal-reasoning configuration, and GPT-5.5 at high effort roughly 84 times its no-reasoning baseline. For some models the extra spend buys an outright decline, since reasoning degrades their grammar judgment even as it raises their cost (§5.4.2); the configurations clustered at the high-cost, middling-MCC corner of the grammar results are largely the adaptive and thinking variants.

\subsection{5.4.4 Low-resource languages are improving quickly}\label{low-resource-languages-are-improving-quickly}

Plotting translation performance against model release date shows that the gap between high- and low-resource languages narrows across successive releases. The reading is cross-sectional: each point is a different model, so the trends describe how newer models compare with older ones, not longitudinal change within any system. High-resource languages are already near the top of the scale and have little room to move, rising from 0.82 to 0.85 on average over the period covered by our models, so part of the difference in slopes is ceiling compression. The absolute levels are therefore the more telling evidence: low-resource averages rise from 0.33 to 0.51 (Figure 4), and the best recent frontier models now reach roughly 0.74 on low-resource translation, where the earliest models in our set averaged around 0.33. Because the translation task is comparable across languages, this trend can be read directly, which is not possible for the grammar task (§3.2.7).

\begin{figure}
\centering
\pandocbounded{\includegraphics[keepaspectratio,alt={Mean hard-subset translation score against model release date, grouped by language resource tier (T1, T2, T3; §3.1). Each point is one model's mean score within a tier, using the best-scoring configuration per model (a frontier view of peak capability; n = 53); lines are per-tier ordinary least squares trends with 95\% confidence bands, and the labels give each tier's change over the period. Release date is a coarse axis and the set is not balanced by provider over time, so the trends describe the field in aggregate rather than any single development path; the set contains few models released before 2025, so all trends are least certain at the left edge, most visibly for T3, which also varies more.}]{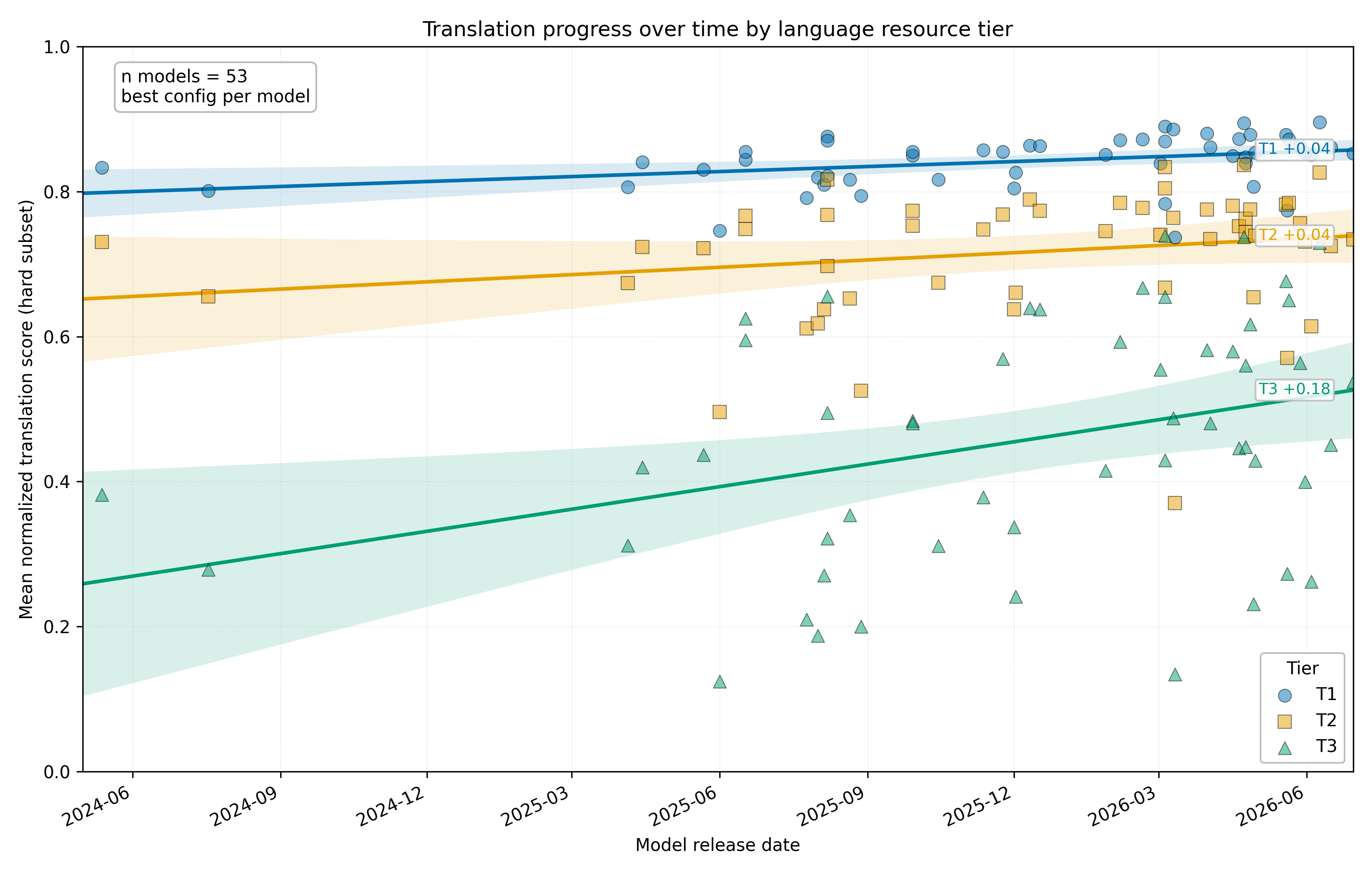}}
\caption{Mean hard-subset translation score against model release date, grouped by language resource tier (T1, T2, T3; §3.1). Each point is one model's mean score within a tier, using the best-scoring configuration per model (a frontier view of peak capability; n = 53); lines are per-tier ordinary least squares trends with 95\% confidence bands, and the labels give each tier's change over the period. Release date is a coarse axis and the set is not balanced by provider over time, so the trends describe the field in aggregate rather than any single development path; the set contains few models released before 2025, so all trends are least certain at the left edge, most visibly for T3, which also varies more.}
\end{figure}

This is the result that most directly motivates a continuously maintained benchmark with deep low-resource coverage: the frontier is moving fastest exactly where existing benchmarks have the least to say.

The same upward direction appears on the grammar task, where low-resource languages also improve across successive model releases. Because grammar items differ by language, those scores cannot be compared across languages or placed on a common axis with translation (§3.2.7); we therefore show the grammar trend within a single language instead. Figure 5 plots Nepali, the example §3.2.7 uses for the within-language reading: scores trend upward across releases while individual models scatter widely at every date.

\begin{figure}
\centering
\pandocbounded{\includegraphics[keepaspectratio,alt={Grammar MCC on Nepali against model release date, each base model shown at its best-on-Nepali configuration (the per-language maximum across that model's configurations) at temperature 0; the line is an OLS trend with a 95\% band, and the horizontal line marks chance. Because the configuration is chosen to maximize the Nepali score, this per-language selection mildly inflates the level and slope relative to fixing each model at its globally best configuration (as in Figure 4).}]{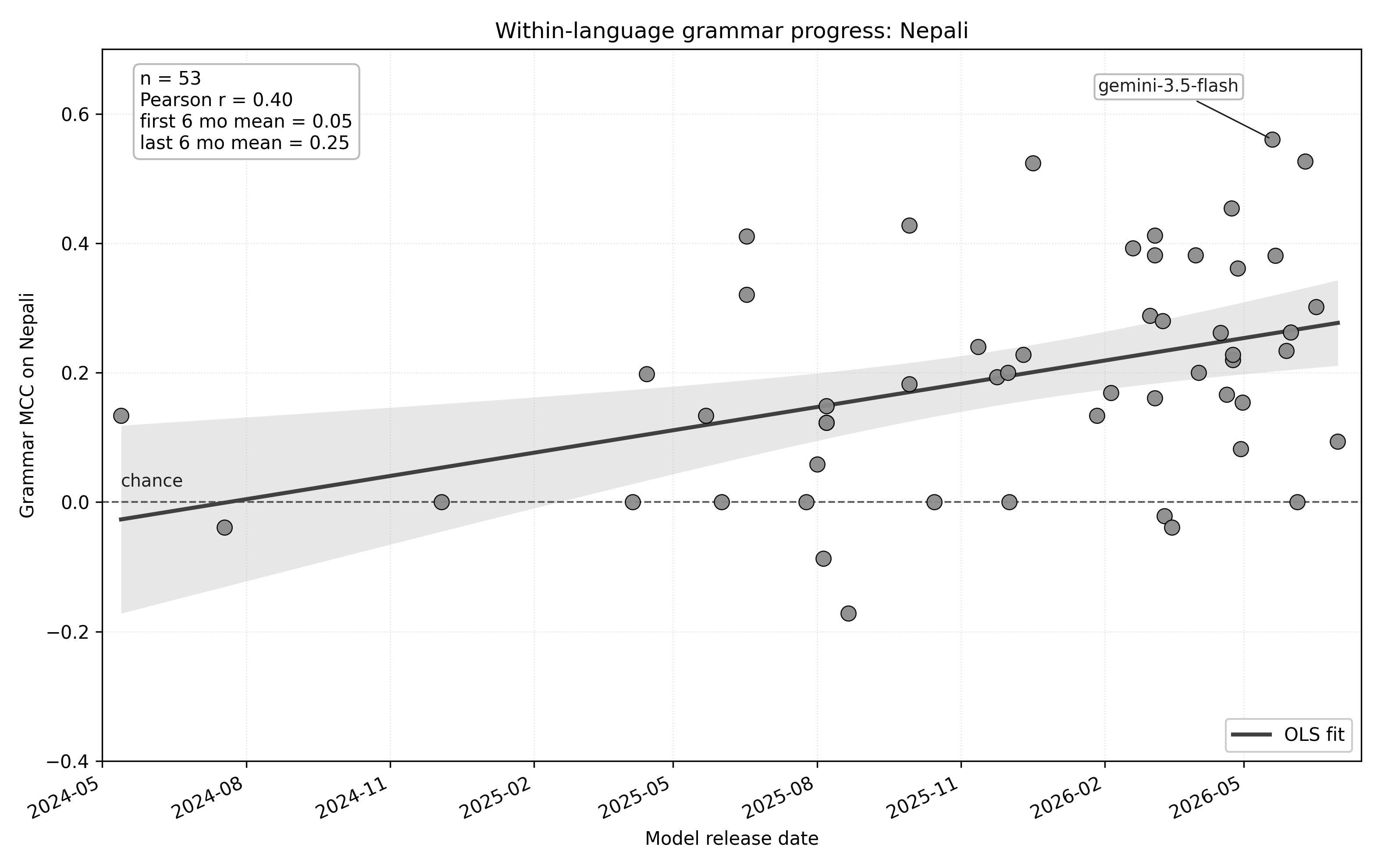}}
\caption{Grammar MCC on Nepali against model release date, each base model shown at its best-on-Nepali configuration (the per-language maximum across that model's configurations) at temperature 0; the line is an OLS trend with a 95\% band, and the horizontal line marks chance. Because the configuration is chosen to maximize the Nepali score, this per-language selection mildly inflates the level and slope relative to fixing each model at its globally best configuration (as in Figure 4).}
\end{figure}

\subsection{5.4.5 Detecting grammatical errors remains near chance for most models}\label{detecting-grammatical-errors-remains-near-chance-for-most-models}

The grammar task is hard for current models, and this is our starkest result --- one that reflects the deliberate difficulty of the items, not an inability to detect ordinary errors: the same models reach an MCC of 0.68 to 0.94 on non-adversarially-selected items (detailed below and in Appendix D.5). Even taking each model at its best configuration, 29 of the 53 models we evaluate score at or below chance on the balanced error-detection task, and the best models reach an MCC of only 0.36: Gemini 3.1 Pro and Claude Fable 5, the only two above 0.3 (Figure 6). The median model is statistically indistinguishable from chance (MCC −0.02).

\begin{figure}
\centering
\pandocbounded{\includegraphics[keepaspectratio,alt={Recall (errors correctly flagged) versus specificity (clean sentences correctly passed), one point per model at its best configuration (n = 53, temperature 0). The diagonal marks chance on the pooled confusion counts (MCC = 0), and the inset's above-chance count (21/53) is taken on those pooled counts, i.e., points above the diagonal; §5.4.5's ``at or below chance'' count instead uses each model's mean per-language MCC, so for a few models near chance the two measures disagree and the counts need not sum to 53. The shaded region marks under-flagging (better at passing clean text than catching errors), into which the field leans. Item selection is adversarial (§3.2.3), so proximity to chance reflects the targeted difficulty of these items, not models' ability to judge ordinary sentences with the more common errors found in everyday text.}]{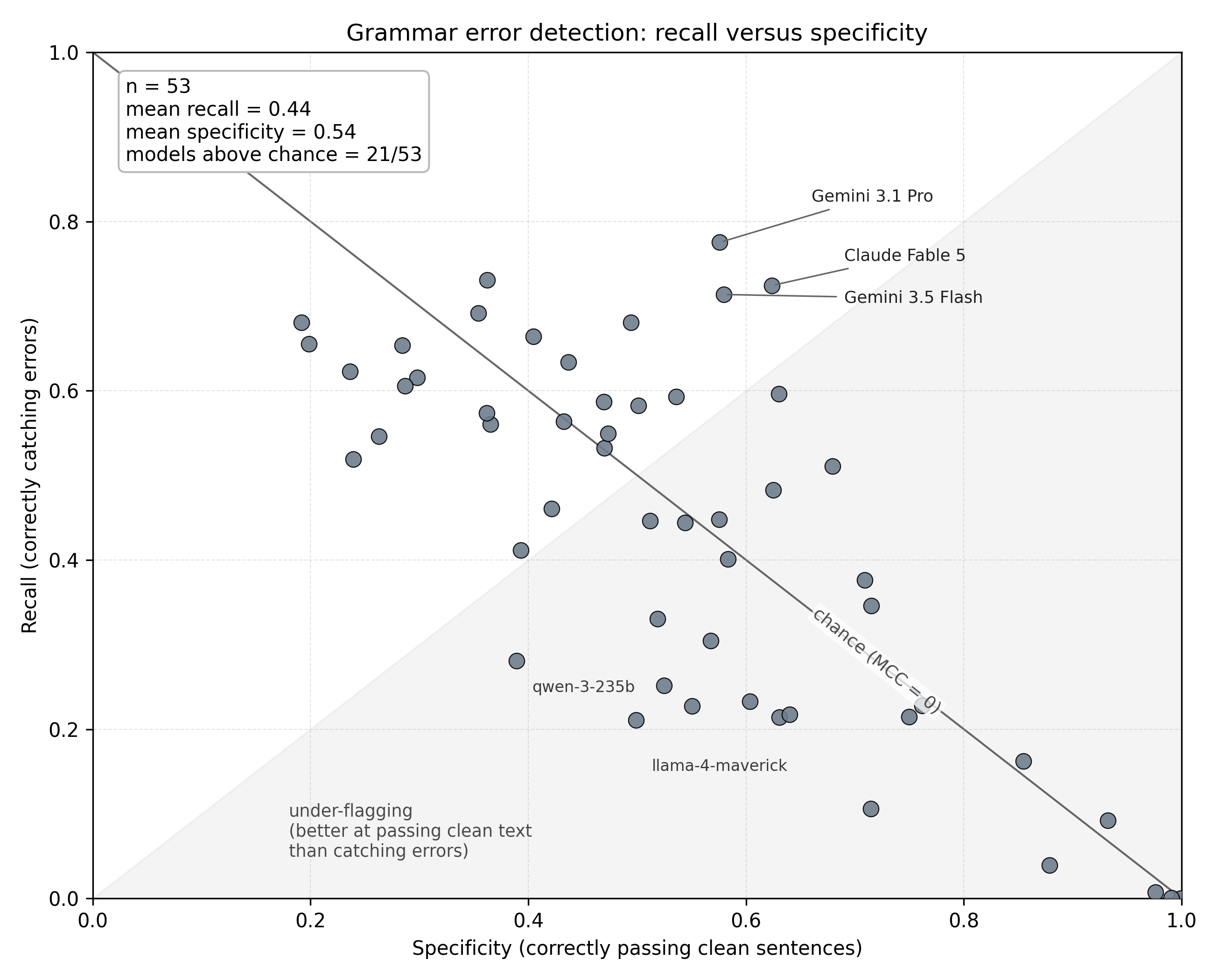}}
\caption{Recall (errors correctly flagged) versus specificity (clean sentences correctly passed), one point per model at its best configuration (n = 53, temperature 0). The diagonal marks chance on the pooled confusion counts (MCC = 0), and the inset's above-chance count (21/53) is taken on those pooled counts, i.e., points above the diagonal; §5.4.5's ``at or below chance'' count instead uses each model's mean per-language MCC, so for a few models near chance the two measures disagree and the counts need not sum to 53. The shaded region marks under-flagging (better at passing clean text than catching errors), into which the field leans. Item selection is adversarial (§3.2.3), so proximity to chance reflects the targeted difficulty of these items, not models' ability to judge ordinary sentences with the more common errors found in everyday text.}
\end{figure}

The mistakes are not random. Models systematically under-flag: pooled across the 53 best configurations, false negatives outnumber false positives by about 1.2 to one, and on a set balanced evenly between correct and erroneous sentences, models flag an error only 45\% of the time. Per-model averages show the same lean: mean recall across models is 0.44 against mean specificity of 0.54 (Figure 6). Current models therefore err toward accepting text as grammatical, and the stumper set is built to exploit exactly this tendency. The lean is not an artifact of item selection. During collection, before any selection pressure was applied, linguists found false-positive items, that is, grammatical sentences that models wrongly flag, the harder category to produce in most languages (§3.2.2), which is independent evidence that models false-alarm less readily than they miss errors.

This difficulty is by design. The items are selected to be the hardest available for each language (§3.2.3), so these scores measure performance on deliberately demanding, language-specific stumpers rather than on everyday grammaticality. As a direct check, we re-scored a sample of linguist-authored items that were validated but not adversarially selected (Appendix D.5): the same models that sit near chance on the benchmark reach an MCC of 0.68 to 0.94 on these ordinary errors (for example, Gemini 3.1 Pro rises from 0.36 to 0.94, and GPT-5.5 high from 0.18 to 0.90). The near-chance benchmark scores therefore reflect the targeted difficulty of the selected items, not an inability to detect grammatical errors in general.

Read that way, the results are informative in two constructive respects. First, the task has clear headroom and rewards genuine competence: the models that separate from the pack, led by Gemini 3.1 Pro and Claude Fable 5, show that the benchmark distinguishes real grammatical ability from surface pattern-matching and can register progress as models improve. The separation comes from the error side of the task: Gemini 3.1 Pro flags 0.78 of erroneous sentences and Claude Fable 5 0.72, against a field mean recall of 0.44, while their specificity (0.58 and 0.62) remains at or above the field mean of 0.54; their advantage lies in catching more errors rather than in a more conservative flagging threshold, and both in fact flag more often than the field, not less (Figure 6). Gemini 3.1 Pro's lead comes despite a selection headwind: it served as a tie-breaker during item selection, with ties resolved toward items it answered incorrectly (§3.2.3), so its top score is, if anything, conservative. Second, the shared under-flagging pattern gives model developers a concrete and addressable target, since current models are better at accepting correct text than at noticing the subtle, language-specific errors that native speakers catch. This also puts the earlier findings in context, since the reasoning gains and cost differences of §5.4.2 and §5.4.3 are all movements around a task with substantial room left to improve.

\section{6. Discussion}\label{discussion}

\section{6.1 Fluency is not proficiency}\label{fluency-is-not-proficiency}

The clearest lesson from our results is that producing fluent multilingual text and understanding a language well enough to judge it are different capabilities, and current models are far stronger at the former. On the translation task most models preserve meaning competently in well-resourced languages, yet on the grammar task the same models sit near chance at deciding whether a sentence is correct (§5.4.5), and they do so with a consistent direction: they lean toward accepting text as grammatical and miss genuine errors more often than they raise false alarms. A model can therefore generate and accept plausible text in a language while failing to notice the subtle, language-specific mistakes that an educated native speaker would catch.

Part of this tendency likely reflects what the models learned from. Many of our error items are built around mistakes that native speakers themselves make frequently (§3.2.1), and such mistakes are correspondingly common in the text the models were trained on. Because these errors are common in real usage, a model that has absorbed its statistical distribution will tend to accept them as normal rather than flag them. Seen this way, the grammar task probes exactly the distinction we set out to measure: whether a model has internalized the prescriptive standard of a language, or has instead learned the descriptive regularities of how the language is actually written, errors and all. The under-flagging pattern suggests that the latter often dominates.

This gap matters most where linguistic precision is the point rather than a nicety. It bears most directly on tools whose core function is to check text against a linguistic standard: automated writing assistance, grammar-checking tools, and localization quality assurance. In these, the under-flagging we observe is the failure mode that makes such a checker unreliable, since errors pass through with apparent confidence. The same accept-by-default tendency is a concern, if a less direct one, wherever a model is relied on to catch problems rather than wave text through, such as reviewing legal, medical, or regulatory content or moderating content outside English, even though grammatical correctness is only one of the qualities that matter there. Our results suggest that current models should not yet be trusted as autonomous grammatical reviewers where linguistic correctness is critical, even for languages where their generative fluency is high. That the two tasks measure different things is not incidental; it is why the benchmark includes both, and why a single ``multilingual'' score would obscure exactly the weakness that matters for careful deployment.

\section{6.2 There is no single best multilingual model}\label{there-is-no-single-best-multilingual-model}

Taken together, our findings argue against the idea of one model that is best for multilingual work. Reasoning helps generative translation universally but degrades the binary error-detection judgment for some models (§5.4.2); spending more buys clear gains on translation but little on grammar (§5.4.3); and strength on one task does not imply strength on the other. The right choice depends on the task, the output format, and the reasoning setting, none of which a single leaderboard number captures.

For practitioners, this points to a concrete discipline: evaluate a candidate model on the actual deployment task and in the actual output format, choose the reasoning setting per task rather than globally, and do not assume that a model strong at translation will also be a strong error detector, or that a more expensive configuration is worth its cost for analytic tasks. For model developers, the interaction between reasoning and the binary judgment is the more surprising signal. That extended reasoning can improve generation while making a constrained yes-or-no judgment worse suggests that current reasoning training optimizes for producing good output more than for disciplined analytic decisions, and that grammatical error detection in particular remains an underserved capability. The encouraging counterpart is that the models leading the grammar task do so by catching more errors rather than by becoming more cautious (§5.4.5), so the capability is tractable and is a clear target for targeted post-training.

\section{6.3 Lessons for building multilingual benchmarks}\label{lessons-for-building-multilingual-benchmarks}

Building the benchmark surfaced several methodological points that we think generalize. Items crafted by linguists to target language-specific phenomena expose failures that translated English test items cannot, because the hardest phenomena are precisely those absent from English. Selecting the most difficult available items per language keeps the task discriminating as models improve and, as a side effect, makes systematic model tendencies such as under-flagging visible. LLM-as-judge evaluation without per-language reference translations can be made trustworthy, but only with care: a multi-provider panel, an explicit measurement of each judge's bias toward its own provider (§3.3.4), and direct validation against professional annotators (§4) were all necessary before we were willing to rely on it. Finally, comparability has to be designed in rather than assumed, which is why grammar is reported within a language and translation across a shared item set, and why the benchmark is kept private with a living leaderboard: the same design that resists contamination is what let us observe that low-resource languages are improving fastest (§5.4.4), a trend a static, high-resource benchmark would have missed entirely.

\section{6.4 Relation to concurrent work}\label{relation-to-concurrent-work}

As noted in §2.2, the closest work to ours is the concurrent LiT benchmark of Skorobogat et al. (2026), which independently arrives at the same core idea: that translated knowledge benchmarks do not measure multilingual proficiency, and that round-trip translation offers a reference-free alternative whose judgment can be carried out in English. We read the convergence of two independent efforts on this approach as evidence of its merit. The two benchmarks also share a key practical property: judging a round-trip requires only an English-to-English comparison, and therefore does not depend on a judge more capable than the models under test.

The benchmarks differ in three respects, each following from a different emphasis. The first is granularity. LiT translates each English source serially through a fixed chain of four languages before returning to English, a design its authors adopt to stress-test cross-lingual robustness but whose serial sequences, as they note, ``obscure single-language performance.'' M-GATE instead performs a single English-to-target-to-English round-trip for each language. This forgoes the compounding stress of a chain, but in exchange it yields a score that can be attributed to one language at a time. The two designs answer different questions: LiT measures how well meaning survives a demanding multilingual relay, while ours measures how well each individual language is handled.

The second difference is in judging and validation. LiT uses a single judge model under a Multidimensional Quality Metrics penalty scheme, and validates the approach indirectly by correlating its scores with human preference ratings from LMArena across six model configurations. We instead average three judges drawn from three different providers on a 1-to-5 rubric, measure and report the residual bias each judge shows toward its own provider's outputs (§3.3.4), and validate the panel directly against professional human annotators on a stratified sample of 300 round-trips (§4). The two validation strategies are complementary rather than competing: LiT's shows that round-trip scores track real-world human preference, while ours shows that the automated panel reproduces expert human judgment of individual round-trips.

The third difference is scope. In our benchmark, round-trip translation is one of two tasks, paired with a grammatical error-detection task built from linguist-crafted, language-specific items, so that the benchmark probes both meaning preservation and grammatical competence rather than meaning preservation alone. Our language set is likewise chosen for typological diversity and for individual coverage of low-resource languages, rather than grouped by geographic region to form translation sequences, as in LiT.

\section{7. Data Release, Contamination, and Reproducibility}\label{data-release-contamination-and-reproducibility}

A benchmark of this kind faces a direct tension between two things we want from it. Its value as a measurement depends on the test items staying out of model training data, since once items are public they are liable to be absorbed into later pretraining corpora and inflate scores without reflecting genuine capability (§2.4). Science, however, depends on transparency and reproducibility, which usually means releasing the data. We resolve this by protecting only what must be protected and publishing everything else: the live test items are kept private, and the methodology, the results, and a set of illustrative examples are made public.

The methodology is specified in full in this paper: the task designs, the exact prompts (§3.2.4, §3.3.3, §3.3.4), the item construction and validation procedures (§3.2.1--3.2.3, §3.3.1--3.3.2), the corpus-hardening process (§3.3.5), the judge panel and its rubric (§3.3.4), and the human-validation protocol (§4). We plan to release a simplified version of the evaluation harness so that others can run their own material through a comparable pipeline. Results are published on a continuously updated public leaderboard (\href{https://m-gate.ai}{m-gate.ai}) covering every evaluated model and configuration, and we release a set of illustrative items drawn from sentences that were constructed but never included in the live benchmark, on the order of ten to twenty non-live grammar items per language together with illustrative translation items, so that readers can see the form and difficulty of the material without any scored item being exposed.

What we withhold is only the live test items themselves: the grammar stumper sentences for each language, the English translation source sentences, and their authored target meanings and scoring anchors. These are the assets whose leakage would compromise the benchmark, and they are the smallest set that must be protected to keep it meaningful. Because evaluating a model necessarily sends these items to its provider, we use no-training or zero-retention terms wherever a provider offers them. Not every provider does, however, so for some models the items are transmitted under standard API terms. We do not currently rotate the item set, so this is a genuine residual exposure rather than one we can claim to have eliminated, and we flag it under Limitations.

This division preserves reproducibility in the sense that matters here. Because the construction and evaluation procedures are fully specified, an independent team could build a comparable item set and reproduce our methodology; and because the judge panel is validated against professional annotators (§4), a reader can trust the reported scores without inspecting the items that produced them. The living leaderboard, with frozen and versioned subsets (§3.3.5) and a versioned judge panel (§3.3.4), lets results be compared over time and re-derived as models and panels change. We evaluate publicly available models on a rolling basis; a provider that wishes to have an unreleased model tested in advance can do so through a direct collaboration, which lets the leaderboard track the frontier without the items becoming public.

We regard the cost of this design as acceptable. Independent researchers cannot rerun our exact items to replicate a specific score, and must instead rely on the leaderboard, the published methodology, the illustrative examples, and the human-validation evidence. We take this to be the right trade-off for a benchmark intended to stay meaningful as models improve, and we discuss its limits further under Limitations.

\section{8. Conclusion}\label{conclusion}

Multilingual language models are now deployed across a hundred or more languages, yet the benchmarks used to judge them largely measure whether a model can carry out a task in a language, not whether it commands the language itself. M-GATE targets that gap. It measures linguistic proficiency through two complementary tasks: grammatical error detection on linguist-crafted items that turn on phenomena absent from English, and round-trip translation scored by an English-to-English comparison that needs no reference translations. The benchmark spans 30 typologically diverse languages reaching well into the low-resource range, and is built to stay discriminating and contamination-resistant, with adversarially selected items kept private behind a public leaderboard and a judge panel validated against professional annotators.

The results show that fluency and proficiency come apart. Models that translate competently sit near chance at detecting grammatical errors, and grammatical and translation ability correlate only moderately across models. Translation quality tracks pretraining representation closely, so the lowest-resource languages lose the most meaning, though that gap is closing quickly as new models appear. Enabling reasoning reliably improves translation, while its effect on error detection is smaller and for some models negative; spending more buys clear gains on translation but little on grammar, where inexpensive models are often as good or better. Taken together, these point to grammatical error detection, especially in lower-resource languages, as a specific and underserved capability that current training improves only slowly.

Several directions follow, though not all are straightforward. Extending the benchmark to further facets of proficiency is attractive, but our own construction experience is a caution here: many candidate phenomena we tried proved too easy or failed to separate models, so identifying new tasks that yield real signal is part of the difficulty rather than a matter of simply adding more. Within the existing tasks, the grammar task could be extended to ask not only whether an error is present but whether the model identifies the intended one, though scoring that reliably is hard: it would require judging explanations in the target language, precisely where automated judges are weakest (§3.2.6), and may need human linguists. A human performance baseline on the tasks would separately help calibrate how far models remain from expert competence.

Beyond new task types, the living design also leaves room for the benchmark itself to grow along its existing dimensions: future versions may add languages that meet the inclusion criteria of §3.1, or expand the per-language item sets to tighten statistical resolution and refresh the hard subset. We have no fixed roadmap for such extensions, but the versioned hard subset (§3.3.5) and judge panel (§3.3.4) were designed precisely so that the item set and task suite can evolve without breaking comparability with previously published results.

\section*{Limitations}\label{limitations}

Our test items are private (§7). This is a deliberate choice to resist contamination, but it limits independent reproducibility: others cannot rerun our exact items to replicate a specific score, and must rely on the published methodology, the leaderboard, the illustrative examples, and the human-validation evidence. Because evaluation necessarily sends items to model providers, and not all providers offer terms that exclude inputs from training, the items are exposed for some models, and because we do not rotate the set, that exposure accumulates over time. Two features limit how much it can inflate scores, however. First, only inputs reach the evaluated model: grammar items are sent without their labels, translation uses no reference translations, and the meaning anchors used by the judges never reach the model under test. Second, the grammar set is balanced and not ordered by label, so the sentences alone do not reveal which of them contain errors. Absorbing the items therefore does not by itself reveal the correct answers, which for these items could only be reconstructed through professional linguistic analysis. This limits the risk without eliminating it, and it remains the clearest contamination concern we cannot fully close.

Grammar scores are relative, not absolute. The stumper items are adversarially selected to be the hardest available for each language (§3.2.3), so near-chance performance reflects the difficulty of these targeted items rather than a model's general grammatical ability, and the scores should not be read as an estimate of how well a model handles ordinary text; they also measure prompted judgments, which can undersell a model's latent linguistic knowledge (§3.2.6). Because the items differ by language, grammar results are not comparable across languages (§3.2.7): grammar should be compared only within a language, while cross-language comparisons rely on the translation task, which uses the same source sentences everywhere. Grammar performance is also unstable across runs at default temperature (§5.2), which is why we headline the temperature-0 score.

Round-trip translation is an indirect measure. It captures whether meaning survives a round trip, not the quality of the intermediate translation, and it can in principle be passed by compensating errors on the two legs or by luck; it also cannot separate comprehension failures from generation failures (§3.3.6). The benchmark is English-anchored, measuring English to target and back rather than the full translation matrix, so it says little about translation between two non-English languages.

Automated judging is validated but imperfect. The translation judges agree with professional annotators about as closely as the annotators agree with one another, and are in fact more internally consistent than the human panel (§4); agreement is high but not perfect, bounded by the irreducible subjectivity of meaning-preservation scoring, on which even expert annotators do not fully agree. We therefore treat the panel as a validated stand-in for human judgment rather than as ground truth. The panel also shows a small residual bias toward each provider's own outputs (§3.3.4), and the human validation covers twelve representative languages rather than all twenty-nine. By design the annotators, like the judges, score only the English round-trip and are blind to the intermediate language; an annotator who knew that language might, for instance, recognize a calque or a subtle intermediate-language artifact that reads as fluent English, so our validation speaks to English-to-English meaning agreement rather than to detecting such artifacts. Finally, the translation source sentences are model-generated and then expert-reviewed (§3.3.1), unlike the grammar items, which were authored by linguists (LLM brainstorming was permitted but, per feedback from the linguists, rarely used, and every item was linguist-verified and cross-validated).

Scale and coverage are bounded. Each task uses 100 items per language, chosen for diagnostic value rather than size; we report confidence intervals for the headline correlations and per-model intervals on the leaderboard, and per-model grammar differences smaller than roughly 0.05 MCC are within sampling noise. The benchmark covers 30 languages, and our inclusion criteria (§3.1) bias the set toward languages with a settled, institutionally supported written standard, excluding languages with contested standards, high diglossia, or unsettled orthography, some of them widely spoken. Common Crawl share, our proxy for pretraining representation, is coarse and undercounts non-Latin scripts (§5.4.1).

Finally, results are a snapshot. Model capabilities change with each release, so any single set of scores dates quickly; we mitigate this by maintaining a living leaderboard with versioned subsets rather than reporting a one-time result.

\section*{Ethics Statement}\label{ethics-statement}

\textbf{Human contributors.} The benchmark relies substantially on professional linguists, both to construct and cross-validate the grammar items and to carry out the human-validation study. All were engaged through our established supplier network as professional language experts and were compensated at market rates, which we audit and update regularly. Where the actual workload exceeded our initial estimates, we increased payment to cover the additional time spent. Everyone involved took part with full knowledge of the project's purpose and of how their contributions would be used.

\textbf{Data.} The benchmark contains no personal or sensitive data; all items are purpose-built linguistic examples, authored or curated for evaluation rather than drawn from individuals' writing. To resist contamination, we keep the live items private and release only illustrative, non-live examples (§7).

\textbf{Intended use and impact.} M-GATE is intended to make the multilingual proficiency of language models more visible, including for low-resource languages that existing benchmarks often omit, so that developers and deployers have a clearer basis for their decisions. We caution against two misuses of our results: treating a single score as a complete measure of a model's competence in a language, and deploying models as autonomous grammatical reviewers in settings where our results show they remain unreliable (§6). Cross-language comparisons should follow the guidance in §3.2.7 and §3.3.7.

\section*{Acknowledgements}\label{acknowledgements}

This benchmark would not exist without the professional linguists from the RWS language community who authored candidate sentences, probed them against live models, and revised them through repeated review across all 30 languages; the demanding, language-specific character of the grammar task is their work. We equally thank the professional annotators whose independent scoring of round-trip translations underpins the human validation study of §4. Both groups remain anonymous, and both were compensated at professional rates. We are grateful to Marina Pantcheva, Phoebe Liu, and Nicolò Busetto for reviewing earlier drafts of this paper and materially improving it. This work was conducted by the TrainAI team at RWS.

\section*{References}\label{references}

\protect\phantomsection\label{refs}

\appendix

\section{Appendix A: Language Tiering Methodology and Per-Language Rationale}\label{appendix-a-language-tiering-methodology-and-per-language-rationale}

This appendix details the tiering methodology summarized in §3.1.

\subsection{A.1 Signals}\label{a.1-signals}

Three independent signals were combined to assign each of the 30 languages to a resource tier.

\textbf{Ethnologue Digital Language Support (DLS)} (Simons et al. 2022) is the primary digital signal. The DLS scale measures digital tooling presence across a fixed inventory of digital tools (encoding, fonts, spellcheck, machine translation, speech, virtual assistants) on five levels: Still, Emerging, Ascending, Vital, Thriving. SIL Global re-harvests the inventory for each edition, from web data collected in the fourth quarter of the preceding year; the inventory has grown over successive editions (143 tools in the published method), so DLS levels are not directly comparable across editions.

\textbf{Ethnologue EGIDS} (Lewis and Simons 2010) is the institutional signal: a 13-level scale from 0 (International) to 10 (Extinct) capturing institutional support, intergenerational transmission, and official status, independent of digital tooling.

\textbf{Common Crawl page share} is the pretraining-data signal: the percentage of pages in crawl CC-MAIN-2026-12 identified as the language by CLD2. This is the closest available public proxy for what frontier LLMs see during pretraining. It is required because DLS alone does not discriminate within the benchmark scope: 28 of the 30 languages are DLS Vital or higher, and DLS treats ``present'' tooling and ``deep'' tooling similarly.

A secondary check against the NLP resource classes of Joshi et al. (2020) was used for sanity but treated as stale: it predates the GPT-4 era and materially under-rates several languages whose digital position has shifted since.

\subsection{A.2 Tier definitions}\label{a.2-tier-definitions}

The Ethnologue signals were weighted above the Joshi classes, and Common Crawl share was added as a third axis to discriminate within DLS Vital. The resulting cuts:

\begin{itemize}
\tightlist
\item
  \textbf{Tier 1} --- EGIDS 0--1, DLS Thriving, and dedicated investment from every major frontier lab. The functional cut is ``every frontier model gets specific post-training and evaluation for this language.''
\item
  \textbf{Tier 2} --- EGIDS 1--2, DLS Vital, and a Common Crawl share around 0.009\% or above. Meaningful pretraining presence and strong institutional support, but not part of frontier labs' dedicated post-training set.
\item
  \textbf{Tier 3} --- DLS Vital or lower with a Common Crawl share near or below the Tier 2 floor, or otherwise weak signals across multiple axes. Frontier LLM performance is expected to be more variable here.
\end{itemize}

These cuts describe the tiers rather than defining them by a strict threshold: membership follows the combined Ethnologue + Common Crawl signal, so a language whose Common Crawl share sits just above the nominal 0.009\% floor can still fall in Tier 3 when its other signals are weak (Kyrgyz; A.3, item 6). Common Crawl share is used to discriminate within DLS Vital, not as a hard boundary.

Per-language signal values (speaker counts, EGIDS, DLS, and Common Crawl share) are given in Table 1 (§3.1).

\subsection{A.3 Per-language judgment calls}\label{a.3-per-language-judgment-calls}

A clean signal-driven cut produces an uneven distribution and conflicts in a few places with institutional reality. The following calls were made explicitly.

\begin{enumerate}
\def\labelenumi{\arabic{enumi}.}
\tightlist
\item
  \textbf{Korean and Arabic kept in Tier 1 despite low Common Crawl share.} Korean (0.84\%) and Arabic (0.65\%) sit below Czech (1.15\%) and Indonesian (1.10\%) on Common Crawl. They were retained in Tier 1 because every frontier lab does dedicated Korean and Arabic post-training and evaluation, and because CLD2 likely undercounts Arabic substantially due to dialect fragmentation and script handling.
\item
  \textbf{Czech and Indonesian kept in Tier 2 despite high Common Crawl share.} Both rank above Korean and Arabic on raw web volume, but neither receives the dedicated frontier-lab investment that distinguishes Tier 1. Czech and Indonesian are the strongest entries in Tier 2 by web data alone.
\item
  \textbf{Khmer promoted to Tier 2 despite Joshi Class 1.} Joshi et al. (2020) placed Khmer in the low-resource bucket. Five years later, Khmer is DLS upper-Vital, the sole national language of Cambodia, and has a monolingual web with no English-diglossia dilution. The institutional and DLS signals overrule the stale Joshi class.
\item
  \textbf{Tagalog placed in Tier 3 despite roughly 80 million speakers.} Tagalog's Common Crawl share (0.010\%) is below Khmer's because the Philippine internet is English-dominant. From a pretraining-data perspective, frontier LLMs likely see less Tagalog than Khmer, regardless of speaker count.
\item
  \textbf{Maltese placed in Tier 3 despite strong DLS.} EU tooling investment makes Maltese DLS Vital with high-quality translation pairs, but raw web volume (0.003\%) is below several Tier 3 entries. Pretraining-data weight wins over institutional polish here.
\item
  \textbf{Kyrgyz placed in Tier 3 despite a Common Crawl share above the nominal Tier 2 floor.} Kyrgyz (0.013\%) sits just above the \textasciitilde0.009\% figure that otherwise marks the low end of Tier 2, yet we place it in Tier 3. The Tier 2/3 boundary is set by the combined Ethnologue + Common Crawl signal rather than Common Crawl alone, and Kyrgyz, Tagalog (0.010\%), and Khmer (0.009\%) fall within roughly 1.5× of one another on Common Crawl share --- too close for that axis to separate them. Khmer is promoted to Tier 2 on the strength of its institutional and DLS position (item 3), while Kyrgyz and Tagalog, without comparable frontier-relevant support, stay in Tier 3. This is the boundary most sensitive to the underlying signals, and reasonable alternative cuts are defensible (§3.1).
\end{enumerate}

\subsection{A.4 Caveats}\label{a.4-caveats}

\begin{itemize}
\tightlist
\item
  DLS counts the presence of tooling, not its depth. A language at DLS Vital may still have shallow MT and weak LLM coverage.
\item
  Common Crawl share is biased toward Latin-script and English-adjacent regions. Languages with non-Latin scripts (Khmer, Tigrinya, Assamese, Dzongkha) are likely undercounted relative to their actual digital footprint.
\item
  Quechua (Cusco-Collao) and Guarani (Paraguayan) show diglossia and dialect fragmentation that may make some grammaticality judgments contested; they sit closest to the boundary of the inclusion criteria (§3.1).
\item
  Tigrinya's EGIDS depends on country: 1 (National) in Eritrea, 2 (Provincial) in the Tigray region of Ethiopia. The \texttt{ti-ET} tag targets the Ethiopian variety, hence EGIDS 2 in Table 1.
\item
  Speaker counts are L1 + L2 from Ethnologue, 27th edition, rounded, except where footnoted in Table 1.
\end{itemize}

Ethnologue-derived values were fixed at the 27th edition so that the tiering is reproducible; the 29th edition (2026) was current at the time of writing. Because DLS is re-harvested on a growing tool inventory each edition, its levels are not directly comparable across editions, and refreshing them would change the measurement rather than update it. EGIDS is stable for the benchmark's languages (all are levels 0--2). The Tier 2 / Tier 3 boundary remains the most sensitive to any future refresh.

\begin{center}\rule{0.5\linewidth}{0.5pt}\end{center}

\section{Appendix B: Evaluated Models, Versions, and Configurations}\label{appendix-b-evaluated-models-versions-and-configurations}

\subsection{B.1 Conventions}\label{b.1-conventions}

A \emph{configuration} is a model together with a reasoning setting (§3.5). Configuration identifiers carry the reasoning setting as a suffix (\texttt{-high}, \texttt{-medium}, \texttt{-minimal}, \texttt{-adaptive}, \texttt{-reasoning}). The reasoning-setting column takes these values: \texttt{base} is the model's default configuration with no reasoning parameter sent and no reasoning emitted by default (a non-reasoning model); \texttt{reasoning-default} is a model that emits reasoning by default without a parameter being set, run in that single default mode (identified by its output-token profile: hundreds to thousands of output tokens on the one-word grammar task, versus a handful for \texttt{base} models); \texttt{reasoning-none} is a model whose provider enables reasoning by default but which we evaluate with reasoning explicitly disabled (each such configuration is paired with a \texttt{reasoning-high} configuration, so the two together measure the effect of turning reasoning on); and \texttt{reasoning-minimal}, \texttt{-medium}, \texttt{-high}, \texttt{-adaptive}, and \texttt{-enabled} request reasoning at the named effort level, following each provider's own terminology. In Table 4, \texttt{base} models correspond to a ``None'' reasoning-configurations entry, \texttt{reasoning-default} models to an ``On by default'' entry, and the remaining labels to models evaluated in more than one reasoning setting. \emph{Class} is the capability class of §3.5 (Frontier / Medium / Small). \emph{Inference provider} names the serving route used for our runs: the provider's first-party API (\texttt{openai}, \texttt{google\_genai}, \texttt{mistral}), Azure AI Foundry (\texttt{azure}), Amazon Bedrock (\texttt{bedrock}), or OpenRouter (\texttt{openrouter}). Prices are US dollars per million tokens as recorded in the July 2026 reporting snapshot and apply to all configurations of a model regardless of when it was tested (§3.5). Where a model is available through multiple hosts, one endpoint was evaluated and duplicates excluded.

\subsection{B.2 Full configuration table}\label{b.2-full-configuration-table}

Table B.1 lists all 82 evaluated configurations of the 53 models, grouped by developer.

\begin{landscape}
\footnotesize
\setlength{\tabcolsep}{4pt}
{\def\LTcaptype{none} 
\begin{longtable}[]{@{}
  >{\raggedright\arraybackslash\ttfamily}p{(\linewidth - 14\tabcolsep) * \real{0.3000}}
  >{\raggedright\arraybackslash}p{(\linewidth - 14\tabcolsep) * \real{0.1500}}
  >{\raggedright\arraybackslash}p{(\linewidth - 14\tabcolsep) * \real{0.0800}}
  >{\raggedright\arraybackslash}p{(\linewidth - 14\tabcolsep) * \real{0.1200}}
  >{\raggedright\arraybackslash}p{(\linewidth - 14\tabcolsep) * \real{0.1600}}
  >{\raggedleft\arraybackslash}p{(\linewidth - 14\tabcolsep) * \real{0.0600}}
  >{\raggedleft\arraybackslash}p{(\linewidth - 14\tabcolsep) * \real{0.0600}}
  >{\raggedright\arraybackslash}p{(\linewidth - 14\tabcolsep) * \real{0.0700}}@{}}
\toprule\noalign{}
\begin{minipage}[b]{\linewidth}\raggedright
Configuration
\end{minipage} & \begin{minipage}[b]{\linewidth}\raggedright
Display name
\end{minipage} & \begin{minipage}[b]{\linewidth}\raggedright
Class
\end{minipage} & \begin{minipage}[b]{\linewidth}\raggedright
Inference provider
\end{minipage} & \begin{minipage}[b]{\linewidth}\raggedright
Reasoning
\end{minipage} & \begin{minipage}[b]{\linewidth}\raggedleft
\$/1M in
\end{minipage} & \begin{minipage}[b]{\linewidth}\raggedleft
\$/1M out
\end{minipage} & \begin{minipage}[b]{\linewidth}\raggedright
Release
\end{minipage} \\
\midrule\noalign{}
\endhead
\bottomrule\noalign{}
\endlastfoot
\texttt{gpt-4.1} & GPT-4.1 & frontier & openai & base & 2.00 & 8.00 & 2025-04-14 \\
\texttt{gpt-4o} & GPT-4o & frontier & openai & base & 2.50 & 10.00 & 2024-05-13 \\
\texttt{gpt-4o-mini} & GPT-4o Mini & medium & openai & base & 0.15 & 0.60 & 2024-07-18 \\
\texttt{gpt-5} & GPT-5 & frontier & openai & reasoning-default & 1.25 & 10.00 & 2025-08-07 \\
\texttt{gpt-5-mini} & GPT-5-Mini & medium & openai & reasoning-default & 0.25 & 2.00 & 2025-08-07 \\
\texttt{gpt-5-nano} & GPT-5-Nano & small & openai & reasoning-default & 0.05 & 0.40 & 2025-08-07 \\
\texttt{gpt-5.1} & GPT-5.1 & frontier & openai & base & 1.25 & 10.00 & 2025-11-12 \\
\texttt{gpt-5.2} & GPT-5.2 & frontier & openai & base & 1.75 & 14.00 & 2025-12-11 \\
\texttt{gpt-5.2-high} & GPT-5.2 (High) & frontier & openai & reasoning-high & 1.75 & 14.00 & 2025-12-11 \\
\texttt{gpt-5.2-medium} & GPT-5.2 (Medium) & frontier & openai & reasoning-medium & 1.75 & 14.00 & 2025-12-11 \\
\texttt{gpt-5.4} & GPT-5.4 & frontier & openai & base & 2.50 & 15.00 & 2026-03-05 \\
\texttt{gpt-5.4-high} & GPT-5.4 (High) & frontier & openai & reasoning-high & 2.50 & 15.00 & 2026-03-05 \\
\texttt{gpt-5.4-medium} & GPT-5.4 (Medium) & frontier & openai & reasoning-medium & 2.50 & 15.00 & 2026-03-05 \\
\texttt{gpt-5.4-mini} & GPT-5.4 Mini & medium & openai & base & 0.75 & 4.50 & 2026-03-05 \\
\texttt{gpt-5.4-mini-high} & GPT-5.4 Mini (High) & medium & openai & reasoning-high & 0.75 & 4.50 & 2026-03-05 \\
\texttt{gpt-5.4-mini-medium} & GPT-5.4 Mini (Medium) & medium & openai & reasoning-medium & 0.75 & 4.50 & 2026-03-05 \\
\texttt{gpt-5.4-nano} & GPT-5.4 Nano & small & openai & base & 0.20 & 1.25 & 2026-03-05 \\
\texttt{gpt-5.4-nano-high} & GPT-5.4 Nano (High) & small & openai & reasoning-high & 0.20 & 1.25 & 2026-03-05 \\
\texttt{gpt-5.4-nano-medium} & GPT-5.4 Nano (Medium) & small & openai & reasoning-medium & 0.20 & 1.25 & 2026-03-05 \\
\texttt{gpt-5.5} & GPT-5.5 & frontier & openai & base & 5.00 & 30.00 & 2026-04-23 \\
\texttt{gpt-5.5-high} & GPT-5.5 (High) & frontier & openai & reasoning-high & 5.00 & 30.00 & 2026-04-23 \\
\texttt{gpt-5.5-medium} & GPT-5.5 (Medium) & frontier & openai & reasoning-medium & 5.00 & 30.00 & 2026-04-23 \\
\texttt{gpt-oss-120b} & GPT-OSS 120B & medium & openrouter & reasoning-default & 0.04 & 0.19 & 2025-08-05 \\
\texttt{claude-fable-5} & Claude Fable 5 & frontier & azure & reasoning-default & 10.00 & 50.00 & 2026-06-09 \\
\texttt{claude-haiku-4.5} & Claude Haiku 4.5 & small & bedrock & base & 1.10 & 5.50 & 2025-10-15 \\
\texttt{claude-haiku-4.5-thinking} & Claude Haiku 4.5 (Thinking) & small & bedrock & reasoning-enabled & 1.10 & 5.50 & 2025-10-15 \\
\texttt{claude-opus-4.5} & Claude Opus 4.5 & frontier & bedrock & base & 5.00 & 25.00 & 2025-11-24 \\
\texttt{claude-opus-4.5-thinking} & Claude Opus 4.5 (Thinking) & frontier & bedrock & reasoning-enabled & 5.00 & 25.00 & 2025-11-24 \\
\texttt{claude-opus-4.6} & Claude Opus 4.6 & frontier & bedrock & base & 5.00 & 25.00 & 2026-02-05 \\
\texttt{claude-opus-4.6-adaptive} & Claude Opus 4.6 (Adaptive) & frontier & bedrock & reasoning-adaptive & 5.00 & 25.00 & 2026-02-05 \\
\texttt{claude-opus-4.6-thinking} & Claude Opus 4.6 (Thinking) & frontier & bedrock & reasoning-enabled & 5.00 & 25.00 & 2026-02-05 \\
\texttt{claude-opus-4.7} & Claude Opus 4.7 & frontier & bedrock & base & 5.00 & 25.00 & 2026-04-16 \\
\texttt{claude-opus-4.7-adaptive} & Claude Opus 4.7 (Adaptive) & frontier & bedrock & reasoning-adaptive & 5.00 & 25.00 & 2026-04-16 \\
\texttt{claude-opus-4.8} & Claude Opus 4.8 & frontier & bedrock & base & 5.00 & 25.00 & 2026-05-28 \\
\texttt{claude-opus-4.8-adaptive} & Claude Opus 4.8 (Adaptive) & frontier & bedrock & reasoning-adaptive & 5.00 & 25.00 & 2026-05-28 \\
\texttt{claude-sonnet-4} & Claude Sonnet 4 & medium & openrouter & base & 3.00 & 15.00 & 2025-05-22 \\
\texttt{claude-sonnet-4.5} & Claude Sonnet 4.5 & medium & bedrock & base & 3.30 & 16.50 & 2025-09-29 \\
\texttt{claude-sonnet-4.5-thinking} & Claude Sonnet 4.5 (Thinking) & medium & bedrock & reasoning-enabled & 3.30 & 16.50 & 2025-09-29 \\
\texttt{claude-sonnet-4.6} & Claude Sonnet 4.6 & medium & bedrock & base & 3.30 & 16.50 & 2026-02-17 \\
\texttt{claude-sonnet-4.6-adaptive} & Claude Sonnet 4.6 (Adaptive) & medium & bedrock & reasoning-adaptive & 3.30 & 16.50 & 2026-02-17 \\
\texttt{claude-sonnet-4.6-thinking} & Claude Sonnet 4.6 (Thinking) & medium & bedrock & reasoning-enabled & 3.30 & 16.50 & 2026-02-17 \\
\texttt{claude-sonnet-5} & Claude Sonnet 5 & medium & bedrock & reasoning-default & 3.30 & 16.50 & 2026-06-30 \\
\texttt{gemini-2.5-flash} & Gemini 2.5 Flash & medium & google\_genai & reasoning-default & 0.30 & 2.50 & 2025-06-17 \\
\texttt{gemini-2.5-pro} & Gemini 2.5 Pro & frontier & google\_genai & reasoning-default & 1.25 & 10.00 & 2025-06-17 \\
\texttt{gemini-3-flash-preview} & Gemini 3 Flash Preview & medium & google\_genai & reasoning-default & 0.50 & 3.00 & 2025-12-17 \\
\texttt{gemini-3-flash-preview-minimal} & Gemini 3 Flash Preview (Minimal) & medium & google\_genai & reasoning-minimal & 0.50 & 3.00 & 2025-12-17 \\
\texttt{gemini-3.1-flash-lite-preview} & Gemini 3.1 Flash Lite Preview & small & google\_genai & base & 0.25 & 1.50 & 2026-03-02 \\
\texttt{gemini-3.1-pro-preview} & Gemini 3.1 Pro Preview & frontier & google\_genai & reasoning-default & 2.00 & 12.00 & 2026-02-19 \\
\texttt{gemini-3.5-flash} & Gemini 3.5 Flash & medium & google\_genai & reasoning-default & 1.50 & 9.00 & 2026-05-19 \\
\texttt{gemini-3.5-flash-high} & Gemini 3.5 Flash (High) & medium & google\_genai & reasoning-high & 1.50 & 9.00 & 2026-05-19 \\
\texttt{gemini-3.5-flash-minimal} & Gemini 3.5 Flash (Minimal) & medium & google\_genai & reasoning-minimal & 1.50 & 9.00 & 2026-05-19 \\
\texttt{gemma-4-31b-it} & Gemma 4 31B & small & google\_genai & reasoning-default & 0.12 & 0.35 & 2026-04-02 \\
\texttt{mistral-large-3} & Mistral Large 3 & frontier & mistral & base & 0.50 & 1.50 & 2025-12-02 \\
\texttt{mistral-medium-3.1} & Mistral Medium 3.1 & medium & mistral & base & 2.70 & 8.10 & 2025-08-01 \\
\texttt{mistral-medium-3.5} & Mistral Medium 3.5 & medium & mistral & base & 1.50 & 7.50 & 2026-04-29 \\
\texttt{mistral-medium-3.5-high} & Mistral Medium 3.5 (High) & medium & mistral & reasoning-high & 1.50 & 7.50 & 2026-04-29 \\
\texttt{mistral-small-3.2} & Mistral Small 3.2 & small & mistral & base & 0.10 & 0.30 & 2025-06-01 \\
\texttt{mistral-small-4} & Mistral Small 4 & small & mistral & base & 0.15 & 0.60 & 2026-03-16 \\
\texttt{mistral-small-4-high} & Mistral Small 4 (High) & small & mistral & reasoning-high & 0.15 & 0.60 & 2026-03-16 \\
\texttt{deepseek-v3.1} & DeepSeek V3.1 & frontier & azure & base & 0.27 & 1.00 & 2025-08-21 \\
\texttt{deepseek-v3.2} & DeepSeek V3.2 & frontier & azure & base & 0.58 & 1.68 & 2025-12-01 \\
\texttt{deepseek-v4-flash} & DeepSeek V4 Flash & medium & openrouter & base & 0.14 & 0.28 & 2026-04-24 \\
\texttt{deepseek-v4-flash-high} & DeepSeek V4 Flash (High) & medium & openrouter & reasoning-high & 0.14 & 0.28 & 2026-04-24 \\
\texttt{deepseek-v4-pro} & DeepSeek V4 Pro & frontier & openrouter & reasoning-none & 1.74 & 3.48 & 2026-04-24 \\
\texttt{deepseek-v4-pro-high} & DeepSeek V4 Pro (High) & frontier & openrouter & reasoning-high & 1.74 & 3.48 & 2026-04-24 \\
\texttt{qwen-3-235b} & Qwen3 235B & frontier & bedrock & base & 0.22 & 0.88 & 2025-07-25 \\
\texttt{qwen-3.6-max-preview} & Qwen 3.6 Max Preview & frontier & openrouter & base & 1.04 & 6.24 & 2026-04-27 \\
\texttt{qwen-3.6-max-preview-high} & Qwen 3.6 Max Preview (High) & frontier & openrouter & reasoning-high & 1.04 & 6.24 & 2026-04-27 \\
\texttt{qwen-3.7-max} & Qwen 3.7 Max & frontier & openrouter & reasoning-none & 2.50 & 7.50 & 2026-05-21 \\
\texttt{qwen-3.7-max-high} & Qwen 3.7 Max (High) & frontier & openrouter & reasoning-high & 2.50 & 7.50 & 2026-05-21 \\
\texttt{grok-4.20} & Grok 4.20 & frontier & openrouter & base & 1.25 & 2.50 & 2026-03-31 \\
\texttt{grok-4.20-reasoning} & Grok 4.20 (Reasoning) & frontier & openrouter & reasoning-enabled & 1.25 & 2.50 & 2026-03-31 \\
\texttt{grok-4.3} & Grok 4.3 & frontier & openrouter & reasoning-none & 1.25 & 2.50 & 2026-04-30 \\
\texttt{llama-3.3-70b} & Llama 3.3 70B & medium & openrouter & base & 0.50 & 1.00 & 2024-12-04 \\
\texttt{llama-4-maverick} & Llama 4 Maverick & medium & azure & base & 1.41 & 0.35 & 2025-04-05 \\
\texttt{kimi-k2.5} & Kimi K2.5 & frontier & openrouter & reasoning-default & 0.45 & 2.20 & 2026-01-27 \\
\texttt{kimi-k2.6} & Kimi K2.6 & frontier & openrouter & reasoning-default & 0.95 & 4.00 & 2026-04-20 \\
\texttt{nemotron-3-super} & Nemotron 3 Super & medium & openrouter & reasoning-default & 0.10 & 0.50 & 2026-03-11 \\
\texttt{nemotron-3-ultra} & Nemotron 3 Ultra & medium & openrouter & reasoning-default & 0.50 & 2.50 & 2026-06-04 \\
\texttt{bytedance-seed-2.0-lite} & Seed 2.0 Lite & medium & openrouter & reasoning-default & 0.25 & 2.00 & 2026-03-10 \\
\texttt{glm-5.2} & GLM 5.2 & frontier & openrouter & reasoning-default & 1.40 & 4.40 & 2026-06-16 \\
\texttt{minimax-m3} & MiniMax M3 & medium & openrouter & reasoning-default & 0.30 & 1.20 & 2026-05-31 \\
\end{longtable}
}

\textbf{Table B.1:} \emph{All 82 evaluated configurations. Prices are the July 2026 snapshot; configurations of the same model share prices unless the provider bills reasoning differently.}
\end{landscape}

\subsection{B.3 Construction-panel model versions (§3.2.1)}\label{b.3-construction-panel-model-versions-3.2.1}

The five-model construction panel used in the stumper collection app resolved to the following versions:

{\def\LTcaptype{none} 
\begin{longtable}[]{@{}lll@{}}
\toprule\noalign{}
Panel model & Version string & Endpoint \\
\midrule\noalign{}
\endhead
\bottomrule\noalign{}
\endlastfoot
Gemini 3 Flash & \texttt{gemini-3-flash-preview} & google\_genai \\
GPT-4o & \texttt{gpt-4o} & openai \\
Llama 4 Maverick & \texttt{Llama-4-Maverick-17B-128E-Instruct-FP8} & azure \\
Grok 4.1 Fast & \texttt{grok-4.1-fast} & openrouter \\
DeepSeek V3.1 & \texttt{DeepSeek-V3.1} & azure \\
\end{longtable}
}

\textbf{Table B.2:} \emph{Construction-panel versions. All models run at temperature 0.}

The two held-out tie-break models of §3.2.3 were \texttt{gemini-3.1-pro-preview} (google\_genai) and \texttt{gpt-5.4} at medium reasoning effort (openai).

\subsection{B.4 Judge-panel model versions (§3.3.4)}\label{b.4-judge-panel-model-versions-3.3.4}

{\def\LTcaptype{none} 
\begin{longtable}[]{@{}lll@{}}
\toprule\noalign{}
Judge & Version string & Endpoint \\
\midrule\noalign{}
\endhead
\bottomrule\noalign{}
\endlastfoot
GPT-5.4 Mini & \texttt{gpt-5.4-mini} & openai \\
Claude Haiku 4.5 & \texttt{eu.anthropic.claude-haiku-4-5-20251001-v1:0} & bedrock \\
Gemini 3.1 Flash Lite & \texttt{gemini-3.1-flash-lite-preview} & google\_genai \\
\end{longtable}
}

\textbf{Table B.3:} \emph{Judge-panel versions. All judges run at temperature 0.}

The fallback classifier for grammar-task response parsing (§3.2.4.1) is the same Claude Haiku 4.5 version.

\begin{center}\rule{0.5\linewidth}{0.5pt}\end{center}

\section{Appendix C: Stumper Dataset Collection Instructions}\label{appendix-c-stumper-dataset-collection-instructions}

\begin{Shaded}
\begin{Highlighting}[]
\NormalTok{\# M{-}GATE Sentence Creation Guide}

\NormalTok{\# What is this project?}

\NormalTok{M{-}GATE is an automated multilingual benchmark for evaluating LLM linguistic proficiency. Your task is to create sentences containing grammatical errors that LLMs fail to detect.}

\NormalTok{Because LLMs are statistical machines, they can be tripped up by mistakes that are common online — even when those mistakes would be obvious to a trained linguist. We\textquotesingle{}re exploiting this weakness to expose gaps in their language capabilities.}

\NormalTok{\# Your task}

\NormalTok{Create \textasciitilde{}50 sentences in your language that: }

\NormalTok{* Contain a genuine grammatical error (in standard written form) where at least 2 LLMs are tricked into saying the sentence has no errors, or vice versa: no error is present but at least 2 LLMs claim there are errors (hard requirement)}

\NormalTok{* At least 5 sentences should also trick Gemini 3 Flash}

\NormalTok{* Provide a balanced number of sentences with and without errors (25 sentences with errors and 25 without errors)}

\NormalTok{* We will not provide payment for any additional sentences outside of the scope}

\NormalTok{You\textquotesingle{}ll submit your sentences via [this Google Form], which includes a field for explaining the error and providing the corrected version.}

\NormalTok{***}

\NormalTok{\# Requirements}

\NormalTok{\#\# The errors must be undisputable}

\NormalTok{* Every sentence must contain an error that is unambiguously wrong according to linguistic authorities in your language. (Or vice versa for sentences with no errors.)}

\NormalTok{* Avoid stylistic issues, neologisms, or anything that could be debated.}

\NormalTok{* Avoid word forms or constructions that have been accepted by authorities over time and are no longer considered errors (many take a descriptivist approach).}

\NormalTok{* Avoid rules that have changed in the last 12–18 months.}

\NormalTok{* If your language has multiple regional variants, stick to standard/universal rules where possible.}

\NormalTok{Rule of thumb: If a language expert or official grammar authority would without a doubt confirm it\textquotesingle{}s an error, it qualifies.}

\NormalTok{\#\# Variety matters}

\NormalTok{* Each specific error (instance) — such as a particular word form or verb conjugation — should appear in no more than 2 sentences.}

\NormalTok{* For example, if you find that LLMs struggle with "I" vs. "me" in sentences like "That was a surprise for Jim and I/me", don\textquotesingle{}t use this particular error more than twice.}

\NormalTok{\#\# Sentence format}

\NormalTok{* Single sentences only (though they can be complex)}

\NormalTok{* Always fill in the Error explanation field: describe what\textquotesingle{}s wrong and provide the correct version}

\NormalTok{***}

\NormalTok{\# How to test your sentences}

\NormalTok{Use the testing app to check whether your sentences trick the models:}

\NormalTok{\textless{}REDACTED URL\textgreater{}}

\NormalTok{Select your language, provide a sentence (one sentence at a time), mark whether the sentence has a mistake or not, and click the “Test All Models” button. You will see the results of five separate LLM models and the final result on the top right side. If you trick two models or more, the sentence is qualified and needs to be provided in the [Google Form].}

\NormalTok{A sentence qualifies if:}

\NormalTok{* The top{-}right result displays “Tricked x/5 models {-} Qualifies”}

\NormalTok{* It genuinely contains an error, and at least 2 models answer "No", therefore marking their answer as “Wrong” (i.e., they miss the error)}

\NormalTok{* [Example] for sentences with errors}

\NormalTok{  It does not contain an error, and at least 2 models answer “Yes”, therefore marking their answer as “Wrong” (i.e., they claim there is an error even though there isn’t any)}

\NormalTok{  * [Example] for sentences without errors}

\NormalTok{If you get an “Unparsable” error from a model, it is not considered an error of grammatical judgment and therefore does not count. We will not provide payment for any sentences that do not qualify or sentences outside of the scope.}

\NormalTok{Bonus points if your sentence tricks Gemini 3 Flash. Aim to have at least 5 sentences that trick Gemini 3 Flash, not just other models.}

\NormalTok{***}


\NormalTok{\# Tips for finding good errors}

\NormalTok{\#\# Research common mistakes}

\NormalTok{Look up articles about mistakes native speakers make in your language. These are great inspiration — but don\textquotesingle{}t copy example sentences verbatim. Create your own sentence that incorporates the same type of error.}

\NormalTok{Important: Always verify that it\textquotesingle{}s still considered an error. Authorities sometimes accept previously incorrect forms over time.}

\NormalTok{\#\# Use LLMs to generate candidates}

\NormalTok{Asking an LLM to help brainstorm can be effective. For example, using Gemini or ChatGPT with Deep Research mode:}

\NormalTok{I need you to look up common mistakes that \textless{}YOUR\_LANGUAGE\textgreater{} native speakers make (all kinds of mistakes), or known issues that LLMs have when producing \textless{}YOUR\_LANGUAGE\textgreater{} text. Then generate candidate sentences with those mistakes worked in.}

\NormalTok{The point is to create sentences with mistakes that LLMs would fail to spot. Make it challenging, but ensure these are definite mistakes according to \textless{}YOUR\_LANGUAGE\textgreater{} linguistic authorities.}

\NormalTok{Give me 30–50 candidate sentences.}

\NormalTok{Important: LLM suggestions often include sentences that are actually acceptable. Always double{-}check every candidate against authoritative sources.}

\NormalTok{\#\# Verify the model is failing for the right reason}

\NormalTok{If you suspect a model might be flagging a sentence for the wrong reason (e.g., it says there\textquotesingle{}s an error but you\textquotesingle{}re not sure it\textquotesingle{}s detecting your error):}

\NormalTok{1. Fix the real error and test the corrected sentence}

\NormalTok{2. If the model still claims there\textquotesingle{}s an error in the corrected sentence, you\textquotesingle{}ve found a false positive — a correct sentence that tricks models into thinking it\textquotesingle{}s wrong}

\NormalTok{3. You can submit this as a sentence without an error (it still counts as tricking the models)}

\NormalTok{***}

\NormalTok{\# Quick checklist before submitting}

\NormalTok{* [ ] The sentence contains a genuine, undisputable grammatical error}

\NormalTok{* [ ] You\textquotesingle{}ve verified the error against linguistic authorities}

\NormalTok{* [ ] At least 2 models fail to detect the error in the testing app}

\NormalTok{* [ ] You\textquotesingle{}ve filled in the Error explanation field (what\textquotesingle{}s wrong + correct version)}

\NormalTok{* [ ] This specific error (instance) doesn\textquotesingle{}t appear in more than 2 of your sentences}
\end{Highlighting}
\end{Shaded}

\begin{center}\rule{0.5\linewidth}{0.5pt}\end{center}

\section{Appendix D: Grammar Item Length Analysis}\label{appendix-d-grammar-item-length-analysis}

This appendix reports the per-language length distributions and the full per-model results of the length-shortcut analysis summarized in §3.2.3.

\subsection{D.1 Method}\label{d.1-method}

The length feature is the raw Unicode character count of each item. The label is the item's category (error / no error). All analyses run on the final selected benchmark items (50 error / 50 no-error per language). Because pooled analysis can mask per-language artifacts, metrics are computed independently within each language and only then summarized.

Two diagnostics quantify how much label signal length carries in principle. First, a per-language ROC-AUC using \texttt{−length} as the score (higher = ``more likely error'', following the observation that error items skew shorter). Second, an optimistic in-sample upper bound: for each language, all observed lengths are swept as thresholds for the rule ``shorter than \emph{t} ⇒ predict error,'' and the maximum \textbar MCC\textbar{} retained; threshold selection and evaluation use the same items, so this deliberately overstates what length alone could achieve.

Whether models \emph{actually} exploit the correlation is tested directly on model outputs in the four languages with the strongest length--label association (Assamese, Fijian, Quechua, Kinyarwanda). For each language, a median-length heuristic (below the language's median length ⇒ predict error) is defined; items are split by whether the heuristic agrees with the gold label; and for each model the accuracy gap, accuracy(agree) − accuracy(disagree), is computed at temperature 0. Systematically positive gaps would indicate length-riding; near-zero or mixed-sign gaps argue against it.

\subsection{D.2 Per-language length distributions}\label{d.2-per-language-length-distributions}

{\def\LTcaptype{none} 
\begin{longtable}[]{@{}
  >{\raggedright\arraybackslash}p{(\linewidth - 12\tabcolsep) * \real{0.1154}}
  >{\raggedright\arraybackslash}p{(\linewidth - 12\tabcolsep) * \real{0.1154}}
  >{\raggedleft\arraybackslash}p{(\linewidth - 12\tabcolsep) * \real{0.1538}}
  >{\raggedleft\arraybackslash}p{(\linewidth - 12\tabcolsep) * \real{0.1538}}
  >{\raggedleft\arraybackslash}p{(\linewidth - 12\tabcolsep) * \real{0.1538}}
  >{\raggedleft\arraybackslash}p{(\linewidth - 12\tabcolsep) * \real{0.1538}}
  >{\raggedleft\arraybackslash}p{(\linewidth - 12\tabcolsep) * \real{0.1538}}@{}}
\toprule\noalign{}
\begin{minipage}[b]{\linewidth}\raggedright
Language
\end{minipage} & \begin{minipage}[b]{\linewidth}\raggedright
n (error / no error)
\end{minipage} & \begin{minipage}[b]{\linewidth}\raggedleft
Median length, error
\end{minipage} & \begin{minipage}[b]{\linewidth}\raggedleft
Median length, no error
\end{minipage} & \begin{minipage}[b]{\linewidth}\raggedleft
Mean length, error
\end{minipage} & \begin{minipage}[b]{\linewidth}\raggedleft
Mean length, no error
\end{minipage} & \begin{minipage}[b]{\linewidth}\raggedleft
Length-only AUC
\end{minipage} \\
\midrule\noalign{}
\endhead
\bottomrule\noalign{}
\endlastfoot
en-US & 50/50 & 60 & 42 & 74.6 & 51.3 & 0.298 \\
de & 50/50 & 49 & 116 & 81.4 & 154.7 & 0.645 \\
ja & 50/50 & 21 & 28 & 21.9 & 28.7 & 0.691 \\
zh-CN & 50/50 & 15 & 11 & 15.6 & 13.0 & 0.367 \\
es & 50/50 & 48 & 54 & 48.8 & 54.7 & 0.561 \\
fr & 50/50 & 38 & 52 & 38.2 & 51.5 & 0.747 \\
pt-BR & 50/50 & 56 & 51 & 59.4 & 62.2 & 0.484 \\
pl & 50/50 & 38 & 34 & 38.7 & 39.8 & 0.466 \\
ko & 50/50 & 32 & 38 & 33.7 & 39.6 & 0.627 \\
ar & 50/50 & 32 & 26 & 35.4 & 30.2 & 0.385 \\
cs & 50/50 & 39 & 30 & 47.4 & 32.2 & 0.352 \\
id & 50/50 & 49 & 51 & 52.7 & 53.0 & 0.514 \\
fi & 50/50 & 52 & 81 & 54.8 & 96.5 & 0.713 \\
th & 50/50 & 32 & 44 & 33.3 & 48.3 & 0.672 \\
ne & 50/50 & 28 & 29 & 39.8 & 56.7 & 0.550 \\
sq & 50/50 & 39 & 50 & 43.6 & 85.0 & 0.576 \\
ta & 50/50 & 63 & 50 & 59.5 & 58.1 & 0.425 \\
eu & 50/50 & 30 & 28 & 32.6 & 32.7 & 0.468 \\
is & 50/50 & 34 & 46 & 38.2 & 58.1 & 0.653 \\
km & 50/50 & 33 & 25 & 34.0 & 32.0 & 0.386 \\
ky & 50/50 & 30 & 30 & 33.2 & 31.5 & 0.428 \\
tl & 50/50 & 36 & 36 & 40.4 & 41.9 & 0.530 \\
as & 50/50 & 18 & 28 & 20.0 & 29.8 & 0.846 \\
mt & 50/50 & 41 & 50 & 45.7 & 55.5 & 0.636 \\
rw & 50/50 & 24 & 54 & 25.7 & 64.4 & 0.962 \\
ti-ET & 50/50 & 20 & 22 & 20.8 & 23.2 & 0.609 \\
qu & 50/50 & 30 & 49 & 33.6 & 49.8 & 0.762 \\
gn & 50/50 & 54 & 70 & 59.1 & 76.1 & 0.607 \\
fj & 50/50 & 27 & 56 & 30.3 & 60.3 & 0.858 \\
dz & 50/50 & 25 & 26 & 25.9 & 29.1 & 0.580 \\
\end{longtable}
}

\textbf{Table D.1:} \emph{Character-length distributions by label and length-only ROC-AUC (score = −length), per language. Languages in Table 1 (tier) order. Pooled across all 3,000 items, the median length is 35 characters for error items and 39 for no-error items; the median per-language AUC is 0.57.}

The association is weak-to-moderate on average but strong in a few languages: Kinyarwanda (AUC ≈ 0.96), Fijian (0.86), Assamese (0.85), and Quechua (0.76) stand out. The optimistic best-threshold bound shows the same picture: the median per-language best \textbar MCC\textbar{} is 0.30, the maximum 0.80 (Kinyarwanda), while the pooled best \textbar MCC\textbar{} across all items is only 0.13.

\subsection{D.3 Direct exploitation test}\label{d.3-direct-exploitation-test}

Item-level agreement of the median-length heuristic with gold labels in the four target languages: Assamese 80/100, Fijian 76/100, Quechua 72/100, Kinyarwanda 90/100 (thresholds: 23.5, 36.0, 39.0, and 38.5 characters respectively).

{\def\LTcaptype{none} 
\begin{longtable}[]{@{}lr@{}}
\toprule\noalign{}
Model & Aggregate gap (agree − disagree) \\
\midrule\noalign{}
\endhead
\bottomrule\noalign{}
\endlastfoot
grok-4.1-fast & −0.096 \\
gpt-5.4-medium & −0.094 \\
qwen-3-235b & −0.054 \\
deepseek-v3.1 & −0.047 \\
gemini-3.1-pro-preview & −0.045 \\
gemini-3-flash-preview & −0.038 \\
grok-4.3 & +0.003 \\
gpt-4o & +0.049 \\
llama-4-maverick & +0.056 \\
\end{longtable}
}

\textbf{Table D.2:} \emph{Accuracy gap between heuristic-agreeing and heuristic-disagreeing items, aggregated over the four highest-association languages, per model (temperature 0). Positive values would indicate length-riding.}

Gaps cluster near zero; the largest magnitudes are negative, the opposite of length-riding, and the largest positive gaps are small. Together with the mechanism argument --- models are evaluated zero-shot on independent single-sentence requests and cannot learn a dataset-level correlation --- we conclude that the length--label correlation present in some languages' items is not exploited under zero-shot evaluation (§3.2.3).

\subsection{D.4 Leave-one-family-out selection-bias check}\label{d.4-leave-one-family-out-selection-bias-check}

Because five model families (Google, OpenAI, Meta, xAI, DeepSeek) supplied the construction panel and also appear on the leaderboard, item selection could in principle depress not only the panel models' scores (§3.2.1) but those of their same-family relatives, if grammatical blind spots run in families. We test this directly. For each panel family F, we form the \textbf{F-blind subset}: the live items that tricked at least two of the four \emph{non-F} panel members at collection time, so their qualification never consulted family F. Every configuration is then re-scored (MCC, temperature 0) on each F-blind subset. Since the blind subsets are easier than the full set by construction, the diagnostic is the differential shift Δ\_F: the mean score change (blind minus full) of family-F configurations minus the mean change of all other configurations, with a 95\% bootstrap CI over items. For families whose panel member itself appears on the leaderboard (Meta, DeepSeek), we additionally report Δ\_F with the panel member excluded from the family mean, isolating pure family transfer.

{\def\LTcaptype{none} 
\begin{longtable}[]{@{}
  >{\raggedright\arraybackslash}p{(\linewidth - 8\tabcolsep) * \real{0.2121}}
  >{\raggedleft\arraybackslash}p{(\linewidth - 8\tabcolsep) * \real{0.1212}}
  >{\raggedleft\arraybackslash}p{(\linewidth - 8\tabcolsep) * \real{0.1212}}
  >{\raggedright\arraybackslash}p{(\linewidth - 8\tabcolsep) * \real{0.2727}}
  >{\raggedright\arraybackslash}p{(\linewidth - 8\tabcolsep) * \real{0.2727}}@{}}
\toprule\noalign{}
\begin{minipage}[b]{\linewidth}\raggedright
Family (panel member)
\end{minipage} & \begin{minipage}[b]{\linewidth}\raggedleft
Items retained (/3,000)
\end{minipage} & \begin{minipage}[b]{\linewidth}\raggedleft
Error / no-error
\end{minipage} & \begin{minipage}[b]{\linewidth}\raggedright
Δ\_F (95\% CI)
\end{minipage} & \begin{minipage}[b]{\linewidth}\raggedright
Δ\_F excl. panel member (95\% CI)
\end{minipage} \\
\midrule\noalign{}
\endhead
\bottomrule\noalign{}
\endlastfoot
Google (Gemini 3 Flash) & 2,405 & 1,379/1,026 & +0.022 {[}+0.013, +0.032{]} & --- \\
OpenAI (GPT-4o) & 2,298 & 1,308/990 & −0.014 {[}−0.021, −0.006{]} & --- \\
Meta (Llama 4 Maverick) & 2,196 & 1,262/934 & +0.113 {[}+0.094, +0.133{]} & +0.116 {[}+0.090, +0.144{]} \\
xAI (Grok 4.1 Fast†) & 2,299 & 1,382/917 & +0.068 {[}+0.060, +0.077{]} & --- \\
DeepSeek (V3.1) & 2,198 & 1,235/963 & +0.069 {[}+0.058, +0.079{]} & +0.058 {[}+0.047, +0.068{]} \\
\end{longtable}
}

\textbf{Table D.3:} \emph{Leave-one-family-out differential shift in grammar MCC. †Grok 4.1 Fast is not itself on the leaderboard, so the xAI effect is entirely family transfer to its relatives (Grok 4.20, Grok 4.3).}

Two conclusions follow. First, for the families that headline the leaderboard the selection bias is immaterial: OpenAI configurations do not improve on items chosen without consulting GPT-4o (Δ = −0.014), and the Google effect (+0.022) is below the per-model sampling noise of ±0.03--0.04 (§5.1) --- so the low grammar scores of strong GPT configurations, and the high scores of Gemini ones, are not artifacts of the panel's composition. Second, a real family-transfer effect exists for the panel's own families: Meta (+0.12), DeepSeek (+0.07), and xAI (+0.07) configurations gain materially on their family-blind subsets, and the effect persists when the panel member itself is excluded. Grammar scores of construction-panel models \emph{and their same-family relatives} should therefore be read with this selection bias in mind, extending the caution of §3.2.1.

Limitations: this analysis restricts the realized live item set rather than simulating a counterfactual selection over the full candidate pool, and tie-breaking exposure (Gemini 3.1 Pro Preview, GPT-5.4; §3.2.3) cannot be removed by the family-blind restriction. It was run against the live database (all published configurations at the time of analysis).

\subsection{D.5 Difficulty reference: performance on non-adversarial items}\label{d.5-difficulty-reference-performance-on-non-adversarial-items}

The headline grammar scores (§5.4.5) are measured on items adversarially selected to be the hardest available per language (§3.2.3), which raises the question of how much of the near-chance result is intrinsic difficulty of grammatical error detection versus an artifact of that selection. This subsection provides a reference point with two checks that share no selection dependence with the headline number.

\textbf{Selection gradient.} Items were selected because they tricked at least two of the five construction-panel models at collection time (§3.2.1). Grouping the full validated candidate pool by how many panel models each item tricked shows a monotonic selection pressure: 9\% of items that tricked no panel model were selected, rising through 33\% (one model), 62\% (two), 81\% (three), to 95\% (four or five). Restricting to the 3,000 live items and grouping by the same trick count, leaderboard models' MCC falls steeply with item difficulty. The field-median MCC is +0.19 on the two-trick stratum (the easiest items admitted to the benchmark), 0.00 on the three-trick stratum, and −0.35 on the four-to-five-trick stratum; the top configurations show the same gradient (e.g., Gemini 3.1 Pro 0.51 / 0.44 / −0.03, Claude Fable 5 0.53 / 0.44 / −0.05). The strata are not label-balanced (the harder strata skew toward error items), so these values are read as a gradient rather than as calibrated accuracies.

\textbf{Typical-error calibration.} For a reference set with no adversarial selection at all, we sampled 541 linguist-authored items across all 30 languages that were validated but not selected for the benchmark, restricted to items that tricked at most one panel model and had full five-model panel coverage (balanced toward 10 error / 10 no-error per language where the pool allowed; live-benchmark sentences excluded). Five configurations were re-scored on this set at temperature 0 using the production prompt, parsing, and fallback (§3.2.4--§3.2.4.1). These are non-live items and are not part of the scored benchmark.

{\def\LTcaptype{none} 
\begin{longtable}[]{@{}lrr@{}}
\toprule\noalign{}
Configuration & Benchmark MCC & Calibration MCC (typical errors) \\
\midrule\noalign{}
\endhead
\bottomrule\noalign{}
\endlastfoot
Gemini 3.1 Pro Preview & 0.36 & 0.94 \\
Claude Fable 5 & 0.36 & 0.89 \\
Gemini 3.5 Flash & 0.30 & 0.91 \\
GPT-5.5 High & 0.18 & 0.90 \\
GPT-5.4 & −0.16 & 0.68 \\
\end{longtable}
}

\textbf{Table D.4:} \emph{Grammar MCC on the adversarially selected benchmark versus a non-selected typical-error calibration set (541 items, temperature 0). Benchmark MCC is the frozen-snapshot value (identical to §5.4.5 and Appendix J.1); calibration MCC is from the separate typical-error run. Fixed sampling seed; per-language sampling audit and per-item results accompany the analysis.}

On ordinary, non-selected errors the same models that sit near chance on the benchmark detect grammatical errors reliably (MCC 0.68--0.94), and the gap is largest for the models with the lowest benchmark scores (GPT-5.4, +0.84). Together with the selection gradient, this locates the near-chance benchmark result in the targeted difficulty of the selected items, not in a general inability to detect grammatical errors, and confirms that the benchmark measures the demanding tail rather than everyday grammaticality (§5.4.5).

\begin{center}\rule{0.5\linewidth}{0.5pt}\end{center}

\section{Appendix E: Judge Self-Preference Measurement}\label{appendix-e-judge-self-preference-measurement}

This appendix details the same-maker (self-preference) analysis summarized in §3.3.4. The judge panel spans three providers (Anthropic, Google, OpenAI; Appendix B.4), all of whose models are also evaluated.

\subsection{E.1 Method}\label{e.1-method}

Two comparisons are computed over all judge calls. The \textbf{raw} comparison contrasts each judge's average score on responses produced by models of its own maker against responses from other makers; this mixes self-preference with genuine quality differences between makers' models. The \textbf{controlled} comparison removes the quality confound: for each evaluated response, the judge's score is expressed as a residual against the mean score of the other two judges on the same response, and same-maker residuals are compared with other-maker residuals. The controlled metric is the primary one. Both are computed on the full corpus and on the hard subset (hard).

\subsection{E.2 Results}\label{e.2-results}

{\def\LTcaptype{none} 
\begin{longtable}[]{@{}llrrr@{}}
\toprule\noalign{}
Subset & Judge maker & Same-maker avg & Other-maker avg & Raw diff \\
\midrule\noalign{}
\endhead
\bottomrule\noalign{}
\endlastfoot
full & Anthropic & 4.141 & 3.857 & +0.284 \\
full & Google & 4.279 & 3.925 & +0.355 \\
full & OpenAI & 4.172 & 3.991 & +0.181 \\
hard & Anthropic & 3.866 & 3.606 & +0.260 \\
hard & Google & 3.932 & 3.633 & +0.299 \\
hard & OpenAI & 3.900 & 3.728 & +0.173 \\
\end{longtable}
}

\textbf{Table E.1:} \emph{Raw same-maker gaps. These are large but mostly reflect that each maker's own models tend to be strong translators.}

{\def\LTcaptype{none} 
\begin{longtable}[]{@{}
  >{\raggedright\arraybackslash}p{(\linewidth - 10\tabcolsep) * \real{0.1364}}
  >{\raggedright\arraybackslash}p{(\linewidth - 10\tabcolsep) * \real{0.1364}}
  >{\raggedleft\arraybackslash}p{(\linewidth - 10\tabcolsep) * \real{0.1818}}
  >{\raggedleft\arraybackslash}p{(\linewidth - 10\tabcolsep) * \real{0.1818}}
  >{\raggedleft\arraybackslash}p{(\linewidth - 10\tabcolsep) * \real{0.1818}}
  >{\raggedleft\arraybackslash}p{(\linewidth - 10\tabcolsep) * \real{0.1818}}@{}}
\toprule\noalign{}
\begin{minipage}[b]{\linewidth}\raggedright
Subset
\end{minipage} & \begin{minipage}[b]{\linewidth}\raggedright
Judge maker
\end{minipage} & \begin{minipage}[b]{\linewidth}\raggedleft
Same-maker residual
\end{minipage} & \begin{minipage}[b]{\linewidth}\raggedleft
Other-maker residual
\end{minipage} & \begin{minipage}[b]{\linewidth}\raggedleft
Diff
\end{minipage} & \begin{minipage}[b]{\linewidth}\raggedleft
p
\end{minipage} \\
\midrule\noalign{}
\endhead
\bottomrule\noalign{}
\endlastfoot
full & Anthropic & −0.073 & −0.080 & +0.007 & 0.029 \\
full & Google & +0.000 & −0.027 & +0.027 & 7.8×10⁻¹⁰ \\
full & OpenAI & +0.102 & +0.103 & −0.001 & 0.68 \\
hard & Anthropic & −0.002 & 0.000 & −0.002 & 0.76 \\
hard & Google & +0.024 & 0.000 & +0.024 & 0.0011 \\
hard & OpenAI & +0.001 & 0.000 & +0.001 & 0.84 \\
\end{longtable}
}

\textbf{Table E.2:} \emph{Controlled same-maker effect (residual against the other two judges on the same response). The effect is concentrated in the Google judge; the Anthropic and OpenAI judges show negligible or non-significant self-preference after controlling for response quality.}

Pooled across judges, the controlled same-maker lift is +0.025 points on the 1--5 scale on the full set (95\% CI {[}0.021, 0.028{]}, p ≈ 5×10⁻⁴²) and +0.034 on the hard subset (95\% CI {[}0.028, 0.040{]}, p ≈ 1×10⁻³⁰). Because each response is scored by three judges of which at most one can be same-maker, this translates to roughly +0.008 (full) to +0.011 (hard) on the final three-judge mean, or +0.002--0.003 on the normalized {[}0,1{]} scale: well below the differences that separate models, whose hard-subset scores span 0.40 to 0.82 (§5.3).

{\def\LTcaptype{none} 
\begin{longtable}[]{@{}llr@{}}
\toprule\noalign{}
Subset & Response maker & Shift in 3-judge mean \\
\midrule\noalign{}
\endhead
\bottomrule\noalign{}
\endlastfoot
full & Anthropic & −0.025 \\
full & Google & +0.000 \\
full & OpenAI & +0.034 \\
hard & Anthropic & −0.018 \\
hard & Google & −0.013 \\
hard & OpenAI & +0.039 \\
\end{longtable}
}

\textbf{Table E.3:} \emph{Approximate net shift in the final three-judge average for responses from each maker, combining each judge's controlled bias. All shifts are within ±0.04 points on the 1--5 scale.}

\textbf{Model-identical case.} All three judges are themselves evaluated models (Appendix B), so the sharpest form of self-preference is a judge scoring a round-trip produced by its own exact model. Restricting to those cases and comparing each judge's residual (its score minus the mean of the other two judges on the same round-trip) against a cross-maker baseline, the model-identical lift is +0.022 for Claude Haiku 4.5, +0.031 for Gemini 3.1 Flash Lite, and +0.036 for GPT-5.4 Mini; pooled, +0.035 (95\% CI {[}0.014, 0.056{]}; n = 4,337 round-trips, hard subset). This is no larger than the provider-level same-maker effect (+0.034 on the hard subset), so even in its strongest form self-preference stays small, moving the final three-judge mean by only about +0.01, since at most one of the three judges is ever the same model as the response.

By construction, the residual comparison detects only biases that differ between judges; a bias shared by all three, such as a panel-wide preference for LLM-typical phrasing in back-translations, would cancel in the residuals and is instead bounded by the human validation of §4, where the panel mean tracks the human mean at r = 0.78 and scores 0.21 points \emph{lower} than the human annotators on average.

\begin{center}\rule{0.5\linewidth}{0.5pt}\end{center}

\section{Appendix F: Corpus Hardening Probe Details}\label{appendix-f-corpus-hardening-probe-details}

This appendix details the probe campaign behind the corpus hardening of §3.3.5. All probes ran through the full production translation-and-judging pipeline (forward translation, back-translation, three-judge scoring) without writing to the active dataset, over seven intermediate languages spanning all resource tiers: Czech, Polish, Finnish, Japanese, Simplified Chinese, Fijian, and Kinyarwanda.

\subsection{F.1 Probe rounds}\label{f.1-probe-rounds}

133 candidate items were probed across six rounds, each testing a hypothesized source of round-trip difficulty:

{\def\LTcaptype{none} 
\begin{longtable}[]{@{}
  >{\raggedright\arraybackslash}p{(\linewidth - 6\tabcolsep) * \real{0.1250}}
  >{\raggedleft\arraybackslash}p{(\linewidth - 6\tabcolsep) * \real{0.1667}}
  >{\raggedright\arraybackslash}p{(\linewidth - 6\tabcolsep) * \real{0.3333}}
  >{\raggedright\arraybackslash}p{(\linewidth - 6\tabcolsep) * \real{0.3750}}@{}}
\toprule\noalign{}
\begin{minipage}[b]{\linewidth}\raggedright
Round
\end{minipage} & \begin{minipage}[b]{\linewidth}\raggedleft
Items
\end{minipage} & \begin{minipage}[b]{\linewidth}\raggedright
Hypothesis probed
\end{minipage} & \begin{minipage}[b]{\linewidth}\raggedright
Outcome
\end{minipage} \\
\midrule\noalign{}
\endhead
\bottomrule\noalign{}
\endlastfoot
Hard candidates, round 1 & 28 & Fine-grained linguistic phenomena and documented MT failure modes (negation scope, modality, aspect, nominalization) & Mostly robust; a minority of items showed consistent loss \\
Hard candidates, round 2 & 30 & Referential precision (numbers, dates, measurements) and structural complexity under stress & Numbers/dates near ceiling; selective signal in structure \\
Paragraph-level items & 20 & Multi-sentence discourse structure; difficulty compounding with length & Little added discrimination over single sentences \\
Cross-lingual asymmetries & 35 & Lexical asymmetries between English and target languages (kinship systems, granular vocabulary distinctions) & Largely mitigated by current frontier models \\
Competing coreference & 15 & Two same-gender referents whose pronouns are resolved only by pragmatic inference & \textbf{Strongest and most consistent discriminator}; actor attribution is dropped while propositional content survives \\
Same-actor double action & 5 & Complement hypothesis: round-trips split one actor's action chain across two actors & Did not replicate; unambiguous predicate chains survive \\
\end{longtable}
}

\textbf{Table F.1:} \emph{The six probe rounds (133 items total). The competing-coreference and same-actor rounds were each run twice, once under each of two frontier target models (a Gemini frontier configuration and GPT-5.4 at medium reasoning effort), to check that the signal was not model-specific.}

Most hypothesized difficulty sources proved robust to round-trip translation under current frontier models: number and date handling, idiom preservation, quantifier scope, modal force, counterfactuals, and anaphora chains all scored near ceiling, and many failure modes documented in 2024--2025 work were largely mitigated by intervening model progress (§3.3.5).

\subsection{F.2 Promotion}\label{f.2-promotion}

22 probed items were promoted into the live corpus, replacing the 22 lowest-discrimination items; the corpus remained at 100 items. By category, the promoted items comprise 15 Discourse pragmatics, 4 Referential precision, and 3 Structural complexity items, which produces the post-hardening category distribution of §3.3.2 (in particular the growth of Discourse pragmatics to 26 items). Promoted items are live benchmark items and are not reproduced here.

\subsection{F.3 Illustrative non-promoted probe item (non-live)}\label{f.3-illustrative-non-promoted-probe-item-non-live}

For the flavor of a \emph{robust} probe: the round-1 item ``He didn't leave because he was angry.'' (negation-scope ambiguity) round-tripped with a score of 5 from every judge in every probed language --- the ambiguity survives because most target languages preserve the surface structure. Items like this were not promoted precisely because they no longer discriminate between current models.

\begin{center}\rule{0.5\linewidth}{0.5pt}\end{center}

\section{Appendix G: Tokenizer Efficiency}\label{appendix-g-tokenizer-efficiency}

\textbf{Measurement.} We measure tokenizer efficiency as the average number of characters (Unicode code points) encoded per token. For each model and language we take the grammar-task sentences, concatenate them into a single message with no additional text, send it as the request payload, read the input token count reported by the API, and divide the total character count by it. We report results as a per-character token-cost multiplier relative to English (English = 1.0): the target language's characters-per-token divided into the same tokenizer family's own English characters-per-token, so no multiplier mixes measurements from different tokenizers.

\textbf{Why concatenation.} We arrived at this design after first sending sentences individually, where some models add extra input tokens in certain configurations (for example, when reasoning is enabled) --- in some cases as many tokens as the sentence itself. Because these are a fixed per-request overhead, concatenating 100 sentences reduces their relative contribution by roughly 100×, to a low single-digit percentage. The per-request inflation is nonetheless real when sending many short requests in deployment, and practitioners should measure it for their chosen models.

\textbf{Tokenizer families.} Tokenizer efficiency clusters tightly by provider: models from the same provider almost always produce identical chars/token profiles, indicating a shared tokenizer. We group models accordingly. Table G.1 reports a selected set of scripts and languages spanning the observed range; where a provider's models show distinct chars/token profiles, the families are listed separately. Anthropic and Alibaba each changed tokenizers between model generations, with clearly different profiles; OpenAI's two rows (gpt-oss and the GPT-4o/GPT-5 line) are near-identical, consistent with variants of the same o200k vocabulary, and are kept separate only for completeness.

\textbf{Interpretation across scripts.} Within a single language, every tokenizer encodes the same sentences, so differences between tokenizer families are content-matched and directly reflect tokenizer efficiency; a lower chars-per-token value on the same text means a less efficient tokenizer, and the within-language cost comparisons in the table are valid as realized deployment cost. Across writing systems, the multiplier instead measures per-character cost, because the underlying sentences differ in content and characters carry different amounts of information in different scripts (a Han character conveys more than a Latin one; compact scripts need fewer characters for the same content). A cross-script multiplier therefore bounds but does not equal the cost of processing equivalent content: Chinese at 3.2× per character corresponds to a much smaller content-level premium once its character economy is accounted for, while for scripts that are not more compact than Latin the per-character and content-level readings roughly coincide. Estimating content-level multipliers would require parallel text. A natural candidate is the translation task's own forward translations of the shared English sources, filtered to highly rated round-trips; we may adopt this in a future version if a sufficient-quality parallel set can be assembled for every language. For now, translation quality in the lowest-resource languages (§5.4.1) makes model-translated text unreliable as a measurement corpus there, so we prefer the grammar sentences, which are linguist-validated native text in all 30 languages.

\begin{landscape}

{\def\LTcaptype{none} 
\begin{longtable}[]{@{}
  >{\raggedright\arraybackslash}p{(\linewidth - 18\tabcolsep) * \real{0.2500}}
  >{\raggedright\arraybackslash}p{(\linewidth - 18\tabcolsep) * \real{0.0833}}
  >{\raggedright\arraybackslash}p{(\linewidth - 18\tabcolsep) * \real{0.0833}}
  >{\raggedright\arraybackslash}p{(\linewidth - 18\tabcolsep) * \real{0.0833}}
  >{\raggedright\arraybackslash}p{(\linewidth - 18\tabcolsep) * \real{0.0833}}
  >{\raggedright\arraybackslash}p{(\linewidth - 18\tabcolsep) * \real{0.0833}}
  >{\raggedright\arraybackslash}p{(\linewidth - 18\tabcolsep) * \real{0.0833}}
  >{\raggedright\arraybackslash}p{(\linewidth - 18\tabcolsep) * \real{0.0833}}
  >{\raggedright\arraybackslash}p{(\linewidth - 18\tabcolsep) * \real{0.0833}}
  >{\raggedright\arraybackslash}p{(\linewidth - 18\tabcolsep) * \real{0.0833}}@{}}
\toprule\noalign{}
\begin{minipage}[b]{\linewidth}\raggedright
Tokenizer family
\end{minipage} & \begin{minipage}[b]{\linewidth}\raggedright
English chars/tok (baseline)
\end{minipage} & \begin{minipage}[b]{\linewidth}\raggedright
Tigrinya
\end{minipage} & \begin{minipage}[b]{\linewidth}\raggedright
Dzongkha
\end{minipage} & \begin{minipage}[b]{\linewidth}\raggedright
Khmer
\end{minipage} & \begin{minipage}[b]{\linewidth}\raggedright
Chinese
\end{minipage} & \begin{minipage}[b]{\linewidth}\raggedright
Japanese
\end{minipage} & \begin{minipage}[b]{\linewidth}\raggedright
Korean
\end{minipage} & \begin{minipage}[b]{\linewidth}\raggedright
Thai
\end{minipage} & \begin{minipage}[b]{\linewidth}\raggedright
German
\end{minipage} \\
\midrule\noalign{}
\endhead
\bottomrule\noalign{}
\endlastfoot
Google: Gemini, Gemma & 4.4 & 3.4 & 3.3 & 2.2 & 3.2 & 2.8 & 2.7 & 1.7 & 1.1 \\
Meta: Llama 4 & 4.7 & 4.3 & 5.1 & 2.6 & 3.3 & 3.1 & 2.7 & 1.9 & 1.1 \\
OpenAI: gpt-oss & 4.6 & 8.2 & 6.9 & 2.9 & 3.8 & 3.7 & 3.0 & 2.2 & 1.1 \\
OpenAI: GPT-4o / GPT-5 (o200k) & 4.8 & 8.4 & 7.1 & 3.0 & 3.8 & 3.8 & 3.1 & 2.3 & 1.1 \\
ByteDance: Seed 2.0 & 4.2 & 5.3 & 7.7 & 3.4 & 3.1 & 4.4 & 3.5 & 2.6 & 1.1 \\
Anthropic: Claude 4.7 / 4.8\textsuperscript{†} & 3.1 & 4.5 & 4.2 & 3.5 & 3.4 & 3.0 & 3.4 & 2.9 & 1.5 \\
Anthropic: Claude 4.5 / 4.6 & 4.2 & 6.2 & 5.8 & 4.7 & 4.7 & 4.1 & 4.6 & 3.9 & 1.4 \\
Alibaba: Qwen 3.6 / 3.7-max & 4.4 & 6.3 & 6.2 & 4.4 & 2.7 & 2.6 & 2.5 & 1.4 & 1.1 \\
Alibaba: Qwen 3-235B & 4.7 & 5.9 & 6.2 & 5.7 & 3.0 & 3.2 & 3.4 & 2.7 & 1.3 \\
xAI: Grok 4.x & 4.4 & 9.9 & 5.1 & 5.0 & 3.3 & 2.9 & 3.0 & 2.2 & 1.2 \\
Moonshot: Kimi & 4.7 & 6.6 & 4.3 & 5.9 & 2.5 & 3.8 & 4.3 & 3.8 & 1.4 \\
DeepSeek v3.1--v4 & 4.7 & 8.7 & 5.3 & 6.0 & 2.7 & 3.3 & 3.5 & 2.1 & 1.3 \\
MiniMax: M3 & 4.3 & 10.3 & 8.8 & 8.5 & 2.9 & 2.4 & 2.6 & 4.5 & 1.2 \\
Mistral & 4.6 & 10.8 & 10.4 & 13.1 & 4.1 & 3.5 & 2.5 & 2.7 & 1.1 \\
NVIDIA: Nemotron 3 & 4.6 & 10.8 & 10.4 & 13.1 & 4.1 & 3.5 & 2.5 & 2.7 & 1.1 \\
\end{longtable}
}

\textbf{Table G.1:} \emph{Token cost relative to English by tokenizer family. The first numeric column is the absolute baseline (average characters per token in English); each remaining column is a multiplier against that family's own baseline: how many times more tokens a similar text requires than English text under the same tokenizer, so English is 1.0 by construction. Values are averaged over models sharing a tokenizer. Google's tokenizer carries the smallest multilingual tax (worst script ≤3.4×), while the Mistral and NVIDIA tokenizers, numerically identical across all languages and likely sharing a lineage, reach ≈13× on Khmer; within a single language, tokenizer choice alone changes cost by 4--6×. † Claude 4.7/4.8 use a different, less efficient tokenizer than earlier Claude models, including in English (3.1 chars/token vs ≈4.2--4.8 elsewhere); because multipliers are computed against each family's own English baseline, this family's lower multipliers reflect its lower baseline, not cheaper multilingual processing, and its absolute per-language cost is higher than the multipliers suggest.}

\end{landscape}

\begin{center}\rule{0.5\linewidth}{0.5pt}\end{center}

\section{Appendix H: Temperature Equivalence Details}\label{appendix-h-temperature-equivalence-details}

This appendix reports the full results of the temperature-equivalence test summarized in §3.5.2.

\subsection{H.1 Design}\label{h.1-design}

Five models (Gemini 3 Flash Preview, GPT-5.4, Grok 4.3, DeepSeek V4 Flash, Qwen 3.6 Max Preview) were compared at temperature 0 versus their default temperature on six languages spanning all resource tiers (German, Japanese, Czech, Thai, Tagalog, Assamese), all 100 items per language per task. Per-item values are built asymmetrically by design: the deterministic single run at temperature 0, against the mean of three runs at default temperature. Translation scores are normalized to {[}0, 1{]} via (score − 1)/4; grammar item values are correctness in \{0, 1\} (temperature 0) or the mean of three runs. For each (model, language, item) the paired difference d = x\_default − x\_temp0 is computed; equivalence is tested with two one-sided tests (TOST) at \(\alpha\) = 0.05 with margin Δ = 0.05 on the normalized scale: equivalent iff the 90\% CI of mean(d) lies within (−0.05, +0.05). Two translation items were dropped for known failures; no grammar items were dropped. Pooled translation results treat items as independent even though the same English source recurs across languages, so pooled standard errors are a simplification.

\subsection{H.2 Pooled results by model}\label{h.2-pooled-results-by-model}

{\def\LTcaptype{none} 
\begin{longtable}[]{@{}
  >{\raggedright\arraybackslash}p{(\linewidth - 10\tabcolsep) * \real{0.1316}}
  >{\raggedright\arraybackslash}p{(\linewidth - 10\tabcolsep) * \real{0.2632}}
  >{\raggedleft\arraybackslash}p{(\linewidth - 10\tabcolsep) * \real{0.1053}}
  >{\raggedleft\arraybackslash}p{(\linewidth - 10\tabcolsep) * \real{0.1053}}
  >{\raggedright\arraybackslash}p{(\linewidth - 10\tabcolsep) * \real{0.2632}}
  >{\raggedright\arraybackslash}p{(\linewidth - 10\tabcolsep) * \real{0.1316}}@{}}
\toprule\noalign{}
\begin{minipage}[b]{\linewidth}\raggedright
Task
\end{minipage} & \begin{minipage}[b]{\linewidth}\raggedright
Model
\end{minipage} & \begin{minipage}[b]{\linewidth}\raggedleft
n
\end{minipage} & \begin{minipage}[b]{\linewidth}\raggedleft
mean(d)
\end{minipage} & \begin{minipage}[b]{\linewidth}\raggedright
90\% CI
\end{minipage} & \begin{minipage}[b]{\linewidth}\raggedright
Equivalent
\end{minipage} \\
\midrule\noalign{}
\endhead
\bottomrule\noalign{}
\endlastfoot
Grammar & DeepSeek V4 Flash & 600 & −0.034 & {[}−0.065, −0.002{]} & no \\
Grammar & Gemini 3 Flash Preview & 600 & +0.017 & {[}−0.001, +0.034{]} & yes \\
Grammar & GPT-5.4 & 600 & −0.001 & {[}−0.023, +0.021{]} & yes \\
Grammar & Grok 4.3 & 600 & −0.001 & {[}−0.034, +0.033{]} & yes \\
Grammar & Qwen 3.6 Max Preview & 600 & +0.019 & {[}−0.005, +0.044{]} & yes \\
Translation & DeepSeek V4 Flash & 600 & −0.003 & {[}−0.013, +0.007{]} & yes \\
Translation & Gemini 3 Flash Preview & 598 & +0.005 & {[}−0.003, +0.012{]} & yes \\
Translation & GPT-5.4 & 600 & +0.003 & {[}−0.007, +0.012{]} & yes \\
Translation & Grok 4.3 & 600 & −0.033 & {[}−0.046, −0.020{]} & yes \\
Translation & Qwen 3.6 Max Preview & 600 & −0.040 & {[}−0.052, −0.029{]} & no \\
\end{longtable}
}

\textbf{Table H.1:} \emph{Paired equivalence test (TOST, Δ = 0.05 normalized), pooled by model across the six languages. Eight of ten model-task cells are equivalent. The two exceptions both run toward lower default-temperature scores: DeepSeek V4 Flash on grammar (−0.034 accuracy) and Qwen 3.6 Max Preview on translation (−0.040 normalized ≈ 0.16 points on the 1--5 scale).}

\subsection{H.3 MCC supplement}\label{h.3-mcc-supplement}

Because the headline grammar metric is MCC, equivalence was additionally checked on MCC computed from the pooled confusion matrix per (model, language, condition), with margin Δ\_MCC = 0.10. Pooled over all 30 model-language pairs, the mean difference is +0.009 (90\% CI {[}−0.019, +0.037{]}), within the margin. Per model, Gemini 3 Flash Preview (+0.035), GPT-5.4 (+0.002), and Grok 4.3 (+0.004) pass, while DeepSeek V4 Flash (−0.064) and Qwen 3.6 Max Preview (+0.068) have CIs that exceed the margin. Because MCC responds more sharply than accuracy to shifts in the balance of false positives and negatives, it flags Qwen, which was equivalent on accuracy (Table H.1). The two flagged models also differ in direction, with DeepSeek scoring lower at default temperature and Qwen higher, so neither points to a consistent temperature effect on grammar.

\subsection{H.4 Slice-level results}\label{h.4-slice-level-results}

At the (model × language) slice level (n ≈ 100), many slices fail the ±0.05 equivalence test simply because they are underpowered at this margin; we therefore report observed differences rather than equivalence flags (§3.5.2). Translation slices with non-equivalent results and their observed differences:

{\def\LTcaptype{none} 
\begin{longtable}[]{@{}llrl@{}}
\toprule\noalign{}
Model & Language & mean(d) & 90\% CI \\
\midrule\noalign{}
\endhead
\bottomrule\noalign{}
\endlastfoot
GPT-5.4 & as & −0.018 & {[}−0.054, +0.019{]} \\
GPT-5.4 & th & +0.029 & {[}+0.003, +0.055{]} \\
Grok 4.3 & as & −0.047 & {[}−0.091, −0.003{]} \\
Grok 4.3 & ja & −0.033 & {[}−0.059, −0.006{]} \\
Grok 4.3 & th & −0.051 & {[}−0.084, −0.019{]} \\
Grok 4.3 & tl & −0.038 & {[}−0.073, −0.003{]} \\
Qwen 3.6 Max Preview & as & −0.067 & {[}−0.105, −0.029{]} \\
Qwen 3.6 Max Preview & cs & −0.023 & {[}−0.054, +0.009{]} \\
Qwen 3.6 Max Preview & ja & −0.051 & {[}−0.073, −0.028{]} \\
Qwen 3.6 Max Preview & th & −0.037 & {[}−0.062, −0.011{]} \\
Qwen 3.6 Max Preview & tl & −0.055 & {[}−0.090, −0.021{]} \\
\end{longtable}
}

\textbf{Table H.2:} \emph{Non-equivalent translation slices (normalized scale). Grammar slices show the same pattern with wider intervals; the direction of non-equivalent slices is predominantly negative (default temperature scoring lower), so reporting at temperature 0 does not understate performance.}

\begin{center}\rule{0.5\linewidth}{0.5pt}\end{center}

\section{Appendix I: Human Validation Study Details}\label{appendix-i-human-validation-study-details}

\subsection{I.1 Sampling procedure}\label{i.1-sampling-procedure}

The validation sample contains 300 round-trip translation evaluations, drawn to be balanced across resource tiers and judge-score bands. Allocation is 100 items per tier (T1, T2, T3), and within each tier 25 items in each of four judge-score bands defined on the item-level mean judge score: {[}1, 2), {[}2, 3), {[}3, 4), and {[}4, 5{]}. Sampling used a fixed random seed for reproducibility.

Tier labels are taken from the language metadata (see §3.1 and Table 1). Within each tier we sampled from four representative languages chosen for script and family diversity:

{\def\LTcaptype{none} 
\begin{longtable}[]{@{}
  >{\raggedright\arraybackslash}p{(\linewidth - 2\tabcolsep) * \real{0.3529}}
  >{\raggedright\arraybackslash}p{(\linewidth - 2\tabcolsep) * \real{0.6471}}@{}}
\toprule\noalign{}
\begin{minipage}[b]{\linewidth}\raggedright
Tier
\end{minipage} & \begin{minipage}[b]{\linewidth}\raggedright
Languages
\end{minipage} \\
\midrule\noalign{}
\endhead
\bottomrule\noalign{}
\endlastfoot
T1 (high-resource) & German (de), Japanese (ja), Arabic (ar), Simplified Chinese (zh-CN) \\
T2 (mid-resource) & Czech (cs), Thai (th), Tamil (ta), Khmer (km) \\
T3 (low-resource) & Tagalog (tl), Assamese (as), Tigrinya (ti-ET), Dzongkha (dz) \\
\end{longtable}
}

Only evaluations with exactly three successfully parsed judge scores were eligible. To avoid over-concentration, we capped how many sampled items could share the same model or language, preferring candidates from less-represented models and languages during selection.

Each sampled item was presented to annotators in a blinded form containing the original English sentence, the round-trip English, the item's target meaning, and (where available) the worked scoring examples. Judge scores and rater identities were not shown. Annotators applied the same 1--5 rubric used by the LLM judges (§3.3.4).

\subsection{I.2 Metrics}\label{i.2-metrics}

We compute the following agreement measures, separately for two cohorts: the three LLM judges and the three human annotators.

\begin{itemize}
\tightlist
\item
  \textbf{Krippendorff's \(\alpha\) (ordinal):} chance-corrected agreement appropriate for ordered 1--5 ratings, computed jointly over the raters in the cohort.
\item
  \textbf{Pearson r:} computed for each rater pair, then averaged across pairs using Fisher-z transformation.
\item
  \textbf{Exact agreement:} proportion of rating pairs with identical scores.
\item
  \textbf{Within-one agreement:} proportion of rating pairs differing by at most one point (\textbar a − b\textbar{} ≤ 1).
\end{itemize}

For the panel-level analysis we compute, per item, the mean of the three judge scores and the mean of the three human scores, then report Pearson r and Spearman \(\rho\) between these two vectors across items, with 95\% confidence intervals from 5,000 bootstrap resamples over items. We also report the mean signed difference (judge mean minus human mean) and the mean absolute error. Panel-level results are given overall and by tier in Table 8 (§4.2).

Matching between human and judge ratings is done at the level of the individual evaluation, so each item contributes one judge panel mean and one human panel mean.

Within-band agreement is dominated by range restriction and is not informative as a validity measure; band balancing serves only to ensure the overall estimate spans the full score range.

\subsection{I.3 Pairwise rater agreement}\label{i.3-pairwise-rater-agreement}

Table I.1 reports agreement for each individual rater pair over the full sample. Human raters are anonymized (H1, H2, H3).

{\def\LTcaptype{none} 
\begin{longtable}[]{@{}
  >{\raggedright\arraybackslash}p{(\linewidth - 12\tabcolsep) * \real{0.1625}}
  >{\raggedright\arraybackslash}p{(\linewidth - 12\tabcolsep) * \real{0.2000}}
  >{\raggedright\arraybackslash}p{(\linewidth - 12\tabcolsep) * \real{0.2000}}
  >{\raggedleft\arraybackslash}p{(\linewidth - 12\tabcolsep) * \real{0.1375}}
  >{\raggedleft\arraybackslash}p{(\linewidth - 12\tabcolsep) * \real{0.0625}}
  >{\raggedleft\arraybackslash}p{(\linewidth - 12\tabcolsep) * \real{0.0875}}
  >{\raggedleft\arraybackslash}p{(\linewidth - 12\tabcolsep) * \real{0.1500}}@{}}
\toprule\noalign{}
\begin{minipage}[b]{\linewidth}\raggedright
Pair type
\end{minipage} & \begin{minipage}[b]{\linewidth}\raggedright
Rater A
\end{minipage} & \begin{minipage}[b]{\linewidth}\raggedright
Rater B
\end{minipage} & \begin{minipage}[b]{\linewidth}\raggedleft
Pearson r
\end{minipage} & \begin{minipage}[b]{\linewidth}\raggedleft
\(\alpha\)
\end{minipage} & \begin{minipage}[b]{\linewidth}\raggedleft
Exact
\end{minipage} & \begin{minipage}[b]{\linewidth}\raggedleft
Within-one
\end{minipage} \\
\midrule\noalign{}
\endhead
\bottomrule\noalign{}
\endlastfoot
Judge--judge & claude-haiku-4.5 & gemini-3.1-flash-lite-preview & 0.83 & 0.82 & 61\% & 92\% \\
Judge--judge & claude-haiku-4.5 & gpt-5.4-mini & 0.79 & 0.79 & 57\% & 89\% \\
Judge--judge & gemini-3.1-flash-lite-preview & gpt-5.4-mini & 0.86 & 0.84 & 59\% & 95\% \\
Human--human & H1 & H2 & 0.72 & 0.72 & 53\% & 82\% \\
Human--human & H1 & H3 & 0.69 & 0.67 & 46\% & 78\% \\
Human--human & H2 & H3 & 0.70 & 0.69 & 53\% & 78\% \\
Judge--human & H1 & claude-haiku-4.5 & 0.70 & 0.69 & 42\% & 82\% \\
Judge--human & H1 & gemini-3.1-flash-lite-preview & 0.65 & 0.64 & 40\% & 77\% \\
Judge--human & H1 & gpt-5.4-mini & 0.59 & 0.59 & 37\% & 73\% \\
Judge--human & H2 & claude-haiku-4.5 & 0.69 & 0.68 & 44\% & 81\% \\
Judge--human & H2 & gemini-3.1-flash-lite-preview & 0.60 & 0.59 & 41\% & 73\% \\
Judge--human & H2 & gpt-5.4-mini & 0.60 & 0.60 & 40\% & 75\% \\
Judge--human & H3 & claude-haiku-4.5 & 0.70 & 0.65 & 45\% & 77\% \\
Judge--human & H3 & gemini-3.1-flash-lite-preview & 0.69 & 0.66 & 50\% & 72\% \\
Judge--human & H3 & gpt-5.4-mini & 0.66 & 0.63 & 45\% & 76\% \\
\end{longtable}
}

\textbf{Table I.1:} \emph{All 15 pairwise rater agreements over the full sample (n = 300 each). Human raters anonymized. Cohort-level Pearson r (Table 7) is the Fisher-z average of the unrounded pair correlations, so it can differ by up to 0.01 from re-averaging the rounded values shown here.}

\subsection{I.4 Within-cohort agreement by resource tier}\label{i.4-within-cohort-agreement-by-resource-tier}

The paper's claim that the judge panel is more internally consistent than the human annotators within every resource tier (§4.2) is supported by the per-tier breakdown:

{\def\LTcaptype{none} 
\begin{longtable}[]{@{}llrrrrr@{}}
\toprule\noalign{}
Cohort & Tier & Krippendorff \(\alpha\) & Pearson r & Exact & Within-one & n \\
\midrule\noalign{}
\endhead
\bottomrule\noalign{}
\endlastfoot
LLM judges & T1 & 0.84 & 0.85 & 63\% & 93\% & 100 \\
LLM judges & T2 & 0.78 & 0.79 & 56\% & 91\% & 100 \\
LLM judges & T3 & 0.84 & 0.85 & 58\% & 92\% & 100 \\
Human annotators & T1 & 0.64 & 0.64 & 50\% & 75\% & 100 \\
Human annotators & T2 & 0.71 & 0.72 & 54\% & 82\% & 100 \\
Human annotators & T3 & 0.72 & 0.75 & 48\% & 81\% & 100 \\
\end{longtable}
}

\textbf{Table I.2:} \emph{Within-cohort agreement by resource tier (validation sample; 100 items per tier). Judge-panel agreement exceeds human-annotator agreement in every tier, and neither cohort degrades on low-resource languages.}

\subsection{I.5 Agreement by judge-score band}\label{i.5-agreement-by-judge-score-band}

Within-band agreement is dominated by range restriction: conditioning on a one-point slice of the scale removes most of the variance that correlation-based measures rely on, so within-band \(\alpha\) and r are mechanically depressed even when raters agree closely in absolute terms (§4.2). The bands are reported for completeness; within-one agreement is the more interpretable measure here.

{\def\LTcaptype{none} 
\begin{longtable}[]{@{}llrrrrr@{}}
\toprule\noalign{}
Cohort & Band & Krippendorff \(\alpha\) & Pearson r & Exact & Within-one & n \\
\midrule\noalign{}
\endhead
\bottomrule\noalign{}
\endlastfoot
LLM judges & {[}1,2) & 0.11 & 0.08 & 68\% & 96\% & 75 \\
LLM judges & {[}2,3) & −0.11 & −0.20 & 51\% & 88\% & 75 \\
LLM judges & {[}3,4) & −0.26 & −0.23 & 35\% & 85\% & 75 \\
LLM judges & {[}4,5{]} & 0.49 & 0.47 & 82\% & 99\% & 75 \\
Human annotators & {[}1,2) & 0.48 & 0.52 & 67\% & 83\% & 75 \\
Human annotators & {[}2,3) & 0.36 & 0.38 & 35\% & 68\% & 75 \\
Human annotators & {[}3,4) & 0.47 & 0.55 & 36\% & 76\% & 75 \\
Human annotators & {[}4,5{]} & 0.31 & 0.47 & 65\% & 91\% & 75 \\
\end{longtable}
}

\textbf{Table I.3:} \emph{Within-cohort agreement inside each judge-score band (75 items per band, pooled over tiers). Bands are defined on the item-level mean judge score, so LLM-judge scores are range-restricted within bands by construction, which depresses \(\alpha\) and r; the humans, whose scores are not band-restricted, retain moderate within-band correlations. Judge within-one agreement remains 85--99\% in all bands. The signed judge−human difference by band is reported with the systematic-bias discussion in §4.2.}

\begin{center}\rule{0.5\linewidth}{0.5pt}\end{center}

\begin{landscape}

\footnotesize
\setlength{\tabcolsep}{4pt}
\renewcommand{\arraystretch}{0.95}

\section{Appendix J: Full Per-Language Results}\label{appendix-j-full-per-language-results}

This appendix gives the complete model-by-language result tables underlying the heatmaps referenced in §5.1. All values use the frozen July 2026 reporting snapshot; grammar results are at temperature 0 and translation results on the hard subset. Configurations are sorted by their cross-language mean (the \emph{Mean} column; grammar averages all 30 languages, translation all 29 targets). Grammar MCC values are comparable within a language column but not across columns (§3.2.7); translation scores are comparable across languages (§3.3.7). Current results for all configurations, including models added after this snapshot, are at \href{https://m-gate.ai}{m-gate.ai}.

\subsection{J.1 Grammar (MCC, temperature 0)}\label{j.1-grammar-mcc-temperature-0}

\subsubsection{Grammar MCC, Tier 1 languages (temperature 0)}\label{grammar-mcc-tier-1-languages-temperature-0}

{\def\LTcaptype{none} 
\begin{longtable}[]{@{}
  >{\raggedright\arraybackslash}p{(\linewidth - 22\tabcolsep) * \real{0.1698}}
  >{\raggedleft\arraybackslash}p{(\linewidth - 22\tabcolsep) * \real{0.0755}}
  >{\raggedleft\arraybackslash}p{(\linewidth - 22\tabcolsep) * \real{0.0755}}
  >{\raggedleft\arraybackslash}p{(\linewidth - 22\tabcolsep) * \real{0.0755}}
  >{\raggedleft\arraybackslash}p{(\linewidth - 22\tabcolsep) * \real{0.0755}}
  >{\raggedleft\arraybackslash}p{(\linewidth - 22\tabcolsep) * \real{0.0755}}
  >{\raggedleft\arraybackslash}p{(\linewidth - 22\tabcolsep) * \real{0.0755}}
  >{\raggedleft\arraybackslash}p{(\linewidth - 22\tabcolsep) * \real{0.0755}}
  >{\raggedleft\arraybackslash}p{(\linewidth - 22\tabcolsep) * \real{0.0755}}
  >{\raggedleft\arraybackslash}p{(\linewidth - 22\tabcolsep) * \real{0.0755}}
  >{\raggedleft\arraybackslash}p{(\linewidth - 22\tabcolsep) * \real{0.0755}}
  >{\raggedleft\arraybackslash}p{(\linewidth - 22\tabcolsep) * \real{0.0755}}@{}}
\toprule\noalign{}
\begin{minipage}[b]{\linewidth}\raggedright
Configuration
\end{minipage} & \begin{minipage}[b]{\linewidth}\raggedleft
Mean*
\end{minipage} & \begin{minipage}[b]{\linewidth}\raggedleft
en-US
\end{minipage} & \begin{minipage}[b]{\linewidth}\raggedleft
de
\end{minipage} & \begin{minipage}[b]{\linewidth}\raggedleft
ja
\end{minipage} & \begin{minipage}[b]{\linewidth}\raggedleft
zh-CN
\end{minipage} & \begin{minipage}[b]{\linewidth}\raggedleft
es
\end{minipage} & \begin{minipage}[b]{\linewidth}\raggedleft
fr
\end{minipage} & \begin{minipage}[b]{\linewidth}\raggedleft
pt-BR
\end{minipage} & \begin{minipage}[b]{\linewidth}\raggedleft
pl
\end{minipage} & \begin{minipage}[b]{\linewidth}\raggedleft
ko
\end{minipage} & \begin{minipage}[b]{\linewidth}\raggedleft
ar
\end{minipage} \\
\midrule\noalign{}
\endhead
\bottomrule\noalign{}
\endlastfoot
Gemini 3.1 Pro Preview & +0.36 & +0.59 & +0.31 & -0.04 & +0.30 & +0.40 & +0.43 & +0.60 & +0.28 & +0.52 & +0.41 \\
Claude Fable 5 & +0.36 & +0.49 & +0.35 & +0.09 & +0.35 & +0.35 & +0.45 & +0.65 & +0.32 & +0.56 & +0.41 \\
Gemini 3.5 Flash & +0.30 & +0.54 & +0.27 & -0.20 & +0.37 & +0.36 & +0.39 & +0.65 & +0.12 & +0.46 & +0.36 \\
Gemini 3.5 Flash (High) & +0.28 & +0.48 & +0.27 & -0.19 & +0.38 & +0.35 & +0.39 & +0.62 & +0.28 & +0.44 & +0.36 \\
Gemini 3 Flash Preview & +0.23 & +0.44 & +0.11 & -0.32 & +0.30 & +0.33 & +0.36 & +0.60 & +0.12 & +0.24 & +0.24 \\
Grok 4.20 & +0.20 & -0.15 & +0.20 & +0.38 & +0.27 & +0.08 & +0.30 & +0.04 & +0.39 & +0.34 & +0.38 \\
Gemini 3.5 Flash (Minimal) & +0.19 & +0.19 & +0.12 & -0.39 & +0.18 & +0.23 & +0.10 & +0.41 & +0.08 & +0.12 & +0.00 \\
GPT-5.5 High & +0.18 & +0.26 & +0.19 & +0.02 & -0.04 & +0.29 & +0.36 & +0.47 & +0.08 & +0.30 & +0.14 \\
\end{longtable}
}

\textbf{Table J.1:} \emph{Grammar MCC by configuration and Tier 1 language at temperature 0. *The Mean column averages all 30 languages.}

\subsubsection{Grammar MCC, Tier 2 languages (temperature 0)}\label{grammar-mcc-tier-2-languages-temperature-0}

{\def\LTcaptype{none} 
\begin{longtable}[]{@{}
  >{\raggedright\arraybackslash}p{(\linewidth - 22\tabcolsep) * \real{0.1698}}
  >{\raggedleft\arraybackslash}p{(\linewidth - 22\tabcolsep) * \real{0.0755}}
  >{\raggedleft\arraybackslash}p{(\linewidth - 22\tabcolsep) * \real{0.0755}}
  >{\raggedleft\arraybackslash}p{(\linewidth - 22\tabcolsep) * \real{0.0755}}
  >{\raggedleft\arraybackslash}p{(\linewidth - 22\tabcolsep) * \real{0.0755}}
  >{\raggedleft\arraybackslash}p{(\linewidth - 22\tabcolsep) * \real{0.0755}}
  >{\raggedleft\arraybackslash}p{(\linewidth - 22\tabcolsep) * \real{0.0755}}
  >{\raggedleft\arraybackslash}p{(\linewidth - 22\tabcolsep) * \real{0.0755}}
  >{\raggedleft\arraybackslash}p{(\linewidth - 22\tabcolsep) * \real{0.0755}}
  >{\raggedleft\arraybackslash}p{(\linewidth - 22\tabcolsep) * \real{0.0755}}
  >{\raggedleft\arraybackslash}p{(\linewidth - 22\tabcolsep) * \real{0.0755}}
  >{\raggedleft\arraybackslash}p{(\linewidth - 22\tabcolsep) * \real{0.0755}}@{}}
\toprule\noalign{}
\begin{minipage}[b]{\linewidth}\raggedright
Configuration
\end{minipage} & \begin{minipage}[b]{\linewidth}\raggedleft
Mean*
\end{minipage} & \begin{minipage}[b]{\linewidth}\raggedleft
cs
\end{minipage} & \begin{minipage}[b]{\linewidth}\raggedleft
id
\end{minipage} & \begin{minipage}[b]{\linewidth}\raggedleft
fi
\end{minipage} & \begin{minipage}[b]{\linewidth}\raggedleft
th
\end{minipage} & \begin{minipage}[b]{\linewidth}\raggedleft
ne
\end{minipage} & \begin{minipage}[b]{\linewidth}\raggedleft
sq
\end{minipage} & \begin{minipage}[b]{\linewidth}\raggedleft
ta
\end{minipage} & \begin{minipage}[b]{\linewidth}\raggedleft
eu
\end{minipage} & \begin{minipage}[b]{\linewidth}\raggedleft
is
\end{minipage} & \begin{minipage}[b]{\linewidth}\raggedleft
km
\end{minipage} \\
\midrule\noalign{}
\endhead
\bottomrule\noalign{}
\endlastfoot
Gemini 3.1 Pro Preview & +0.36 & +0.55 & +0.54 & +0.43 & +0.23 & +0.39 & +0.48 & +0.36 & +0.49 & +0.53 & -0.04 \\
Claude Fable 5 & +0.36 & +0.67 & +0.62 & +0.34 & +0.27 & +0.47 & +0.52 & +0.31 & +0.50 & +0.69 & -0.16 \\
Gemini 3.5 Flash & +0.30 & +0.45 & +0.50 & +0.10 & +0.05 & +0.42 & +0.48 & +0.36 & +0.34 & +0.53 & -0.16 \\
Gemini 3.5 Flash (High) & +0.28 & +0.54 & +0.50 & +0.13 & +0.10 & +0.40 & +0.42 & +0.22 & +0.28 & +0.53 & -0.08 \\
Gemini 3 Flash Preview & +0.23 & +0.44 & +0.46 & +0.00 & +0.04 & +0.35 & +0.42 & +0.04 & +0.12 & +0.20 & -0.17 \\
Grok 4.20 & +0.20 & +0.06 & +0.00 & +0.41 & +0.44 & -0.27 & +0.18 & +0.27 & +0.38 & +0.35 & +0.42 \\
Gemini 3.5 Flash (Minimal) & +0.19 & +0.29 & +0.30 & +0.21 & +0.08 & +0.56 & +0.35 & +0.16 & +0.16 & +0.32 & -0.14 \\
GPT-5.5 High & +0.18 & +0.49 & +0.34 & +0.30 & -0.12 & +0.44 & +0.47 & -0.12 & +0.30 & +0.30 & -0.26 \\
\end{longtable}
}

\textbf{Table J.2:} \emph{Grammar MCC by configuration and Tier 2 language at temperature 0. *The Mean column averages all 30 languages.}

\subsubsection{Grammar MCC, Tier 3 languages (temperature 0)}\label{grammar-mcc-tier-3-languages-temperature-0}

{\def\LTcaptype{none} 
\begin{longtable}[]{@{}
  >{\raggedright\arraybackslash}p{(\linewidth - 22\tabcolsep) * \real{0.1698}}
  >{\raggedleft\arraybackslash}p{(\linewidth - 22\tabcolsep) * \real{0.0755}}
  >{\raggedleft\arraybackslash}p{(\linewidth - 22\tabcolsep) * \real{0.0755}}
  >{\raggedleft\arraybackslash}p{(\linewidth - 22\tabcolsep) * \real{0.0755}}
  >{\raggedleft\arraybackslash}p{(\linewidth - 22\tabcolsep) * \real{0.0755}}
  >{\raggedleft\arraybackslash}p{(\linewidth - 22\tabcolsep) * \real{0.0755}}
  >{\raggedleft\arraybackslash}p{(\linewidth - 22\tabcolsep) * \real{0.0755}}
  >{\raggedleft\arraybackslash}p{(\linewidth - 22\tabcolsep) * \real{0.0755}}
  >{\raggedleft\arraybackslash}p{(\linewidth - 22\tabcolsep) * \real{0.0755}}
  >{\raggedleft\arraybackslash}p{(\linewidth - 22\tabcolsep) * \real{0.0755}}
  >{\raggedleft\arraybackslash}p{(\linewidth - 22\tabcolsep) * \real{0.0755}}
  >{\raggedleft\arraybackslash}p{(\linewidth - 22\tabcolsep) * \real{0.0755}}@{}}
\toprule\noalign{}
\begin{minipage}[b]{\linewidth}\raggedright
Configuration
\end{minipage} & \begin{minipage}[b]{\linewidth}\raggedleft
Mean*
\end{minipage} & \begin{minipage}[b]{\linewidth}\raggedleft
ky
\end{minipage} & \begin{minipage}[b]{\linewidth}\raggedleft
tl
\end{minipage} & \begin{minipage}[b]{\linewidth}\raggedleft
as
\end{minipage} & \begin{minipage}[b]{\linewidth}\raggedleft
mt
\end{minipage} & \begin{minipage}[b]{\linewidth}\raggedleft
rw
\end{minipage} & \begin{minipage}[b]{\linewidth}\raggedleft
ti-ET
\end{minipage} & \begin{minipage}[b]{\linewidth}\raggedleft
qu
\end{minipage} & \begin{minipage}[b]{\linewidth}\raggedleft
gn
\end{minipage} & \begin{minipage}[b]{\linewidth}\raggedleft
fj
\end{minipage} & \begin{minipage}[b]{\linewidth}\raggedleft
dz
\end{minipage} \\
\midrule\noalign{}
\endhead
\bottomrule\noalign{}
\endlastfoot
Gemini 3.1 Pro Preview & +0.36 & +0.71 & +0.40 & +0.54 & +0.42 & +0.51 & +0.00 & +0.39 & +0.22 & -0.22 & +0.12 \\
Claude Fable 5 & +0.36 & +0.59 & +0.01 & +0.36 & +0.36 & +0.43 & +0.01 & +0.26 & +0.31 & -0.17 & +0.28 \\
Gemini 3.5 Flash & +0.30 & +0.60 & +0.34 & +0.41 & +0.48 & +0.36 & -0.02 & +0.43 & +0.16 & -0.14 & -0.02 \\
Gemini 3.5 Flash (High) & +0.28 & +0.59 & +0.26 & +0.41 & +0.38 & +0.23 & -0.11 & +0.30 & +0.08 & -0.17 & +0.08 \\
Gemini 3 Flash Preview & +0.23 & +0.58 & +0.11 & +0.46 & +0.38 & +0.40 & +0.06 & +0.34 & +0.12 & -0.07 & +0.08 \\
Grok 4.20 & +0.20 & +0.26 & +0.44 & -0.02 & +0.27 & +0.10 & +0.00 & +0.00 & +0.14 & +0.24 & +0.16 \\
Gemini 3.5 Flash (Minimal) & +0.19 & +0.48 & -0.09 & +0.47 & +0.41 & +0.38 & +0.12 & +0.26 & +0.05 & +0.06 & +0.12 \\
GPT-5.5 High & +0.18 & +0.40 & +0.00 & +0.33 & +0.26 & +0.31 & +0.05 & +0.28 & +0.12 & -0.46 & +0.00 \\
\end{longtable}
}

\textbf{Table J.3:} \emph{Grammar MCC by configuration and Tier 3 language at temperature 0. *The Mean column averages all 30 languages.}

\subsection{J.2 Translation (normalized score, hard subset)}\label{j.2-translation-normalized-score-hard-subset}

\subsubsection{Translation normalized score, hard subset, Tier 1 languages}\label{translation-normalized-score-hard-subset-tier-1-languages}

{\def\LTcaptype{none} 
\begin{longtable}[]{@{}
  >{\raggedright\arraybackslash}p{(\linewidth - 20\tabcolsep) * \real{0.1837}}
  >{\raggedleft\arraybackslash}p{(\linewidth - 20\tabcolsep) * \real{0.0816}}
  >{\raggedleft\arraybackslash}p{(\linewidth - 20\tabcolsep) * \real{0.0816}}
  >{\raggedleft\arraybackslash}p{(\linewidth - 20\tabcolsep) * \real{0.0816}}
  >{\raggedleft\arraybackslash}p{(\linewidth - 20\tabcolsep) * \real{0.0816}}
  >{\raggedleft\arraybackslash}p{(\linewidth - 20\tabcolsep) * \real{0.0816}}
  >{\raggedleft\arraybackslash}p{(\linewidth - 20\tabcolsep) * \real{0.0816}}
  >{\raggedleft\arraybackslash}p{(\linewidth - 20\tabcolsep) * \real{0.0816}}
  >{\raggedleft\arraybackslash}p{(\linewidth - 20\tabcolsep) * \real{0.0816}}
  >{\raggedleft\arraybackslash}p{(\linewidth - 20\tabcolsep) * \real{0.0816}}
  >{\raggedleft\arraybackslash}p{(\linewidth - 20\tabcolsep) * \real{0.0816}}@{}}
\toprule\noalign{}
\begin{minipage}[b]{\linewidth}\raggedright
Configuration
\end{minipage} & \begin{minipage}[b]{\linewidth}\raggedleft
Mean*
\end{minipage} & \begin{minipage}[b]{\linewidth}\raggedleft
de
\end{minipage} & \begin{minipage}[b]{\linewidth}\raggedleft
ja
\end{minipage} & \begin{minipage}[b]{\linewidth}\raggedleft
zh-CN
\end{minipage} & \begin{minipage}[b]{\linewidth}\raggedleft
es
\end{minipage} & \begin{minipage}[b]{\linewidth}\raggedleft
fr
\end{minipage} & \begin{minipage}[b]{\linewidth}\raggedleft
pt-BR
\end{minipage} & \begin{minipage}[b]{\linewidth}\raggedleft
pl
\end{minipage} & \begin{minipage}[b]{\linewidth}\raggedleft
ko
\end{minipage} & \begin{minipage}[b]{\linewidth}\raggedleft
ar
\end{minipage} \\
\midrule\noalign{}
\endhead
\bottomrule\noalign{}
\endlastfoot
GPT-5.5 High & 0.82 & 0.91 & 0.88 & 0.87 & 0.87 & 0.92 & 0.95 & 0.90 & 0.86 & 0.89 \\
GPT-5.4 (High) & 0.82 & 0.91 & 0.85 & 0.82 & 0.90 & 0.94 & 0.94 & 0.90 & 0.86 & 0.90 \\
Claude Fable 5 & 0.81 & 0.92 & 0.86 & 0.90 & 0.88 & 0.91 & 0.93 & 0.91 & 0.88 & 0.88 \\
GPT-5.5 Medium & 0.81 & 0.91 & 0.83 & 0.86 & 0.90 & 0.90 & 0.92 & 0.88 & 0.84 & 0.87 \\
GPT-5.4 (Medium) & 0.80 & 0.92 & 0.84 & 0.83 & 0.90 & 0.91 & 0.94 & 0.89 & 0.86 & 0.85 \\
Claude Opus 4.6 (Adaptive) & 0.78 & 0.91 & 0.85 & 0.87 & 0.93 & 0.88 & 0.92 & 0.90 & 0.82 & 0.88 \\
Claude Opus 4.6 (Thinking) & 0.78 & 0.91 & 0.87 & 0.83 & 0.92 & 0.90 & 0.93 & 0.88 & 0.85 & 0.91 \\
GPT-5 & 0.78 & 0.90 & 0.84 & 0.86 & 0.91 & 0.89 & 0.92 & 0.86 & 0.83 & 0.87 \\
\end{longtable}
}

\textbf{Table J.4:} \emph{Normalized hard-subset round-trip translation score by configuration and Tier 1 target language. *The Mean column averages all 29 target languages.}

\subsubsection{Translation normalized score, hard subset, Tier 2 languages}\label{translation-normalized-score-hard-subset-tier-2-languages}

{\def\LTcaptype{none} 
\begin{longtable}[]{@{}
  >{\raggedright\arraybackslash}p{(\linewidth - 22\tabcolsep) * \real{0.1698}}
  >{\raggedleft\arraybackslash}p{(\linewidth - 22\tabcolsep) * \real{0.0755}}
  >{\raggedleft\arraybackslash}p{(\linewidth - 22\tabcolsep) * \real{0.0755}}
  >{\raggedleft\arraybackslash}p{(\linewidth - 22\tabcolsep) * \real{0.0755}}
  >{\raggedleft\arraybackslash}p{(\linewidth - 22\tabcolsep) * \real{0.0755}}
  >{\raggedleft\arraybackslash}p{(\linewidth - 22\tabcolsep) * \real{0.0755}}
  >{\raggedleft\arraybackslash}p{(\linewidth - 22\tabcolsep) * \real{0.0755}}
  >{\raggedleft\arraybackslash}p{(\linewidth - 22\tabcolsep) * \real{0.0755}}
  >{\raggedleft\arraybackslash}p{(\linewidth - 22\tabcolsep) * \real{0.0755}}
  >{\raggedleft\arraybackslash}p{(\linewidth - 22\tabcolsep) * \real{0.0755}}
  >{\raggedleft\arraybackslash}p{(\linewidth - 22\tabcolsep) * \real{0.0755}}
  >{\raggedleft\arraybackslash}p{(\linewidth - 22\tabcolsep) * \real{0.0755}}@{}}
\toprule\noalign{}
\begin{minipage}[b]{\linewidth}\raggedright
Configuration
\end{minipage} & \begin{minipage}[b]{\linewidth}\raggedleft
Mean*
\end{minipage} & \begin{minipage}[b]{\linewidth}\raggedleft
cs
\end{minipage} & \begin{minipage}[b]{\linewidth}\raggedleft
id
\end{minipage} & \begin{minipage}[b]{\linewidth}\raggedleft
fi
\end{minipage} & \begin{minipage}[b]{\linewidth}\raggedleft
th
\end{minipage} & \begin{minipage}[b]{\linewidth}\raggedleft
ne
\end{minipage} & \begin{minipage}[b]{\linewidth}\raggedleft
sq
\end{minipage} & \begin{minipage}[b]{\linewidth}\raggedleft
ta
\end{minipage} & \begin{minipage}[b]{\linewidth}\raggedleft
eu
\end{minipage} & \begin{minipage}[b]{\linewidth}\raggedleft
is
\end{minipage} & \begin{minipage}[b]{\linewidth}\raggedleft
km
\end{minipage} \\
\midrule\noalign{}
\endhead
\bottomrule\noalign{}
\endlastfoot
GPT-5.5 High & 0.82 & 0.90 & 0.77 & 0.78 & 0.88 & 0.81 & 0.90 & 0.81 & 0.75 & 0.92 & 0.86 \\
GPT-5.4 (High) & 0.82 & 0.91 & 0.79 & 0.77 & 0.88 & 0.83 & 0.91 & 0.80 & 0.70 & 0.91 & 0.84 \\
Claude Fable 5 & 0.81 & 0.93 & 0.78 & 0.74 & 0.87 & 0.84 & 0.90 & 0.77 & 0.74 & 0.92 & 0.79 \\
GPT-5.5 Medium & 0.81 & 0.90 & 0.77 & 0.74 & 0.88 & 0.81 & 0.90 & 0.79 & 0.70 & 0.91 & 0.84 \\
GPT-5.4 (Medium) & 0.80 & 0.90 & 0.74 & 0.74 & 0.86 & 0.84 & 0.92 & 0.82 & 0.73 & 0.91 & 0.82 \\
Claude Opus 4.6 (Adaptive) & 0.78 & 0.85 & 0.68 & 0.71 & 0.87 & 0.83 & 0.88 & 0.77 & 0.72 & 0.90 & 0.82 \\
Claude Opus 4.6 (Thinking) & 0.78 & 0.88 & 0.68 & 0.69 & 0.85 & 0.82 & 0.90 & 0.79 & 0.72 & 0.90 & 0.83 \\
GPT-5 & 0.78 & 0.89 & 0.75 & 0.71 & 0.83 & 0.82 & 0.88 & 0.82 & 0.73 & 0.89 & 0.84 \\
\end{longtable}
}

\textbf{Table J.5:} \emph{Normalized hard-subset round-trip translation score by configuration and Tier 2 target language. *The Mean column averages all 29 target languages.}

\subsubsection{Translation normalized score, hard subset, Tier 3 languages}\label{translation-normalized-score-hard-subset-tier-3-languages}

{\def\LTcaptype{none} 
\begin{longtable}[]{@{}
  >{\raggedright\arraybackslash}p{(\linewidth - 22\tabcolsep) * \real{0.1698}}
  >{\raggedleft\arraybackslash}p{(\linewidth - 22\tabcolsep) * \real{0.0755}}
  >{\raggedleft\arraybackslash}p{(\linewidth - 22\tabcolsep) * \real{0.0755}}
  >{\raggedleft\arraybackslash}p{(\linewidth - 22\tabcolsep) * \real{0.0755}}
  >{\raggedleft\arraybackslash}p{(\linewidth - 22\tabcolsep) * \real{0.0755}}
  >{\raggedleft\arraybackslash}p{(\linewidth - 22\tabcolsep) * \real{0.0755}}
  >{\raggedleft\arraybackslash}p{(\linewidth - 22\tabcolsep) * \real{0.0755}}
  >{\raggedleft\arraybackslash}p{(\linewidth - 22\tabcolsep) * \real{0.0755}}
  >{\raggedleft\arraybackslash}p{(\linewidth - 22\tabcolsep) * \real{0.0755}}
  >{\raggedleft\arraybackslash}p{(\linewidth - 22\tabcolsep) * \real{0.0755}}
  >{\raggedleft\arraybackslash}p{(\linewidth - 22\tabcolsep) * \real{0.0755}}
  >{\raggedleft\arraybackslash}p{(\linewidth - 22\tabcolsep) * \real{0.0755}}@{}}
\toprule\noalign{}
\begin{minipage}[b]{\linewidth}\raggedright
Configuration
\end{minipage} & \begin{minipage}[b]{\linewidth}\raggedleft
Mean*
\end{minipage} & \begin{minipage}[b]{\linewidth}\raggedleft
ky
\end{minipage} & \begin{minipage}[b]{\linewidth}\raggedleft
tl
\end{minipage} & \begin{minipage}[b]{\linewidth}\raggedleft
as
\end{minipage} & \begin{minipage}[b]{\linewidth}\raggedleft
mt
\end{minipage} & \begin{minipage}[b]{\linewidth}\raggedleft
rw
\end{minipage} & \begin{minipage}[b]{\linewidth}\raggedleft
ti-ET
\end{minipage} & \begin{minipage}[b]{\linewidth}\raggedleft
qu
\end{minipage} & \begin{minipage}[b]{\linewidth}\raggedleft
gn
\end{minipage} & \begin{minipage}[b]{\linewidth}\raggedleft
fj
\end{minipage} & \begin{minipage}[b]{\linewidth}\raggedleft
dz
\end{minipage} \\
\midrule\noalign{}
\endhead
\bottomrule\noalign{}
\endlastfoot
GPT-5.5 High & 0.82 & 0.71 & 0.73 & 0.81 & 0.88 & 0.67 & 0.77 & 0.64 & 0.74 & 0.65 & 0.78 \\
GPT-5.4 (High) & 0.82 & 0.72 & 0.75 & 0.78 & 0.91 & 0.68 & 0.83 & 0.61 & 0.71 & 0.68 & 0.72 \\
Claude Fable 5 & 0.81 & 0.69 & 0.72 & 0.73 & 0.92 & 0.65 & 0.86 & 0.63 & 0.70 & 0.62 & 0.78 \\
GPT-5.5 Medium & 0.81 & 0.70 & 0.78 & 0.77 & 0.88 & 0.67 & 0.81 & 0.63 & 0.68 & 0.66 & 0.77 \\
GPT-5.4 (Medium) & 0.80 & 0.71 & 0.73 & 0.77 & 0.90 & 0.65 & 0.73 & 0.59 & 0.67 & 0.56 & 0.74 \\
Claude Opus 4.6 (Adaptive) & 0.78 & 0.65 & 0.72 & 0.77 & 0.89 & 0.62 & 0.79 & 0.60 & 0.59 & 0.41 & 0.74 \\
Claude Opus 4.6 (Thinking) & 0.78 & 0.63 & 0.74 & 0.75 & 0.86 & 0.58 & 0.79 & 0.60 & 0.60 & 0.36 & 0.69 \\
GPT-5 & 0.78 & 0.67 & 0.75 & 0.70 & 0.90 & 0.68 & 0.58 & 0.50 & 0.65 & 0.52 & 0.61 \\
\end{longtable}
}

\textbf{Table J.6:} \emph{Normalized hard-subset round-trip translation score by configuration and Tier 3 target language. *The Mean column averages all 29 target languages.}

\subsection{J.3 Judge agreement by language}\label{j.3-judge-agreement-by-language}

{\def\LTcaptype{none} 
\begin{longtable}[]{@{}
  >{\raggedright\arraybackslash}p{(\linewidth - 10\tabcolsep) * \real{0.1304}}
  >{\raggedleft\arraybackslash}p{(\linewidth - 10\tabcolsep) * \real{0.1739}}
  >{\raggedleft\arraybackslash}p{(\linewidth - 10\tabcolsep) * \real{0.1739}}
  >{\raggedleft\arraybackslash}p{(\linewidth - 10\tabcolsep) * \real{0.1739}}
  >{\raggedleft\arraybackslash}p{(\linewidth - 10\tabcolsep) * \real{0.1739}}
  >{\raggedleft\arraybackslash}p{(\linewidth - 10\tabcolsep) * \real{0.1739}}@{}}
\toprule\noalign{}
\begin{minipage}[b]{\linewidth}\raggedright
Language
\end{minipage} & \begin{minipage}[b]{\linewidth}\raggedleft
Krippendorff \(\alpha\)
\end{minipage} & \begin{minipage}[b]{\linewidth}\raggedleft
Mean pairwise r
\end{minipage} & \begin{minipage}[b]{\linewidth}\raggedleft
Exact
\end{minipage} & \begin{minipage}[b]{\linewidth}\raggedleft
Within-one
\end{minipage} & \begin{minipage}[b]{\linewidth}\raggedleft
n
\end{minipage} \\
\midrule\noalign{}
\endhead
\bottomrule\noalign{}
\endlastfoot
ar & 0.74 & 0.73 & 69\% & 93\% & 4099 \\
as & 0.79 & 0.80 & 59\% & 91\% & 4094 \\
cs & 0.73 & 0.73 & 75\% & 95\% & 4100 \\
de & 0.67 & 0.66 & 77\% & 94\% & 4099 \\
dz & 0.90 & 0.90 & 68\% & 95\% & 4071 \\
es & 0.68 & 0.67 & 75\% & 94\% & 4099 \\
eu & 0.74 & 0.75 & 55\% & 89\% & 4090 \\
fi & 0.68 & 0.69 & 55\% & 87\% & 4088 \\
\end{longtable}
}

\textbf{Table J.7:} \emph{Inter-judge agreement per language on the hard subset, over the 82 published configurations. Per-language n is the number of scored round-trips (50 hard items × 82 configurations, less a small number of judge or generation failures; a round-trip is retained if at least two judges returned a parseable score, so per-judge failures do not drop it). Exact and within-one are pairwise agreement rates (Appendix I.2); Krippendorff's \(\alpha\) is ordinal with native missing-data handling, and mean pairwise r is Fisher-z averaged over judge pairs. Cross-language unweighted means (\(\alpha\) 0.77, r 0.77, within-one 93\%) are the §5.3 figures; on the full corpus the means are \(\alpha\) 0.77, r 0.80, within-one 95\% (237,277 scored round-trips against 118,623 on the hard subset). Agreement is computed on the English-to-English judgments and is high in every language, including the lowest-resource ones.}

\end{landscape}

\end{document}